# Message-Based Web Service Composition, Integrity Constraints, and Planning under Uncertainty: A New Connection

**Jörg Hoffmann**                                               JOE.HOFFMANN@SAP.COM
*SAP Research*
*Karlsruhe, Germany*

**Piergiorgio Bertoli**                                               BERTOLI@FBK.EU
*Fondazione Bruno Kessler*
*Trento, Italy*

**Malte Helmert**                                 HELMERT@INFORMATIK.UNI-FREIBURG.DE
*Albert-Ludwigs-Universität Freiburg*
*Freiburg, Germany*

**Marco Pistore**                                               PISTORE@FBK.EU
*Fondazione Bruno Kessler*
*Trento, Italy*

## Abstract

Thanks to recent advances, AI Planning has become the underlying technique for several applications. Figuring prominently among these is automated Web Service Composition (WSC) at the "capability" level, where services are described in terms of preconditions and effects over ontological concepts. A key issue in addressing WSC as planning is that ontologies are not only formal vocabularies; they also axiomatize the possible relationships between concepts. Such axioms correspond to what has been termed "integrity constraints" in the actions and change literature, and applying a web service is essentially a belief update operation. The reasoning required for belief update is known to be harder than reasoning in the ontology itself. The support for belief update is severely limited in current planning tools.

Our first contribution consists in identifying an interesting special case of WSC which is both significant and more tractable. The special case, which we term *forward effects*, is characterized by the fact that every ramification of a web service application involves at least one new constant generated as output by the web service. We show that, in this setting, the reasoning required for belief update simplifies to standard reasoning in the ontology itself. This relates to, and extends, current notions of "message-based" WSC, where the need for belief update is removed by a strong (often implicit or informal) assumption of "locality" of the individual messages. We clarify the computational properties of the forward effects case, and point out a strong relation to standard notions of planning under uncertainty, suggesting that effective tools for the latter can be successfully adapted to address the former.

Furthermore, we identify a significant sub-case, named *strictly forward effects*, where an actual compilation into planning under uncertainty exists. This enables us to exploit off-the-shelf planning tools to solve message-based WSC in a general form that involves powerful ontologies, and requires reasoning about partial matches between concepts. We provide empirical evidence that this approach may be quite effective, using Conformant-FF as the underlying planner.

## 1. Introduction

Since the mid-nineties, AI Planning tools have become several orders of magnitude more scalable, through the invention of automatically generated heuristic functions and other search techniques (see McDermott, 1999; Bonet & Geffner, 2001; Hoffmann & Nebel, 2001; Gerevini, Saetti, & Serina, 2003; Helmert, 2006; Chen, Wah, & Hsu, 2006). This has paved the way to the adoption of planning as the underlying technology for several applications. One such application area is web service composition (WSC), by which in this paper we mean the automated composition of semantic web services (SWS). SWS are pieces of software advertised with a formal description of what they do. *Composing* SWS means to link them together so that their aggregate behavior satisfies a complex user requirement. The ability to automatically compose web services is the key to reducing human effort and time-to-market when constructing integrated enterprise applications. As a result, there is a widely recognized economic potential for WSC.

In the wide-spread SWS frameworks OWL-S[1] and WSMO[2], SWS are described at two distinct "levels". One of these addresses the overall functionality of the SWS, and the other details precisely how to interact with the SWS. At the former level, called "service profile" in OWL-S and "service capability" in WSMO, SWS are described akin to planning operators, with preconditions and effects. Therefore, planning is a prime candidate for realizing WSC at this level. This is the approach we follow in our paper.

In such a setting, a key aspect is that SWS preconditions and effects are described relative to an *ontology* which defines the formal (logical) vocabulary. Indeed, ontologies are much more than just formal vocabularies introducing a set of logical concepts. They also define *axioms* which constrain the behavior of the domain. For instance, an ontology may define a subsumption relationship between two concepts $A$ and $B$, stating that all members of $A$ are necessarily members of $B$. The natural interpretation of such an axiom, in the context of WSC, is that every state that can be encountered – every possible configuration of domain entities – must satisfy the axiom. In that sense, ontology axioms correspond to *integrity constraints* as discussed in the actions and change literature (Ginsberg & Smith, 1988; Eiter & Gottlob, 1992; Brewka & Hertzberg, 1993; Lin & Reiter, 1994; McCain & Turner, 1995; Herzig & Rifi, 1999).[3] Hence WSC as considered here is like planning in the presence of integrity constraints. Since the constraints affect the outcome of action executions, we are facing the frame and ramification problems, and execution of actions corresponds closely to complex notions such as belief update (Lutz & Sattler, 2002; Herzig, Lang, Marquis, & Polacsek, 2001). Unsurprisingly, providing such support for integrity constraints in the modern scalable planning tools mentioned above poses serious challenges. To the best of our knowledge, it has yet to be attempted at all.

Regarding the existing WSC tools, or planning tools employed for solving WSC problems, the situation isn't much better. Most tools ignore the ontology, i.e., they act as if no constraints on the domain behavior were given (Ponnekanti & Fox, 2002; Srivastava, 2002; Narayanan & McIlraith, 2002; Sheshagiri, desJardins, & Finin, 2003; Pistore, Traverso, & Bertoli, 2005b; Pistore, Marconi, Bertoli, & Traverso, 2005a; Agarwal, Chafle, Dasgupta, Karnik, Kumar, Mittal, & Srivastava, 2005a). Other approaches tackle the full generality of belief update by using general reasoners, and suffer from the inevitable performance deficiencies (Eiter, Faber, Leone, Pfeifer, & Polleres, 2003; Giunchiglia, Lee, Lifschitz, McCain, & Turner, 2004).

---

1. For example, see the work of Ankolekar et al. (2002) and Burstein et al. (2004).

2. For example, see the work of Roman et al. (2005) and Fensel et al. (2006).

3. Integrity constraints are sometimes also called *state constraints* or *domain constraints*.

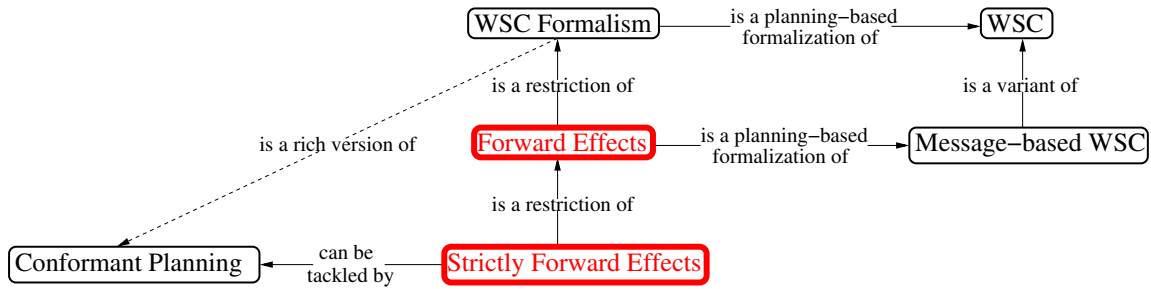

Figure 1: An overview of the planning and WSC frameworks addressed in this paper. Special cases identified herein shown in red / boldface.

Our work addresses the middle ground between these two extremes, i.e., the trade-off between expressivity and scalability in WSC. We do so via the identification of special cases that can be tackled more efficiently. Figure 1 gives an overview of the WSC and planning frameworks involved.

In brief, the forward effects case requires that *every effect and ramification of a web service affects at least one new constant that was generated as the web service's output.* In this situation, the frame problem trivializes, making the planning problem more similar to common notions of conformant planning (Smith & Weld, 1998; Bonet & Geffner, 2000; Cimatti, Roveri, & Bertoli, 2004; Hoffmann & Brafman, 2006). We will discuss how existing tools for the latter, in particular Conformant-FF (Hoffmann & Brafman, 2006), can be extended to deal with WSC under forward effects. With strictly forward effects, where action effects are required to affect *only* outputs, we devise an actual compilation into conformant planning. We thus obtain a scalable tool for interesting WSC problems with integrity constraints. In particular we are able to exploit (some of) the heuristic techniques mentioned above (Hoffmann & Nebel, 2001; Hoffmann & Brafman, 2006).

In what follows, we will explain the various parts of Figure 1 in a little more detail. Our starting point is a WSC formalism, addressing WSC in terms of planning in the presence of integrity constraints, as discussed above. The formalism is essentially an enriched form of conformant planning. Its distinguishing aspects are:

- The initial state description is a conjunction of literals (possibly not mentioning some of the logical facts in the task, and hence introducing uncertainty).

- Actions have a conditional effects semantics, meaning they can be executed in any state, but have an effect only if they are applicable.

- Actions may have output variables, i.e., they may create new constants.

- There is a set of integrity constraints, each of which is a universally quantified clause.

- The semantics of action execution is defined in terms of a belief update operation.

Section 2 below provides more details on these choices, and motivates them with an example and results from the literature. As we will show, planning in the formalism is very hard. Particularly, even just testing whether a given action sequence is a plan is $\Pi_2^p$-complete. This is in contrast to the more common notions of conformant planning, where plan testing is "only" coNP-complete.

As we will see, forward effects remove the additional complexity. Intuitively, the forward effects case covers the situation where a web service outputs some new constants, sets their characteristic properties relative to the inputs, and relies on the ontology axioms to describe any ramifications concerning the new constants. This case is syntactically characterized as follows:

(1) Every effect literal contains at least one output variable.

(2) Within each integrity constraint, every literal has the same set of variables in its arguments.

This definition is best understood with an example. Consider the following variant of the wide-spread "virtual travel agency" (VTA). Web services that book travel and accommodation must be linked. These web services generate new constants corresponding to tickets and reservations. For example, there are integrity constraints stating subsumption, such as $\forall z : trainTicket(z) \Rightarrow ticket(z)$. A web service **bookTicket** may have the input variable $x$, the precondition $train(x)$, the output variable $y$, and the effect $trainTicket(y) \wedge ticketFor(y, x)$. This is a forward effects task: every effect literal contains the output variable $y$, and the integrity constraint has the single variable $z$ which provides the arguments of all literals in the constraint. Say one instantiates the input of **bookTicket** with a constant $c$ and its output with a new constant $d$. When applying the resulting ground action to a state where $train(c)$ holds true, the constant $d$ gets created, and its characteristic properties relative to the inputs – $trainTicket(d) \wedge ticketFor(d, c)$ – are set directly by the action. The integrity constraint takes care of the ramification, establishing that $ticket(d)$ holds. Note that the status of $c$ – apart from its relation to $d$ – is not affected in any way. [4]

The forward effects case is closely related to a wide-spread notion of WSC problems, which we refer to as "message-based WSC". In such approaches, the composition semantics is based on chaining over input and output messages of web services, in one or the other sense. Inferences from ontology axioms can be made in many of these approaches, but only in a restricted way limited by an assumption of "locality" of the individual messages, where the interferences affect only a particular message transfer, and any implications for other transfers are ignored. This locality assumption is usually made in an informal way, and often not stated explicitly at all. One contribution of our work is to shed some light on this issue, via the identification of the forward effects case which lies "in between" message-based WSC and a full planning framework with belief update semantics.

Both message-based WSC and the forward effects case share the focus on output constants. There are two important differences. First, the forward effects case is more restricted than message-based WSC in terms of the ontology axioms allowed. Essentially, forward effects correspond to a special case of WSC where the locality assumption of message-based WSC is actually justified, within a full planning framework. Second, that full framework comes with the benefit of increased flexibility in the combination of services, because locality is not enforced (e.g. the output of one service may be reused at several points in a plan).

From a computational point of view, the key property of the forward effects case is that it removes the need for belief update. In a nutshell, the reason is that actions affect only "new" propositions, i.e., propositions involving at least one output constant. (Recall here the point made about the unchanged status of $c$, in the VTA example above.) The output constant ($d$, in the example) does not exist prior to the application of the action, and hence the previous belief carries no knowledge about it and need not be revised. Consider the characterization of forward effects, as given above. Condition (1) ensures that the immediate effect of the action affects only new propositions. Condition (2) ensures that any changes on new propositions only propagate to new propositions. Since all literals in a constraint share the same variables, the output constant in question is copied to all of them. As we will see, by virtue of these properties the complexity of plan testing is coNP-complete, rather than $\Pi_2^p$-complete, in the forward effects case.

---

4. The latter would *not* be the case if the effect of **bookTicket** included a literal affecting only $x$ (example: $\neg train(x)$), or if there was an integrity constraint capable of mixing "old" and "new" constants (example: $\forall x, y : trainTicket(y) \Rightarrow \neg train(x)$).

This complexity reduction is critical because the reduced complexity is the same as in the more common situation of conformant planning under initial state uncertainty. Therefore it should be feasible to adapt conformant planning tools to address WSC with forward effects. Scalable planning tools for conformant planning have already been developed (Cimatti et al., 2004; Bryce, Kambhampati, & Smith, 2006; Hoffmann & Brafman, 2006; Palacios & Geffner, 2007). Hence this is a promising line of research. As an example, we will focus on the Conformant-FF tool (Hoffmann & Brafman, 2006) (short CFF) and outline the main steps that need to be taken in adapting CFF to handle WSC with forward effects.

We then identify a case where an actual *compilation* into conformant planning under initial state uncertainty exists. For that, one must fix a set of constants a priori. In a manner that is fairly standard (see, e.g., the Settlers domain of Long & Fox, 2003), we simply include in that set a subset of "potential constants" that can be used to instantiate outputs. The more subtle idea we put forward is to identify a condition on the actions under which we can "predict" which properties will be assigned to which potential constants, in case they are created. This enables us to design a compilation that moves all action effects into the initial state formula, and uses actions only to modify the set of constants that already exist. In this way, reasoning about the initial state formula in the compiled task is the same as reasoning about output constants in the original task, and the reasoning mechanisms included in tools such as CFF can be naturally used to implement the latter.

Our trick for predicting output properties is to require that all actions are *compatible* in the sense that they either produce different outputs, or have the same effects. It turns out that this condition is naturally given in a restriction of forward effects, which we call *strictly forward effects*, where the web service effects concern *only* new constants.

Clearly, not being able to reference the inputs is a limitation. For example, we can no longer say, in the above VTA example, that the output $y$ is a *ticket for* the input $x$. Still, the strictly forward effects case describes an interesting class of WSC problems. That class corresponds to web services modeled as in the early versions of OWL-S, for example, where there was no logical connection between inputs and outputs. Further, this class of WSC problems allows powerful ontologies – universally quantified clauses – and makes it possible to combine services very flexibly. Using our compilation, this class of problems can be solved by off-the-shelf tools for planning under uncertainty.

We validate the compilation approach empirically by running a number of tests using CFF as the underlying planner. We use two test scenarios, both of which are scalable in a variety of parameters, covering a range of different problem structures. We examine how CFF reacts to the various parameters. Viewed in isolation, these results demonstrate that large and complex WSC instances can be comfortably solved using modern planning heuristics.

A comparison to alternative WSC tools is problematic due to the widely disparate nature of what kinds of problems these tools can solve, what kinds of input languages they understand, and what purpose the respective developers had in mind. To nevertheless provide some assessment of the comparative benefits of our approach we run tests with the DLVK tool by Eiter et al. (2003) and Eiter, Faber, Leone, Pfeifer, and Polleres (2004). DLVK is one of the few planning tools that deals with ontology axioms – called "static causal rules" – directly, without the need to restrict to forward effects and without the need for a compilation. Since, in the context of our work, the main characteristic of WSC is the presence of ontology axioms, this means that DLVK is one of the few existing "native WSC tools". By comparison, our forward effects compilation approach solves a similar problem, but sacrifices some expressivity. The question is, can we in principle gain anything from this sacrifice? Absolutely, the answer is "yes". DLVK is much slower than compilation+CFF, solving only a small fraction of our test instances even when always provided with the correct plan length bound. We emphasize that we do not wish to over-state these results, due to the above-mentioned differences between the tools. The only conclusion we draw is that the trade-off between expressivity and scalability in WSC is important, and that the forward effects case seems to constitute an interesting point in that trade-off.

The paper is organized as follows. First, Section 2 provides some further background necessary to understand the context and contribution of our work. Section 3 introduces our WSC planning formalism. Section 4 defines and discusses forward effects. Section 5 introduces our compilation to planning under uncertainty, and Section 6 presents empirical results. We discuss the most closely related work at the relevant points during the text, and Section 7 provides a more complete overview. Finally, Section 8 concludes and discusses future work. To improve readability, most proofs are moved into Appendix A and replaced in the text by proof sketches.

## 2. Background

The context of our work is rather intricate. WSC as such is a very new topic posing many different challenges to existing techniques, with the effect that the field is populated by disparate works differing considerably in their underlying purpose and scope. In other words, the "common ground" is fairly thin in this area. Further, our work actually involves three fields of research – WSC, planning, and reasoning about actions and change – which are all relevant to understanding our contribution. For these reasons, we now explain this background in some detail. We first discuss WSC in general, and WSC as Planning in particular. We then state the relevant facts about belief update. We finally consider "message-based" WSC.

### 2.1 WSC, and WSC as Planning

Composition of semantic web services has received considerable attention in the last few years. A general formulation of the problem, shared by a large variety of works, focuses on the "capability" level, where each web service is conceived as an atomic operator that transforms concepts. More specifically, a service is defined via an "IOPE" description: the service receives as input a set $I$ of typed objects, and, provided some precondition $P$ on $I$ holds, produces as output a set $O$ of typed objects for which some effect $E$ is guaranteed to hold. The typing of the objects exchanged by the services is given in terms of their membership in *concepts*. Concepts are classes defined within *ontologies*, which exploit Description Logics (DL), or some other form of logic, to formally define the universe of concepts admitted in the discourse. An ontology can express complex relationships among concepts, like a subsumption hierarchy, or the way objects belonging to a concept are structured into parts referring to other concepts.

This general setting can be instantiated in various ways depending on the kind of conditions admitted as preconditions/effects of services, and on the kind of logics underlying the ontology definitions. Independent of this, the problem of semantic web service composition can be stated as one of *"linking appropriately a set of existing services so that their aggregate behavior is that of a desired service (the goal)"*. To illustrate this problem, consider the following example, which is inspired by the work of Thakkar, Ambite, and Knoblock (2005) on e-services for bioinformatics (and relies on the actual structure of proteins, see for example Petsko & Ringe, 2004; Branden & Tooze, 1998; Chasman, 2003; Fersht, 1998):

**Example 1** *Say we want to compose a web service that provides information about different classes of proteins. The ontology states which classes of proteins exist, and which structural characteristics may occur. We have available an information service for every structural characteristic, and a presentation service that combines a range of information. Given a particular protein class, the composed web service should run the relevant information services, and present their output.*

*Concretely, classes of proteins are distinguished by their location (cell, membrane, intermembrane, ...). This is modeled by predicates protein($x$), cellProtein($x$), membraneProtein($x$), intermembraneProtein($x$), along with sub-concept relations such as $\forall x : cellProtein(x) \Rightarrow protein(x)$. An individual protein is characterized by the following four kinds of structures:*

1. *The "primary structure" states the protein's sequence of amino-acids, e.g., 1kw3($x$) (a protein called "Glyoxalase") and 1n55($x$) (a protein called "Triosephosphate Isomerase").*

2. *The "secondary structure" states the protein's external shape in terms of a DSSP ("Dictionary of Secondary Structure for Proteins") code, admitting a limited set of possible values. For example, G indicates a 3-turn helix, B a $\beta$-sheet, and so on. The total set of values is G,H,I,T,E,B,S.*

3. *The "tertiary structure" categorizes the protein's 3-D shape.*

4. *For a subset of the proteins, a "quaternary structure" categorizes the protein's shape when combined in complexes of proteins (amounting to about 3000 different shapes, see for example 3DComplex.org, 2008).*

*There are various axioms that constrain this domain, apart from the mentioned subconcept relations. First, some obvious axioms specify that each protein has a "value" in each of the four kinds of structures (i.e., the protein has a sequence of amino-acids, an external shape, etc). However, there are also more complex axioms. Particular kinds of proteins come only with particular structure values. This is modeled by axioms such as:*

$$\forall x : \neg cellProtein(x) \lor G(x) \lor \neg 1n55(x)$$

$$\forall x : \neg cellProtein(x) \lor \neg B(x) \lor 1kw3(x) \lor complexBarrel(x)$$

*For each DSSP code Z there is an information service, named* **getInfoDSSP**$_Z$*, whose precondition is Z($x$) and whose effect is InfoDSSP($y$) where $y$ is an output of the service. Similarly, we have information services for amino-acids, 3-D shapes, and shapes in complexes. The presentation service, named* **combineInfo***, requires that information on all four kinds of structures has been created, and has the effect combinedPresentation($y$) (where $y$ is an output of* **combineInfo***).*

*The input to the composed web service is a protein c (a logical constant) and its class. The goal is $\exists x : combinedPresentation(x)$. A solution is to reason about which characteristics may occur, to apply the respective information services, and then to run **combineInfo**. In a variant of the problem, an additional **requestInfo** service is used to initiate the information request, i.e., the output of **requestInfo** is the protein c and its class.*

This example shows how ontology axioms play a crucial role in our form of WSC, formulating complex dependencies between different concepts. Note that applying a web service may have indirect consequences implied by the ontology axioms. In the example, the output of the **requestInfo** service has implications for which kinds of information services are required.

Another interesting aspect of Example 1 is that it requires what the SWS community calls "partial matches", as opposed to "plug-in matches" (Paolucci, Kawamura, Payne, & Sycara, 2002; Li & Horrocks, 2003; Kumar, Neogi, Pragallapati, & Ram, 2007).[5] Consider the situation where one wants to "connect" a web service $w$ to another web service $w'$. That is, $w$ will be executed prior to $w'$, and the output of $w$ will be used to instantiate the input of $w'$. Then $w$ and $w'$ are said to have a *partial match* if, given the ontology axioms, the output of $w$ *sometimes* suffices to provide the necessary input for $w'$. By contrast, $w$ and $w'$ are said to have a *plug-in match* if, given the ontology axioms, the output of $w$ *always* suffices to provide the necessary input for $w'$.

Plug-in matches are tackled by many approaches to WSC, whereas partial matches are tackled only by few. Part of the reason probably is that plug-in matches are easier to handle, in many types of WSC algorithms. Indeed most existing WSC tools support plug-in matches only (see a detailed discussion of WSC tools in Section 7). Example 1 cannot be solved with plug-in matches because each of the information services provides the necessary input for the **combineInfo** service only in some particular cases.

We base our work on a planning formalism that allows to specify web services (i.e., actions) with outputs, and that allows to specify ontology axioms. The axioms are interpreted as integrity constraints, and the resulting semantics corresponds closely to the common intuitions behind WSC, as well as to the existing formal definitions related to WSC (Lutz & Sattler, 2002; Baader, Lutz, Milicic, Sattler, & Wolter, 2005; Liu, Lutz, Milicic, & Wolter, 2006b, 2006a; de Giacomo, Lenzerini, Poggi, & Rosati, 2006). Since one of our main aims is to be able to exploit existing planning techniques, we consider a particular form of ontology axioms, in correspondence with the representations that are used by most of the existing tools for planning under uncertainty. Namely, the axioms are universally quantified clauses. An example is the subsumption relation $\forall x : trainTicket(x) \Rightarrow ticket(x)$ mentioned above, where as usual $A \Rightarrow B$ is an abbreviation for $\neg A \lor B$. A planning task specifies a set of such clauses, interpreted as the conjunction of the clauses. Note that this provides significant modeling power. The meaning of the universal quantification in the clauses is that the clauses hold for all planning "objects" – logical constants – that are known to exist. In that sense, the interpretation of formulas is *closed-world* as is customary in planning tools. However, in contrast to most standard planning formalisms including PDDL, we do not assume a fixed set of constants. Rather, the specification of actions with outputs enables the dynamic creation of new constants. The quantifiers in the ontology axioms range over all constants that exist *in the respective world*. In a similar fashion, the planning goal may contain variables, which are existentially quantified. The constants used to instantiate the goal may have pre-existed, or they may have been generated as the outputs of some of the web services that were applied on the path to the world. Consider for illustration the goal $\exists x : \text{combinedPresentation}(x)$ in Example 1, where the goal variable $x$ will have to be instantiated with an output created by the **combineInfo** service.

---

5. The terminology in these works is slightly different from what we use here, and they also describe additional kinds of matches. Some details are given in Section 7.

Another important aspect of our planning formalism is that we allow incomplete initial state descriptions. The initial state corresponds to the input that the user provides to the composed web service. Certainly we cannot assume that this contains complete information about every aspect of the world. (In Example 1, the initial state tells us which class of proteins we are interested in, but leaves open what the consequences are regarding the possible structural characteristics.) We consider the case where there is no observability, i.e., conformant planning. The outcome of WSC is a sequence of web services that satisfies the user goal in all possible situations.[6] As is customary in conformant planning, the actions have a *conditional effects* semantics, i.e., they fire if their precondition holds true, and otherwise they do nothing. Note that, this way, we obtain a notion of partial matches: the solution employs different actions depending on the situation.

The main difference between our planning formalism and the formalisms underlying most current planning tools is the presence of integrity constraints, and its effect on the semantics of executing actions. That semantics is defined as a belief update operation.

## 2.2 Belief Update

The correspondence of web service applications to belief update was first observed by Lutz and Sattler (2002), and followed by Baader et al. (2005), Liu et al. (2006b, 2006a) and de Giacomo et al. (2006). In the original statement of the belief update problem, we are given a "belief" $\Phi$, i.e., a logical formula defining the worlds considered possible. We are further given a formula $\phi$, the "update". Intuitively, $\phi$ corresponds to some observation telling us that the world has changed in a way so that, now, $\phi$ is true. We want to obtain a formula $\Phi'$ defining the worlds which are possible given this update. Certainly, we need to have that $\Phi' \models \phi$. Ensuring this corresponds to the well-known ramification problem. At the same time, however, the world should not change unnecessarily. That is, we want $\Phi'$ to be "as close as possible to $\Phi$", among the formulas which satisfy $\phi$. This corresponds to the frame problem.

Say we want to apply an action $a$ in the presence of integrity constraints. $\Phi$ describes the worlds that are possible prior to the application of $a$. $\Phi'$ is the resulting set of possible worlds. The integrity constraints correspond to a formula $\Phi_{IC}$ which holds in $\Phi$, and which we require to hold in $\Phi'$. The update formula $\phi$ is given as the conjunction of the action effect with $\Phi_{IC}$, i.e., we have $\phi = \text{eff}_a \wedge \Phi_{IC}$. This means that we update our previous belief with the information that, after $a$, $\text{eff}_a$ is a new formula required to hold, and $\Phi_{IC}$ is still true. For example, we may have an action effect $A(c)$ and a subsumption relation between concepts $A$ and $B$, formulated as a clause $\forall x : \neg A(x) \vee B(x)$. Then the update formula $A(c) \wedge \forall x : \neg A(x) \vee B(x)$ ensures that $B(c)$ is true in $\Phi'$.

Belief update has been widely considered in the literature on AI and databases (see for example Fagin, Kuper, Ullman, & Vardi, 1988; Ginsberg & Smith, 1988; Winslett, 1988, 1990; Katzuno & Mendelzon, 1991; Herzig, 1996; Herzig & Rifi, 1999; Liu et al., 2006b; de Giacomo et al., 2006). The various approaches differ in exactly how $\Phi'$ should be defined. The best consensus is that there is no one approach that is most adequate in every application context. All approaches agree that $\phi$ should hold in the updated state of affairs, $\Phi' \models \phi$. Major differences lie in what exactly it should be taken to mean that $\Phi'$ should be "as close as possible to $\Phi$". Various authors, for example Brewka and Hertzberg (1993), McCain and Turner (1995), Herzig (1996), and Giunchiglia and Lifschitz (1998), argue that a notion of causality is needed, in addition to (or even instead of) a notion of integrity constraints, to model domain behavior in a natural way. We do not counter these arguments, but neither do we follow a causal approach in our work. The reason is that ontologies in the context of WSC, for example ontologies formulated in the web ontology language OWL (McGuinness & van Harmelen, 2004), do not incorporate a notion of causality. All we are given is a set of axioms, made with the intention to describe the behavior of the domain itself, rather than the behavior it exhibits when changed by some particular web services. Our idea in this work is to try to leverage on what we have (or what we are reasonably close to having). Consideration of causal approaches in WSC is left for future work.

---

6. Of course, more generally, observability is partial and web service effects are also uncertain. We do not consider these generalizations here. Extending our notions accordingly should be straightforward, and is future work.

Belief update is a computationally very hard problem. Eiter and Gottlob (1992) and Liberatore (2000) show that, for the non-causal approaches to defining $\Phi'$, reasoning about $\Phi'$ is typically harder than reasoning in the class of formulas used for formulating $\Phi$ and $\phi$. Specifically, deciding whether or not a particular literal is true in $\Phi'$ is $\Pi_2^2$-hard even if $\Phi$ is a complete conjunction of literals (corresponding to a single world state) and $\phi$ is a propositional CNF formula. The same problem is coNP-hard even if $\Phi$ is a single world state and $\phi$ is a propositional Horn formula. We use these results to show that, in our planning formalism, *checking* a plan – testing whether or not a given action sequence is a plan – is $\Pi_p^2$-complete, and deciding polynomially bounded plan existence is $\Sigma_p^3$-complete.

Given this complexity, it is perhaps unsurprising that the support for integrity constraints in current planning tools is severely limited. The only existing planning tools that do support integrity constraints, namely those by Eiter et al. (2003) and Giunchiglia et al. (2004), are based on generic deduction, like satisfiability testing or answer set programming. They hence lack the planning-specific heuristic and search techniques that are the key to scalability in the modern planning tools developed since the mid-nineties. It has not even been investigated yet if and how integrity constraints could be handled in the latter tools. The only existing approach that ventures in this direction implements so-called *derived predicates* in some of the modern planning tools (Thiébaux, Hoffmann, & Nebel, 2005; Gerevini, Saetti, Serina, & Toninelli, 2005; Chen et al., 2006). This approach postulates a strict distinction between "basic" predicates that may be affected by actions, and "derived" predicates that may be affected by integrity constraints taking the form of logic programming rules. If a predicate appears in an action effect, then it is not allowed to appear in the head of a rule. This is not a desirable restriction in the context of WSC, where web services are bound to affect properties that are constrained by ontology axioms.

The existing work connecting WSC with belief update (Lutz & Sattler, 2002; Baader et al., 2005; Liu et al., 2006b, 2006a; de Giacomo et al., 2006) is of a theoretical nature. The actual implemented WSC tools make severe simplifying assumptions. Most often, that assumption is to ignore the ontology axioms (Ponnekanti & Fox, 2002; Srivastava, 2002; McIlraith & Son, 2002; Sheshagiri et al., 2003; Sirin, Parsia, Wu, Hendler, & Nau, 2004; Pistore et al., 2005b, 2005a). Sometimes, the ontology constraints are restricted to subsumption hierarchies, which makes the update problem easy (Constantinescu & Faltings, 2003; Constantinescu, Faltings, & Binder, 2004b, 2004a). Sirin and Parsia (2004) and Sirin, Parsia, and Hendler (2006) discuss the problem of dealing with ontology axioms in WSC, but do not make a connection to belief update, and describe no alternative solution. Finally, some authors, for example Meyer and Weske (2006), do deal with ontology ax-ioms during composition, but do not provide a formal semantics and do not specify exactly how action applications are handled. It seems that these not fully formalized WSC approaches implicitly assume a *message-based* framework. Those frameworks are closely related to the forward effects special case identified herein.

## 2.3 Message-Based WSC

In message-based approaches to WSC, the composition semantics is based on chaining over input and output messages of web services. The word "message" is not a standard term in this context. Most authors use their own individual vocabulary. As far as we are aware, the first appearance of the word "message" in a WSC paper title is in the work by Liu, Ranganathan, and Riabov (2007). This work describes message-based WSC as follows. A solution is a directed acyclic graph (DAG) of web services, where the input needed for web service (DAG graph node) $w$ must be provided by the outputs of the predecessors of $w$ in the graph. That is, the plan determines fixed connections between the actions. Reasoning, then, only takes place "within these connections". Any two connections between different output and input messages, i.e., any two graph edges ending in a different node, are assumed to be mutually independent. Consider the following example for illustration. Say a web service $w$ has the effect $hasAttributeA(c,d)$ where $d$ is an output constant and $c$ is an input (i.e., $c$ existed already prior to application of $w$). Say there is an axiom $\forall x, y : hasAttributeA(x,y) \Rightarrow conceptB(x)$ expressing an "attribute domain restriction". If $x$ has $y$ as a value of attribute $A$, then $x$ must be of concept $B$. Given this, $w$'s effect implies $conceptB(c)$. Now, suppose that our belief prior to applying $w$ did not constrain $c$ to be of concept $B$. Then applying $w$ leads to new knowledge about $c$. Hence we need a non-trivial belief update taking into account the changed status of $c$, and any implications that may have. Message-based WSC simply acts as if the latter is not the case. It only checks whether $w$ correctly supplies the inputs of the web services $w'$ that $w$ is connected to. That is, the new fact $hasAttributeA(c,d)$ may be taken as part of a proof that the effect of $w$ implies the precondition of a connected web service $w'$. But it is not considered at all what implications $hasAttributeA(c,d)$ may have with respect to the previous state of affairs. In that sense, message-based WSC "ignores" the need for belief update.

The intuitions underlying message-based WSC are fairly wide-spread. Many papers use them in a more or less direct way. There are many approaches that explicitly define WSC solutions to be DAGs with local input/output connections as above (Zhan, Arpinar, & Aleman-Meza, 2003; Lecue & Leger, 2006; Lecue & Delteil, 2007; Kona, Bansal, Gupta, & Hite, 2007; Liu et al., 2007; Ambite & Kapoor, 2007). In various other works (Constantinescu & Faltings, 2003; Constantinescu et al., 2004b, 2004a; Meyer & Weske, 2006), the message-based assumptions are more implicit. They manifest themselves mainly in the sense that ontology axioms are only used to infer the properties of output messages, and often only for checking whether the inferences imply that a desired input message is definitely given.

Previous work on message-based WSC does not address at all how message-based WSC relates to the various notions, like belief update, considered in the literature. One contribution of our work is to shed some light on this issue, via the identification of the forward effects case which lies "in between" message-based WSC and a full planning framework with belief update semantics.

Both message-based WSC and the forward effects case share the focus on outputs. Indeed, the output constants generated by our actions can be viewed as "messages". An output constant represents an information object which is created by one web service, and which will form the input of some other web service. In the forward effects case, due to the restriction on axioms, the individual messages do not interact. This is much like message-based WSC. The main difference is this: while message-based WSC ignores any possible interactions, in forward effects there actually aren't any interactions, according to a formal planning-based execution semantics. In that sense, forward effects correspond to a special case of WSC where the assumptions of message-based WSC are justified.

Reconsider our example from above, featuring a web service $w$ with an effect implying that $conceptB(c)$ where $c$ is a pre-existing constant. As explained above, message-based WSC will simply ignore the need for updating the knowledge about $c$. In contrast, the forward effects case disallows the axiom $\forall x, y : hasAttributeA(x, y) \Rightarrow conceptB(x)$ *because* it may lead to new conclusions about the old belief (note that the literals in the axiom refer to different sets of variables).

The forward effects case also differs significantly from most approaches to message-based WSC in terms of the flexibility with which it allows to combine actions into plans. In the message-based approach using DAGs, a solution DAG ensures that the inputs of each service $w$ can always be provided by $w$'s predecessors. That is, we have a plug-in match between the set $W$ of $w$'s predecessors in the DAG, and $w$ itself. Note that this is slightly more general than the usual notion of plug-in matches, in that $|W|$ may be greater than 1, and hence each single service in $W$ may have only a partial match with $w$. This is the notion used, amongst others, by Liu et al. (2007). Other authors, for example Lecue and Leger (2006) and Lecue and Delteil (2007), are more restrictive in that they consider every individual input $x$ of $w$ in turn and require that there exists a $w' \in W$ so that $w'$ has a plug-in match with $x$ (i.e., $w'$ guarantees to always provide $x$). Even in the more generous of these two definitions, partial matches are restricted to appear locally, on DAG links. Every action/web service is required to be always executable at the point where it is applied. In other words, the services are used in a fixed manner, not considering the dynamics of actual execution. In Example 1, this would mean using the same information services regardless of the class of the protein, hence completely ignoring what is relevant and what is not.

The forward effects case incorporates a much more general notion of partial matches. This happens in a straightforward way, exploiting the existing notions from planning, in the form of a conditional effects semantics. The standard notion of a conformant solution defines how partial matches must work together on a global level, to accomplish the goal. To the best of our knowledge, there is only one other line of work on WSC, by Constantinescu et al. (Constantinescu & Faltings, 2003; Constantinescu et al., 2004b, 2004a), that incorporates a comparable notion of partial matches. In that work, web services are characterized in terms of input and output "types". To handle partial matches, so-called "switches" combine several web services in a way that ascertains all relevant cases can be covered. The switches are designed relative to a subsumption hierarchy over the types. Note that subsumption hierarchies are a special case of the much more general integrity constraints – universally quantified clauses – that we consider in our work.

## 3. Formalizing WSC

As a solid basis for addressing WSC, we define a planning formalism featuring integrity constraints, on-the-fly creation of output constants, incomplete initial state descriptions, and actions with a conditional effects semantics. The application of actions is defined as a belief update operation, following the *possible models approach* by Winslett (1988). That definition of belief update is somewhat canonical in that it is very widely used and discussed. In particular it underlies all the recent work relating to formalizations of WSC (Lutz & Sattler, 2002; Baader et al., 2005; Liu et al., 2006b, 2006a; de Giacomo et al., 2006; de Giacomo, Lenzerini, Poggi, & Rosati, 2007). As we will show further below (Section 4.3), most belief update operations are equivalent anyway as soon as we are in the forward effects case. Recall here that the forward effects case is the central object of investigation in this paper.

We first give the syntax of our formalism, which we denote with $\mathcal{WSC}$, then we give its semantics. We conclude with an analysis of its main computational properties.

### 3.1 Syntax

We denote predicates with $G, H, I$, variables with $x, y, z$, and constants with $c, d, e$. *Literals* are possibly negated predicates whose arguments are variables or constants. If all arguments are constants, the literal is *ground*. We refer to positive ground literals as *propositions*. Given a set $\mathcal{P}$ of predicates and a set $C$ of constants, we denote by $\mathcal{P}^C$ the set of all propositions that can be formed from $\mathcal{P}$ and $C$. Given a set $X$ of variables, we denote by $\mathcal{L}^X$ the set of all literals $l$ which use only variables from $X$. Note here that $l$ may use arbitrary predicates and constants.[7] If $l$ is a literal, we write $l[X]$ to indicate that $l$ has the variable arguments $X$. If $X = \{x_1, \ldots, x_k\}$ and $C = (c_1, \ldots, c_k)$, then by $l[c_1, \ldots, c_k/x_1, \ldots, x_k]$ we denote the respective substitution, abbreviated as $l[C]$. In the same way, we use the substitution notation for any construct involving variables. Slightly abusing notation, we use a vector of constants also to denote the set of constants appearing in it. Further, if a function $a$ assigns constants to the variables $X$, then by $l[a/X]$ we denote the substitution where each argument $x \in X$ was replaced with $a(x)$. We are only concerned with first-order logic, that is, whenever we write *formula* we mean a first-order formula. We denote true as $1$ and false as $0$.

A *clause*, or *integrity constraint*, is a disjunction of literals with universal quantification on the outside. The variables quantified over are exactly those that appear in at least one of the literals. For example, $\forall x, y : \neg G(x, y) \vee H(x)$ is an integrity constraint but $\forall x, y, z : \neg G(x, y) \vee H(x)$ and $\forall x : \neg G(x, y) \vee H(x)$ are not. An *operator* $o$ is a tuple $(X_o, \text{pre}_o, Y_o, \text{eff}_o)$, where $X_o, Y_o$ are sets of variables, $\text{pre}_o$ is a conjunction of literals from $\mathcal{L}^{X_o}$, and $\text{eff}_o$ is a conjunction of literals from $\mathcal{L}^{X_o \cup Y_o}$.[8] The intended meaning is that $X_o$ are the inputs and $Y_o$ the outputs, i.e., the new constants created by the operator. For an operator $o$, an *action* $a$ is given by $(\text{pre}_a, \text{eff}_a) \equiv (\text{pre}_o, \text{eff}_o)[C_a/X_o, E_a/Y_o]$ where $C_a$ and $E_a$ are vectors of constants. For $E_a$ we require that the constants are pairwise different – it makes no sense to "output the same new constant twice". Given an action $a$, we will refer to $a$'s inputs and outputs by $C_a$ and $E_a$, respectively. We will also use the notations $\text{pre}_a, \text{eff}_a$ with the obvious meaning.

A $\mathcal{WSC}$ *task*, or *planning task*, is a tuple $(\mathcal{P}, \Phi_{IC}, \mathcal{O}, C_0, \phi_0, \phi_G)$. Here, $\mathcal{P}$ is a set of predicates. $\Phi_{IC}$ is a set of integrity constraints. $\mathcal{O}$ is a set of operators and $C_0$ is a set of constants, the initial constants supply. $\phi_0$ is a conjunction of ground literals, describing the possible initial states. $\phi_G$ is a conjunction of literals with existential quantification on the outside, describing the goal states, e.g., $\exists x, y : G(x) \wedge H(y)$. All predicates are taken from $\mathcal{P}$, and all constants are taken from $C_0$. All constructs (e.g., sets and conjunctions) are finite. We will sometimes identify $\Phi_{IC}$ with the conjunction of the clauses it contains. Note that the existential quantification of the goal variables provides the option to instantiate the goal with constants created during planning – obtaining objects as requested by the goal may be possible only through the use of outputs.

---

7. One could of course introduce more general notations for logical constructs using some set of predicates or constants. However, herein the two notations just given will suffice.

8. As stated, we do not address disjunctive or non-deterministic effects. This is a topic for future work.

The various formulas occurring in $(\mathcal{P}, \Phi_{IC}, \mathcal{O}, C_0, \phi_0, \phi_G)$ may make use of constants from $C_0$. Specifically, this is the case for clauses in $\Phi_{IC}$ and for the goal formula $\phi_G$. Allowing such use of constants does not have any effect on our complexity or algorithmic results. It is conceivable that the feature may be useful. As a simple example, in the VTA domain the user may wish to select a particular train. Say the train company provides a table of trains with their itineraries. That table can be represented in $\phi_0$, possibly with help from $\Phi_{IC}$ stating constraints that hold for particular trains. The user can then select a train, say $ICE107$, and pose as a goal that $\exists y : ticketFor(y, ICE107)$. Constraining the produced ticket in this way would not be possible without the use of pre-existing constants (or would at least require a rather dirty hack, e.g., encoding the desired train in terms of a special predicate).

Operator descriptions, that is, preconditions and effects, may also use constants from $C_0$. The value of this is more benign than for $\Phi_{IC}$ and $\phi_G$ because one can always replace a constant $c$ in the precondition/effect with a new input/output variable $x$, and instantiate $x$ (during planning) with $c$. Note, however, that this would give the planner the option to (uselessly) instantiate $x$ with some other constant, and may hence affect planning performance. In our above example, there might be a special operator booking a ticket for $ICE107$ (e.g., if that train has particular ticketing regulations).

The correspondence of a $\mathcal{WSC}$ task to a web service composition task is fairly obvious. The set $\mathcal{P}$ of predicates is the formal vocabulary used in the underlying ontology. The set $\Phi_{IC}$ of integrity constraints is the set of axioms specified by the ontology, i.e., domain constraints such as subsumption relations. The set $\mathcal{O}$ of operators is the set of web services. Note that our formalization corresponds very closely to the notion of *IOPE descriptions*: inputs, outputs, preconditions, and effects (Ankolekar et al., 2002; Burstein et al., 2004). An action corresponds to a *web service call*, where the web service's parameters are instantiated with the call arguments.

The constructs $C_0$, $\phi_0$, and $\phi_G$ are extracted from the user requirement on the composition. We assume that such requirements also take the form of IOPE descriptions. Then, $C_0$ are the user requirement inputs, and $\phi_0$ is the user requirement precondition. In other words, $C_0$ and $\phi_0$ describe the input given to the composition by the user. Similarly, $\phi_G$ is the user requirement effect – the condition that the user wants to be accomplished – and the user requirement outputs are the (existentially quantified) variables in $\phi_G$.

## 3.2 Semantics

In what follows, assume we are given a $\mathcal{WSC}$ task $(\mathcal{P}, \Phi_{IC}, \mathcal{O}, C_0, \phi_0, \phi_G)$. To be able to model the creation of constants, *states* (also called *world states*) in our formalism are enriched with the set of constants that exist in them. A state $s$ is a pair $(C_s, I_s)$ where $C_s$ is a set of constants, and $I_s$ is a $C_s$-*interpretation*, i.e., a truth value assignment $I_s : \mathcal{P}^{C_s} \mapsto \{0, 1\}$. Quantifiers are taken to range over the constants that exist in a state. That is, if $I$ is a $C$-interpretation and $\phi$ is a formula, then by writing $I \models \phi$ we mean that $I \models \phi^C$ where $\phi^C$ is the same as $\phi$ except that all quantifiers were restricted to range over $C$. To avoid clumsy notation, we will sometimes write $s \models \phi$ to abbreviate $I_s \models \phi$.

The core definition specifies how the application of an action affects a state. This is defined through a form of belief update. Let us first define the latter. Assume a state $s$, a set of constants $C' \supseteq C_s$, and a formula $\phi$. We define $update(s, C', \phi)$ to be the set of interpretations that result from creating the constants $C' \setminus C_s$, and updating $s$ with $\phi$ according to the semantics proposed by Winslett (1988).

Say $I_1$ and $I_2$ are $C'$-interpretations. We define a partial order over such interpretations, by setting $I_1 <_s I_2$ if and only if

$$\{p \in \mathcal{P}^{C_s} \mid I_1(p) \neq I_s(p)\} \subset \{p \in \mathcal{P}^{C_s} \mid I_2(p) \neq I_s(p)\}. \tag{1}$$

In words, $I_1$ is ordered before $I_2$ iff it differs from $s$ in a proper subset of values. Given this, we can now formally define $update(s, C', \phi)$. Let $I$ be an arbitrary $C'$-interpretation. We define

$$I \in update(s, C', \phi) :\Leftrightarrow I \models \phi \text{ and } \{I' \mid I' \models \phi, I' <_s I\} = \emptyset. \tag{2}$$

Hence, $update(s, C', \phi)$ is defined to be the set of all $C'$-interpretations which satisfy $\phi$, and which are minimal with respect to the partial order $<_s$. Put in different terms, $update(s, C', \phi)$ contains all interpretations that differ from $s$ in a set-inclusion minimal set of values.

Now, assume an action $a$. We say that $a$ is *applicable* in $s$, short $appl(s, a)$, if $s \models \text{pre}_a$, $C_a \subseteq C_s$, and $E_a \cap C_s = \emptyset$. That is, on top of the usual precondition satisfaction we require that $a$'s inputs exist and that $a$'s outputs do not yet exist. The result of executing $a$ in $s$ is:

$$res(s, a) := \begin{cases} \{(C', I') \mid C' = C_s \cup E_a, I' \in update(s, C', \Phi_{IC} \wedge \text{eff}_a)\} & appl(s, a) \\ \{s\} & \text{otherwise} \end{cases} \tag{3}$$

Note that $a$ can be executed in $s$ even if it is not applicable. In that case, the outcome is the singleton set containing $s$ itself, i.e., the action does not affect the state. This is an important aspect of our formalism, which we get back to below. If $\Phi_{IC} \wedge \text{eff}_a$ is unsatisfiable, then obviously we get $res(s, a) = \emptyset$. We say in this case that $a$ is *inconsistent*.[9]

The overall semantics of $\mathcal{WSC}$ tasks is now easily defined via a standard notion of *beliefs*. These model our uncertainty about the true state of the world. A belief $b$ is the set of world states that are possible at a given point in time. The *initial belief* is

$$b_0 := \{s \mid C_s = C_0, s \models \Phi_{IC} \wedge \phi_0\}. \tag{4}$$

An action $a$ is inconsistent with a belief $b$ if it is inconsistent with at least one $s \in b$. In the latter case, $res(b, a)$ is undefined. Otherwise, it is defined by

$$res(b, a) := \bigcup_{s \in b} res(s, a). \tag{5}$$

This is extended to action sequences in the obvious way. A *plan* is a sequence $\langle a_1, \ldots, a_n \rangle$ so that

$$\forall s \in res(b_0, \langle a_1, \ldots, a_n \rangle) : s \models \phi_G. \tag{6}$$

For illustration, consider the formalization of our example from Section 2.

**Example 2** *Reconsider Example 1. For the sake of conciseness, we formalize only a part of the example, with simplified axioms. The $\mathcal{WSC}$ task is defined as follows:*

---

9. Unless $\Phi_{IC}$ mentions any constants, if $a$ is based on operator $o$ and $a$ is inconsistent, then *any* action based on $o$ is inconsistent. Such operators can, in principle, be filtered out in a pre-process to planning.

- $\mathcal{P} = \{$ *protein, cellProtein, G, H, I, 1n55, 1kw3, InfoDSSP, Info3D, combinedPresentation* $\}$, *where all the predicates are unary.*

- $\Phi_{IC}$ *consists of the clauses:*

  - $\forall x : \neg$ *cellProtein* $(x) \lor$ *protein* $(x)$ – *[subsumption]*
  - $\forall x : \neg$ *protein* $(x) \lor G(x) \lor H(x) \lor I(x)$ – *[at least one DSSP value]*
  - $\forall x : \neg$ *protein* $(x) \lor 1n55(x) \lor 1kw3(x)$ – *[at least one 3-D shape]*
  - $\forall x : \neg$ *cellProtein* $(x) \lor G(x) \lor 1n55(x)$ – *[dependency]*
  - $\forall x : \neg$ *cellProtein* $(x) \lor H(x) \lor \neg 1n55(x)$ – *[dependency]*

- $\mathcal{O}$ *consists of the operators:*

  - **getInfoDSSP**$_G$: $(\{x\}, G(x), \{y\}, InfoDSSP(y))$
  - **getInfoDSSP**$_H$: $(\{x\}, H(x), \{y\}, InfoDSSP(y))$
  - **getInfoDSSP**$_I$: $(\{x\}, I(x), \{y\}, InfoDSSP(y))$
  - **getInfo3D**$_{1n55}$: $(\{x\}, 1n55(x), \{y\}, Info3D(y))$
  - **getInfo3D**$_{1kw3}$: $(\{x\}, 1kw3(x), \{y\}, Info3D(y))$
  - **combineInfo**: $(\{x_1, x_2\}, InfoDSSP(x_1) \land Info3D(x_2), \{y\}, combinedPresentation(y))$

- $C_0 = \{c\}$, $\phi_0 =$ *cellProtein* $(c)$

- $\phi_G = \exists x :$ *combinedPresentation* $(x)$

*To illustrate the formalism, we now consider a plan for this example task.*

*The initial belief $b_0$ consists of all states $s$ where $C_s = \{c\}$ and $s \models \Phi_{IC} \land$ cellProtein $(c)$. Say we apply the following sequence of actions:*

1. Apply **getInfoDSSP**$_G(c, d)$ *to $b_0$. Then we get to the belief $b_1$ which is the same as $b_0$ except that, from all $s \in b_0$ where $s \models G(c)$, new states are generated that have the constant $d$ and InfoDSSP $(d)$.*

2. Apply **getInfoDSSP**$_H(c, d)$ *to $b_1$. We get the belief $b_2$ where new states with $d$ and InfoDSSP $(d)$ are generated from all $s \in b_1$ where $s \models H(c)$.*

3. Apply **getInfo3D**$_{1n55}(c, e)$ *to $b_2$, yielding $b_3$.*

4. Apply **getInfo3D**$_{1kw3}(c, e)$ *to $b_3$. This yields $b_4$, where we get $e$ and Info3D $(e)$ from all $s \in b_2$ where $s \models 1n55(c)$ or $s \models 1kw3(c)$.*

5. Apply **combineInfo** $(d, e, f)$ *to $b_4$. This brings us to $b_5$ which is like $b_4$ except that from all $s \in b_4$ where $d, e \in C_s$ new states are generated that have $f$ and combinedPresentation $(f)$.*

*From the dependencies in $\Phi_{IC}$ (the last two clauses), we get that any $s \in b_0$ satisfies either $G(c)$ or $H(c)$. From the subsumption clause and the clause regarding 3-D shapes (first and third clauses) we get that any $s \in b_0$ satisfies either 1n55(c) or 1kw3(c). Hence, as is easy to verify, $b_5 \models \phi_G$ and so $\langle$**getInfoDSSP**$_G(c, d)$, **getInfoDSSP**$_H(c, d)$, **getInfo3D**$_{1n55}(c, e)$, **getInfo3D**$_{1kw3}(c, e)$, **combineInfo** $(d, e, f)\rangle$ is a plan.*

*Note that this plan does not make use of* **getInfoDSSP**$_I(c, d)$. *To obtain a plan, in this domain one can always just apply* all *information services. However, this plan is trivial and does not take into account what is relevant and what is not. Reasoning over* $\Phi_{IC}$ *enables us to find better plans.*

Our semantics for executing non-applicable actions is vital for the workings of Example 2. As pointed out above, below the definition of $res(s, a)$ (Equation (3)), $a$ can be executed in $s$ even if it is not applicable. This realizes partial matches: a web service can be called as soon as it might match one of the possible situations. In planning terms, our actions have a *conditional effects* semantics.[10] The contrasting notion would be to enforce preconditions, i.e., to say that $res(s, a)$ is undefined if $a$ is not applicable to $s$. This would correspond to plug-in matches.

In Example 2, the partial match semantics is necessary in order to be able to apply actions that cover only particular cases. For example, consider the action **getInfoDSSP**$_G(c, d)$, which is applied to the initial belief in the example plan. The precondition of that action is $G(c)$. However, there are states in the initial belief which do not satisfy that precondition. The initial belief allows any interpretation satisfying $\Phi_{IC} \wedge \phi_0$ (cf. Equation (4)), and some of these interpretations satisfy $H(c)$ rather than $G(c)$. Due to the partial match semantics, **getInfoDSSP**$_G(c, d)$ does not affect such states – its match with the initial belief is partial.

Clarification is also in order regarding our understanding of constants. First, like every PDDL-like planning formalism (we are aware of), we make a unique name assumption, i.e., different constants refer to different objects. Second, our understanding of web services is that any output they create is a separate individual, i.e., a separate information object.

The latter directly raises the question why we allow actions to share output constants. The answer is that we allow the planner to treat two objects *as if* they were the same. This makes sense if the two objects play the same role in the plan. Consider again Example 2. The actions **getInfoDSSP**$_G(c, d)$ and **getInfoDSSP**$_H(c, d)$ share the same output constant, $d$. This means that $d$ is one name for two separate information objects. These two objects have the same properties, derived from InfoDSSP($d$). The only difference between them is that they are created in different cases, namely from states that satisfy $G(c)$ and $H(c)$ respectively. Having a single name for the two objects is useful because we can take that name as a parameter of actions that do not need to distinguish between the different cases. In the example, **combineInfo**$(d, e, f)$ is such an action.

As hinted, the "cases" in the above correspond to different classes of concrete execution traces. Importantly, on any particular execution trace, each output constant is created at most once. To see this, consider an execution trace $s_0, a_0, s_1, a_1, \ldots, a_k, s_{k+1}$, i.e., an alternating sequence of states and actions where $s_0 \in b_0$, and $s_{i+1} = res(s_i, a_i)$ for all $0 \leq i \leq k$. Say that $a_i$ and $a_j$ share an output constant, $d$. Say further that $a_i$ is applicable in $s_i$, and hence $d \in C_{s_{i+1}}$. Then, quite obviously, we have $d \in C_{s_l}$ for all $i + 1 \leq l \leq k + 1$. In particular, $a_j$ is not applicable in $s_j$: the intersection of its output constants with $C_{s_j}$ is non-empty (cf. the definition of $appl(s, a)$). So, due to our definition of action applicability, it can never happen that the same constant is created twice. In other words, there can never be a reachable state where a single constant name refers to more than one individual information object. In that sense, the use of one name for several objects occurs only at planning time, when the actual execution case which will occur – the actual case which will occur – is not known. For illustration, consider **getInfoDSSP**$_G(c, d)$ and **getInfoDSSP**$_H(c, d)$, and their shared output $d$, in Example 2. Even if the concrete state $s_0 \in b_0$ in which the execution starts satisfies both $G(c)$ and $H(c)$, only one of the actions will fire – namely the one that comes first.

---

10. An obvious generalization is to allow several conditional effects per action, in the style of the ADL language (Pednault, 1989). We omit this here for the sake of simplifying the discussion. An extension in this direction is straight-forward.

We remark that we initially experimented with a definition where actions instantiate only their inputs, and when they are applied to a state $s$ their outputs are, by virtue of the execution semantics, instantiated to constants outside of $C_s$. In such a framework, one can never choose to "share output constants", i.e., to use the same name for two different outputs. The notion we have settled for is strictly richer: the planner can always choose to instantiate the outputs with constants outside of $C_s$. The question is, when does it make sense to share outputs? Answering this question in a domain-independent planner may turn out to be quite non-trivial. We get back to this when we discuss a possible adaptation of CFF in Section 4.5. In the experiments reported herein (Section 6), we use a simple heuristic. Outputs are shared iff the operator effects are identical (giving an indication that the respective outputs may indeed "play the same role" in the plan).

We conclude this sub-section with a final interesting observation regarding modeling in our framework. Negative effects are not an essential part of the $\mathcal{WSC}$ formalism: they can be compiled away. We simply replace any negative effect $\neg G(x_1, \ldots, x_k)$ with $notG(x_1, \ldots, x_k)$ (introducing a new predicate) and state in the integrity constraints that the two are equivalent. That is, we introduce the two new clauses $\forall x_1, \ldots, x_k : G(x_1, \ldots, x_k) \vee notG(x_1, \ldots, x_k)$ and $\forall x_1, \ldots, x_k : \neg G(x_1, \ldots, x_k) \vee \neg notG(x_1, \ldots, x_k)$. While this is a simple compilation technique, the formal details are a little intricate, and are moved to Appendix A. If $a$ is an action in the original task, then $a^+$ denotes the corresponding action in the compiled task, and vice versa. Similarly, if $s$ is an action in the original task, then $s^+$ denotes the corresponding state in the compiled task. We get:

**Proposition 1 (Compilation of Negative Effects in $\mathcal{WSC}$)** *Assume a $\mathcal{WSC}$ task $(\mathcal{P}, \Phi_{IC}, \mathcal{O}, C_0, \phi_0, \phi_G)$. Let $(\mathcal{P}^+, \Phi_{IC}^{'+}, \mathcal{O}^+, C_0, \phi_0, \phi_G)$ be the same task but with negative effects compiled away. Assume an action sequence $\langle a_1, \ldots, a_n \rangle$. Let $b$ be the result of executing $\langle a_1, \ldots, a_n \rangle$ in $(\mathcal{P}, \Phi_{IC}, \mathcal{O}, C_0, \phi_0, \phi_G)$, and let $b^+$ be the result of executing $\langle a_1^+, \ldots, a_n^+ \rangle$ in $(\mathcal{P}^+, \Phi_{IC}^+, \mathcal{O}^+, C_0, \phi_0, \phi_G)$. Then, for any state $s$, we have that $s \in b$ iff $s^+ \in b^+$.*

This can be proved by straightforward application of the relevant definitions. The most important aspect of the result is that the new clauses introduced are allowed in the forward effects and strictly forward effects special cases identified later. Hence, any hardness results transfer directly to tasks without negative effects and dropping negative effects cannot make the algorithms any easier.

### 3.3 Computational Properties

We now perform a brief complexity analysis of the $\mathcal{WSC}$ formalism in its most general form as introduced above. In line with many related works of this kind (Eiter & Gottlob, 1992; Bylander, 1994; Liberatore, 2000; Eiter et al., 2004), we consider the *propositional case*. In our context, this means that we assume a fixed upper bound on the arity of predicates, on the number of input/output parameters of each operator, on the number of variables appearing in the goal, and on the number of variables in any clause. We will refer to $\mathcal{WSC}$ tasks restricted in this way as $\mathcal{WSC}$ tasks *with fixed arity*.

We consider the problems of *checking plans* – testing whether or not a given action sequence is a plan – and of deciding plan existence. For the latter, we distinguish between polynomially bounded plan existence, and unbounded plan existence. We deem these to be particularly relevant decision problems in the context of plan generation. Certainly, plan checks are an integral part of plan gen-eration. Indeed, if a planning tool is based on state space search, then the tool either performs such checks explicitly for (potentially many) plan candidates generated during search, or this complexity is inherent in the effort that underlies the computation of state transitions. Polynomially bounded plan existence is relevant because, in most commonly used planning benchmark domains, plans are of polynomial length (it is also a very wide-spread intuition in the SWS community that composed web services will not contain exceedingly large numbers of web services). Finally, unbounded plan existence is the most general decision problem involved, and thus is of generic interest.

All the problems turn out to be very hard. To prove this, we reuse and adapt various results from the literature. We start with the complexity of plan checking, for which hardness follows from a long established result (Eiter & Gottlob, 1992) regarding the complexity of belief update. For all the results, detailed proofs are available in Appendix A.

**Theorem 1 (Plan Checking in $\mathcal{WSC}$)** *Assume a $\mathcal{WSC}$ task with fixed arity, and a sequence $\langle a_1, \ldots, a_n \rangle$ of actions. It is $\Pi_2^p$-complete to decide whether $\langle a_1, \ldots, a_n \rangle$ is a plan.*

**Proof Sketch:** Membership can be shown by a guess-and-check argument. Guess the proposition values along $\langle a_1, \ldots, a_n \rangle$. Then check whether these values comply with $res$, and lead to an inconsistent action, or to a final state that does not satisfy the goal. $\langle a_1, \ldots, a_n \rangle$ is a plan iff this is not the case for any guess of proposition values. Checking goal satisfaction is polynomial, checking compliance with $res$ is in coNP, checking consistency is in NP.

Hardness follows by a simple adaptation of the proof of Lemma 6.2 from Eiter and Gottlob (1992). That proof uses a reduction from checking validity of a QBF formula $\forall X.\exists Y.\psi[X, Y]$. The lemma considers the case where a propositional belief $\Phi$ is updated with an arbitrary (propositional) formula $\phi$, and the decision problem is to ask whether some other formula $\Phi'$ is implied by the updated belief. In the proof, $\Phi$ is a complete conjunction of literals, i.e., $\Phi$ corresponds to a single world state. $\Phi'$ is a single propositional fact $r$ which is true in $\Phi$. The semantics of $\forall X.\exists Y.\psi[X, Y]$ are encoded in a complicated construction defining the update $\phi$. In a nutshell, $\phi$ is a CNF telling us that for every assignment to $X$ (which will yield a world state $s'$ in the updated belief), we either have to find an assignment to $Y$ so that $\psi[X, Y]$ holds ("completing" $s'$), or we have to falsify $r$.

The difference in our setting lies in our very restricted "update formulas" – action effects – and in the fact that the integrity constraints are supposed to hold in *every* belief. We adapt the above proof by, first, taking the integrity constraints to be the clauses in Eiter and Gottlob's CNF formula $\phi$. We then modify the constraints so that they need only be true if a new fact $t$ holds – i.e., we insert $\neg t$ into every clause. The initial belief has $t$ false, and otherwise corresponds exactly to $\Phi$ as above. The only action of the plan makes $t$ true. The goal is Eiter and Gottlob's fact $r$. □

We remark that membership in Theorem 1 remains valid when allowing actions with multiple conditional effects, when allowing parallel actions, and even when allowing their combination. On the other hand, by virtue of the proof argument as outlined, hardness holds even if the initial state literals $\phi_0$ are complete (describe a single world state), the plan consists of a single world action with a single positive effect literal, and the goal is a single propositional fact that is initially true.

We next consider polynomially bounded plan existence. For this, membership follows directly from Theorem 1. To prove hardness, we construct a planning task that extends Eiter and Gottlob's construction from above with actions that allow to choose a valuation for a third, existentially quantified, set of variables, and hence reduces validity checking of a QBF formula $\exists X.\forall Y.\exists Z.\psi[X, Y, Z]$.

**Theorem 2 (Polynomially Bounded Plan Existence in $\mathcal{WSC}$)** *Assume a $\mathcal{WSC}$ task with fixed arity, and a natural number $b$ in unary representation. It is $\Sigma_3^p$-complete to decide whether there exists a plan of length at most $b$.*

**Proof:** For membership, guess a sequence of at most $b$ actions. By Theorem 1, we can check with a $\Pi_2^p$ oracle whether the sequence is a plan.

For hardness, validity of a QBF formula $\exists X.\forall Y.\exists Z.\psi[X, Y, Z]$, where $\psi$ is in CNF, is reduced to testing plan existence. Say $X = \{x_1, \ldots, x_n\}$. In the planning task, there are $n$ actions (operators with empty input/output parameters) $o^{x_i}$ and $o^{\neg x_i}$ of which the former sets $x_i$ to true and the latter sets $x_i$ to false. Further, there is an action $o^t$ which corresponds to the action used in the hardness proof of Theorem 1. The actions are equipped with preconditions and effects ensuring that any plan must first apply, for all $1 \leq i \leq n$, either $o^{x_i}$ or $o^{\neg x_i}$, and thereafter must apply $o^t$ (of course enforcing the latter also requires a new goal fact that can be achieved only by $o^t$). Hence, choosing a plan candidate in this task is the same as choosing a value assignment $a_X$ for the variables $X$.

In our construction, after all the $o^{x_i}$ and $o^{\neg x_i}$ actions have been executed, one ends up in a belief that contains a single world state, where the value assignment $a_X$ for the variables $X$ corresponds to the chosen actions. This world state basically corresponds to the belief $\Phi$ as in the hardness proof of Theorem 1. The only difference is that the construction has been extended to cater for the third set of variables. This is straightforward. Then, the belief that results from executing $o^t$ satisfies the goal iff Eiter and Gottlob's fact $r$ holds in all its world states. By virtue of similar arguments to those of Eiter and Gottlob, the latter is the case iff $\forall Y.\exists Z.\psi[a_X/X, Y, Z]$, i.e., the substitution of $\exists X.\forall Y.\exists Z.\psi[X, Y, Z]$ with $a_X$, is valid. From this, the claim follows. □

Our final result regards unbounded plan existence in $\mathcal{WSC}$. The result is relatively easy to obtain from the generic reduction described by Bylander (1994) to prove PSPACE-hardness of plan existence in STRIPS. Somewhat shockingly, it turns out that plan existence in $\mathcal{WSC}$ is undecidable even without any integrity constraints, and with a complete initial state description. The source of undecidability is, of course, the ability to generate new constants on-the-fly.

**Theorem 3 (Unbounded Plan Existence in $\mathcal{WSC}$)** *Assume a $\mathcal{WSC}$ task. The decision problem asking whether a plan exists is undecidable.*

**Proof Sketch:** By a modification of the proof by Bylander (1994) that plan existence in propositional STRIPS planning is PSPACE-hard. The original proof proceeds by a generic reduction, constructing a STRIPS task for a Turing Machine with polynomially bounded space. The latter restriction is necessary to model the machine's tape: tape cells are pre-created for all positions within the bound. Exploiting the ability to create constants on-the-fly, we can instead introduce simple operators that allow to extend the tape, at both ends. □

Not being able to decide plan existence is, of course, a significant limitation in principle. However, this limitation is probably of marginal importance in practice, because most planning tools just assume that there is a plan, and they try to find it – rather than trying to prove that there is no plan. In that sense, most planning tools are, by their nature, semi-decision procedures anyway. What matters more than decidability in such a setting is the question whether one can find a plan quickly enough, i.e., before exhausting time or memory.[11] This is also the most relevant question in web service composition.

## 4. Forward Effects

The high complexity of planning in $\mathcal{WSC}$ motivates the search for interesting special cases. We define a special case, called *forward effects*, where every change an action makes to the state involves a newly generated constant.

We start the section by defining the forward effects case and making a core observation about its semantics. We then discuss the modeling power of this special case. Next, we discuss forward effects from a more general perspective of belief update. We analyze the main computational properties of forward effects, and we conclude the section with an assessment of how an existing planning tool could be adapted to handle forward effects.

### 4.1 $\mathcal{WSC}|_{fwd}$ and its Semantics

The forward effects special case of $\mathcal{WSC}$ is defined as follows.

**Definition 1** *Assume a $\mathcal{WSC}$ task $(\mathcal{P}, \Phi_{IC}, \mathcal{O}, C_0, \phi_0, \phi_G)$. The task has* forward effects *iff:*

1. *For all $o \in \mathcal{O}$, and for all $l[X] \in \mathit{eff}_o$, we have $X \cap Y_o \neq \emptyset$.*

2. *For all clauses $\phi \in \Phi_{IC}$, where $\phi = \forall x_1, \ldots, x_k : l_1[X_1] \vee \cdots \vee l_n[X_n]$, we have $X_1 = \cdots = X_n$.*

*The set of all $\mathcal{WSC}$ tasks with forward effects is denoted with $\mathcal{WSC}|_{fwd}$.*

The first condition says that the variables of every effect literal contain at least one output variable. This implies that every ground effect literal of an action contains at least one new constant. The second condition says that, within every integrity constraint, all literals share the same arguments. This implies that effects involving new constants can only affect literals involving new constants. Note that, since $x_1, \ldots, x_k$ are by definition exactly the variables occurring in any of the literals, for each $X_i$ we have $X_i = x_1, \ldots, x_k$. Note further that we may have $k = 0$, i.e., the literals in the clause may be ground. This is intentional. The constants mentioned in the clause must be taken from $C_0$, cf. the discussion in Section 3.1. Therefore, such clauses have no interaction with statements about the new constants generated by a $\mathcal{WSC}|_{fwd}$ action.

We will discuss the modeling power of $\mathcal{WSC}|_{fwd}$ below (Section 4.2). First, we observe that the semantics of $\mathcal{WSC}|_{fwd}$ is much simpler than that of general $\mathcal{WSC}$. One no longer needs the notion of minimal change with respect to the previous state. To state this more precisely, assume a $\mathcal{WSC}$ task with predicates $\mathcal{P}$. Say $I'$ is an interpretation over $\mathcal{P}^{C'}$, where $C'$ is a set of constants. Say that $C \subseteq C'$. We denote by $I'|_C$ the restriction of $I'$ to $\mathcal{P}^C$, i.e., the interpretation of $\mathcal{P}^C$ that coincides with $I'$ on all these propositions. Given a state $s$ and an action $a$, we define:

$$res|_{fwd}(s,a) := \left\{ \begin{array}{ll} \{(C',I') \mid C' = C_s \cup E_a, I'|_{C_s} = I_s, I' \models \Phi_{IC} \wedge \mathit{eff}_a\} & appl(s,a) \\ \{s\} & \text{otherwise} \end{array} \right. \quad (7)$$

---

11. Indeed the planning community is generally rather unconcerned by undecidability, cf. the numeric track of the international planning competitions, and Helmert's (2002) results on the decidability of numerical planning problems.

Compare this to Equation (3), where $I'$ is defined to be a member of $update(s, C', \Phi_{IC} \wedge \text{eff}_a)$, which returns all interpretations that satisfy $\Phi_{IC} \wedge \text{eff}_a$ and that differ minimally from $I_s$. In Equation (7), $I'$ is simply set to be *identical* to $I_s$, on the constants (on the propositions over the constants) that existed beforehand. In other words, the set of new states we get is the cross-product of the old state with all satisfying assignments to $\Phi_{IC} \wedge \text{eff}_a$.

**Lemma 1 (Semantics of $\mathcal{WSC}|_{fwd}$)** *Assume a $\mathcal{WSC}|_{fwd}$ task, a reachable state $s$, and an action $a$. Then $res(s, a) = res|_{fwd}(s, a)$.*

**Proof Sketch:** In $\mathcal{WSC}|_{fwd}$, if $s'$ differs minimally from $s$, then it follows that $s'$ agrees totally with $s$, on the set of propositions $P^{C_s}$ interpreted by $s$. To see this, denote as before with $P^{C_s + E_a}$ the set of all propositions with arguments in $C_s \cup E_a$, and with at least one argument in $E_a$, and denote with $\Phi_{IC}[C_s + E_a]$ the instantiation of $\Phi_{IC}$ with all constants from $C_s \cup E_a$, where in each clause at least one variable is instantiated from $E_a$. The key argument is that $s' \models \Phi_{IC} \wedge \text{eff}_a$ is equivalent to $s' \models \Phi_{IC}[C_s \cup E_a] \wedge \text{eff}_a$, which in turn is equivalent to $s' \models \Phi_{IC}[C_s] \wedge \Phi_{IC}[C_s + E_a] \wedge \text{eff}_a$. In the last formula, $\Phi_{IC}[C_s]$ only uses the propositions $P^{C_s}$, whereas $\Phi_{IC}[C_s + E_a] \wedge \text{eff}_a$ only uses the propositions $P^{C_s + E_a}$. Since $s$ is reachable, we have $s \models \Phi_{IC}[C_s]$. Therefore, to satisfy $\Phi_{IC} \wedge \text{eff}_a$, there is no need to change any of the values assigned by $s$. □

## 4.2 Modeling Power

Intuitively, $\mathcal{WSC}|_{fwd}$ covers the situation where a web service outputs some new constants, sets their characteristic properties relative to the inputs, and relies on the ontology axioms to describe any ramifications concerning the new constants. As was detailed in Section 2, this closely corresponds to the various notions of message-based WSC explored in the literature. In that sense, the modeling power of $\mathcal{WSC}|_{fwd}$ is comparable to that of message-based WSC, one of the most-widespread approaches in the area.

A simple concrete way of assessing the modeling power of $\mathcal{WSC}|_{fwd}$ is to consider the allowed and disallowed axioms. Examples of axioms that are not allowed by $\mathcal{WSC}|_{fwd}$ are: attribute domain restrictions, taking the form $\forall x, y : \neg G(x, y) \vee H(x)$; attribute range restrictions, taking the form $\forall x, y : \neg G(x, y) \vee H(y)$; and relation transitivity, taking the form $\forall x, y, z : \neg G(x, y) \vee \neg G(y, z) \vee G(x, z)$. Note that, for all these axioms, it is easy to construct a case where an action effect, even though it involves a new constant, affects the "old belief". For example, if constants $c$ and $e$ existed beforehand, and an action outputs $d$ and sets $G(c, d) \wedge G(d, e)$, then the axiom $\forall x, y : \neg G(x, y) \vee \neg G(y, z) \vee G(x, z)$ infers that $G(c, e)$ – a statement that does *not* involve the new constant $d$.

Typical ontology axioms that *are* allowed by $\mathcal{WSC}|_{fwd}$ are: subsumption relations, taking the form $\forall x : \neg G(x) \vee H(y)$; mutual exclusion, taking the form $\forall x : \neg G(x) \vee \neg H(y)$; relation reflexivity, taking the form $\forall x : \neg G(x, x)$; and relation symmetry, taking the form $\forall x, y : \neg G(x, y) \vee G(y, x)$. We can also express that a concept $G$ is contained in the union of concepts $H_1, \ldots, H_n$, and more generally we can express any complex dependencies between concepts, taking the form of clausal constraints on the allowed combinations of concept memberships.

One example where complex dependencies are important is the domain of proteins as illustrated in Example 1. Capturing the dependencies is important here in order to be able to select the correct web services. Similar situations arise in many domains that involve complex interdependencies and/or complex regulations. An example for the latter is the Virtual Travel Agency which we discussed before. For example, in the German rail system there are all kinds of regulations regarding which train may be booked with which kind of discount under which conditions. Modeling these regulations would enable a WSC algorithm to select the appropriate booking services. Another interesting case is the hospital domain described by de Jonge, van der Linden, and Willems (2007). There, the problem of hospital asset tracking is handled by means of a set of tracking, logging and filter services, which transform logs to extract various kinds of information. In this setting, it would make sense to model complex dependencies so that the web service composer may determine *which* hospital assets need to be tracked and retrieved. Namely, the latter depends on the type of operation in question, and on the kind of examinations which that operation requires. Accordingly, what we need to model is a categorization of operations, their mapping to sets of required examinations, and how those examinations are associated with hospital assets. Further complications arise since the required examinations/assets may depend on particular circumstances. Clearly, we can express the categorization and dependencies in terms of clauses. While this of course captures only a fraction of what is relevant in a hospital, it is considerably more informed than a composer which always just tracks all the assets.

The main weakness of $\mathcal{WSC}|_{fwd}$ is that it does not allow us to express changes regarding pre-existing objects. This is best illustrated when considering the case of negative effects.[12] In the planning community, these are commonly used to model how previous properties of objects are invalidated by an action. For illustration, reconsider Example 1. Say there is an additional operator **dropCoffeeIn3Dmachine**, with effect $\neg\text{Info3D}(y)$. One would normally expect that, when this operator is applied, the fact $\text{Info3D}(y)$ is deleted and must be re-established. This is not so in $\mathcal{WSC}|_{fwd}$. According to the restrictions this special case imposes, the variable $y$ in $\neg\text{Info3D}(y)$ must be an *output* of **dropCoffeeIn3Dmachine**. That is, dropping coffee into the machine creates a new object, whose characteristic property happens to be $\neg\text{Info3D}(y)$ rather than $\text{Info3D}(y)$. Clearly, this is not the intended semantics of the operator.

To model the intended semantics, we would need to instantiate $y$ with a pre-existing constant. Say that, as in belief $b_3$ in Example 1, a constant $e$ with $\text{Info3D}(e)$ was previously created by **getInfo3D**$_{1n55}(c, e)$. Then $\mathcal{WSC}|_{fwd}$ does allow us to instantiate **dropCoffeeIn3Dmachine** with $e$, so that we have the effect $\neg\text{Info3D}(e)$. However, by virtue of the definition of action applicability, that action will be applicable only in states where $e$ does not yet exist – corresponding to execution paths where **getInfo3D**$_{1n55}(c, e)$ was not executed. Hence the property $\text{Info3D}(e)$ does not get deleted from any state, and $e$ as used by **dropCoffeeIn3Dmachine** is still regarded as a newly created object whose characteristic property is $\neg\text{Info3D}(y)$. The only difference the new action makes is that, now, the plan uses the same name ($e$) to refer to two different information objects (output of **getInfo3D**$_{1n55}(c, e)$ vs. output of **dropCoffeeIn3Dmachine**) that do *not* play the same role in the plan, cf. the discussion in Section 3.2.

An interesting workaround is to let the operators output "time steps", in a spirit reminiscent of the situation calculus (McCarthy & Hayes, 1969; Reiter, 1991). Every operator obtains an extra output variable $t$, which is included into every effect literal. The new time step $t$ is stated to stand in some relation to the previous time steps, e.g., $next(tprev, t)$ where $tprev$ is an input variable instantiated to the previous time step. In such a setting, we can state how the world changes over time. In particular we can state that some object property is different in $t$ than in $tprev$. For example, if an action moves a file $f$ from "RAEDME" to "README" then we could state that $name(f, \text{"RAEDME"}, tprev)$ and $name(f, \text{"README"}, t)$. The problem with such a construction is that the time steps have no special interpretation, they are just ordinary objects.[13] This causes at least two difficulties. (1) If we want to refer to an object property, we have to know the time step in the first place – that is, we have to know whether the actual time step is $t$ or $tprev$. Note here that we cannot maintain a predicate $actualTime(x)$ because this would require us to invalidate a property of $tprev$. (2) There is no solution to the frame problem. The operators must explicitly state every relevant property of the previous time step, and how each property is changed in the new time step.[14]

---

12. Or, in $\mathcal{WSC}$, positive effects triggering negative effects via $\Phi_{IC}$, cf. Proposition 1.

To conclude this sub-section, let us consider how $\mathcal{WSC}|_{fwd}$ can be generalized without losing Lemma 1. Most importantly, instead of requiring that *every* effect literal involves a new constant, one can postulate this only for literals that may actually be affected by the integrity constraints. In particular, if a predicate does not appear in any of the clauses, then certainly an effect literal on that predicate is not harmful even if it does not involve an output constant. One obtains a potentially stronger notion by considering ground literals, rather than predicates. Note that this kind of generalization solves difficulty (1) of the time-step construction, presuming that time steps are not constrained by the clauses. (The frame problem, however, persists.)

Another possibility, deviating somewhat from the way $\mathcal{WSC}$ and $\mathcal{WSC}|_{fwd}$ are currently defined, is to define the integrity constraints in terms of logic programming style rules, along the lines of Eiter et al. (2003, 2004). The requirement on $\mathcal{WSC}|_{fwd}$ can then be relaxed to postulate that the effect literals without new constants do not appear in the rule heads.

We remark that the latter observation suggests a certain strategic similarity with the aforementioned *derived predicates* (Thiébaux et al., 2005) previously used in AI Planning to manage the complexity of integrity constraints. There, the integrity constraints take the form of stratified logic programming style derivation rules, and the predicates appearing in rule heads are not allowed to appear in operator effects. This is an overly restricted solution, in the WSC context. The effects of web services are indeed very likely to affect concepts and relations appearing in the ontology axioms. They may do so in $\mathcal{WSC}|_{fwd}$, as long as output constants are involved.

### 4.3 Belief Update

Lemma 1 is specific to the possible models approach (Winslett, 1988) that underlies our semantics of action applications. It is interesting to consider the semantics of $\mathcal{WSC}|_{fwd}$ from a more general perspective of belief update. Recall that such an update involves a formula characterizing the current belief, and a formula describing the update. We seek a formula that characterizes the updated belief.

A wide variety of definitions has been proposed as to how the updated belief should be defined. However, some common ground exists. Katzuno and Mendelzon (1991) suggest eight postulates, named (U1)...(U8), which every sensible belief update operation should satisfy. Herzig and Rifi (1999) discuss in detail to what degree the postulates are satisfied by a wide range of alternative belief update operators. In particular they call a postulate "uncontroversial" if all update operators under investigation satisfy them. We will take up these results in the following. We examine to what extent we can draw conclusions about the updated belief, $\Phi'$, in the setting of the forward effects case, when relying only on Herzig and Rifi's "uncontroversial postulates".

---

13. Note that here the similarity to the situation calculus ends. Whereas time steps are assigned a specific role in the formulas used in the situation calculus, here they are just ordinary objects handled by actions, as if they were packages or blocks.

14. Despite these difficulties, Theorem 6 below shows that a time step construction can be used to simulate an Abacus machine, and hence to prove undecidability of plan existence in $\mathcal{WSC}|_{fwd}$.

We assume that a planning task with predicates $\mathcal{P}$ is given. We need the following notations:

- If $\Phi$ and $\phi$ are formulas, then $\Phi \circ \phi$ denotes the formula that results from updating the belief $\Phi$ with the update $\phi$, under some semantics for the belief update operator $\circ$.

- Given disjoint sets of constants $C$ and $E$, $\mathcal{P}^{C+E}$ denotes the set of all propositions formed from predicates in $\mathcal{P}$, where all arguments are contained in $C \cup E$ and there exists at least one argument contained in $E$. (Recall that $\mathcal{P}^C$ denotes the set of all propositions formed from predicates in $\mathcal{P}$ and arguments from $C$.)

- Given a set of constants $C$, $\Phi_{IC}[C]$ denotes the instantiation of $\Phi_{IC}$ with $C$. That is, $\Phi_{IC}[C]$ is the conjunction of all clauses that result from replacing the variables of a clause $\phi \in \Phi_{IC}$, $\phi = \forall x_1, \ldots, x_k : l_1[X_1] \vee \cdots \vee l_n[X_n]$, with a tuple $(c_1, \ldots, c_k)$ of constants in $C$.

- Given disjoint sets of constants $C$ and $E$, $\Phi_{IC}[C + E]$ is the conjunction of all clauses that result from replacing the variables of a clause $\phi \in \Phi_{IC}$, $\phi = \forall x_1, \ldots, x_k : l_1[X_1] \vee \cdots \vee l_n[X_n]$, with a tuple $(c_1, \ldots, c_k)$ of constants in $C \cup E$, where at least one constant is taken from $E$.[15]

- If $\psi$ is a ground formula then by $P(\psi)$ we denote the set of propositions occurring in $\psi$.

We will denote the current belief by $\Phi$ and the update by $\phi$. As another convention, given a set of constants $C$, by writing $\psi^C$ we indicate that $P(\psi) \subseteq \mathcal{P}^C$. Similarly, given disjoint sets of constants $C$ and $E$, by writing $\psi^{C+E}$ we indicate that $P(\psi) \subseteq \mathcal{P}^{C+E}$. If $s$ is a state, then by $\psi_s$ we denote the conjunction of literals satisfied by $s$.

We first consider the case where, similar to the claim of Lemma 1, $\Phi$ corresponds to a single concrete world state $s$. We want to apply an action $a$. We wish to characterize the set of states $res(s, a)$, i.e., we wish to construct the formula $\Phi \circ \phi$. For simplicity of notation, denote $C := C_s$ and $E := E_a$. If $a$ is not applicable to $s$, there is nothing to do. Otherwise, we have that:

(I) $\Phi \equiv \Phi_{IC}[C] \wedge \psi^C$ where $P(\psi^C) \subseteq \mathcal{P}^C$.

For example, we can set $\psi^C := \psi_s$. Since $s \models \Phi_{IC}$, we get the desired equivalence. Further, we have that:

(IIa) $\phi \equiv \Phi_{IC}[C] \wedge \Phi_{IC}[C + E] \wedge \text{eff}_a$;

(IIb) $P(\Phi_{IC}[C + E]) \subseteq \mathcal{P}^{C+E}$ and $P(\text{eff}_a) \subseteq \mathcal{P}^{C+E}$.

(IIa) holds trivially: $\phi$ is defined as $\Phi_{IC} \wedge \text{eff}_a$, which is equivalent to $\Phi_{IC}[C \cup E] \wedge \text{eff}_a$ which is equivalent to $\Phi_{IC}[C] \wedge \Phi_{IC}[C + E] \wedge \text{eff}_a$. As for (IIb), this is a consequence of the forward effects case. Every effect literal contains at least one output constant, hence $\text{eff}_a$ contains only propositions from $\mathcal{P}^{C+E}$. For $\Phi_{IC}[C + E]$, we have that at least one variable in each clause is instantiated with a constant $e \in E$. Since, by definition, all literals in the clause share the same variables, $e$ appears in every literal and therefore $\Phi_{IC}[C + E]$ contains only propositions from $\mathcal{P}^{C+E}$.

As an illustration, consider our simple VTA example. There are four predicates, $train(x)$, $ticket(x)$, $trainTicket(x)$, and $ticketFor(x, y)$. The set of integrity constraints $\Phi_{IC}$ consists of the single axiom $\forall x : trainTicket(x) \Rightarrow ticket(x)$. In our current state $s$, we have $C_s = \{c\}$, and $I_s$ sets all propositions to 0 except for $train(c)$. We consider the application of the action $a = \textbf{bookTicket}(c, d)$, whose precondition is $train(c)$, whose set $E$ of output constants is $\{d\}$, and whose effect $\text{eff}_a$ is $trainTicket(d) \land ticketFor(d, c)$. In this setting, we have: $\Phi_{IC}[C] = (trainTicket(c) \Rightarrow ticket(c))$; $\psi^C = (train(c) \land \neg ticket(c) \land \neg trainTicket(c) \land \neg ticketFor(c, c))$; and $\Phi_{IC}[C + E] = (trainTicket(d) \Rightarrow ticket(d))$.

---

15. If no clause in $\Phi_{IC}$ contains any variable, then $\Phi_{IC}[C + E]$ is empty. As is customary, an empty conjunction is taken to be true, i.e., 1.

We will derive in the following that:

(III) $\Phi \circ \phi \equiv (\Phi_{IC}[C] \land \psi^C) \land (\Phi_{IC}[C + E] \land \text{eff}_a)$.

That is, we can characterize the updated belief simply by the conjunction of the previous belief with the action effect and the extended instantiation of the ontology axioms. This corresponds exactly to Lemma 1. To illustrate, we will continue the VTA example. The left hand side of (III) refers to the four propositions based only on $c$, and sets them according to $s$. The right hand side refers to propositions based only on $d$ – $trainTicket(d)$ and $ticket(d)$ – as well as the proposition $ticketFor(d, c)$ which links $c$ and $d$.

As one prerequisite of our derivation of (III), we have to make an assumption which, to the best of our knowledge, is not discussed anywhere in the belief update literature:

(IV) Let $\psi_1, \psi_1', \psi_2, \psi_2'$ be formulas where $P(\psi_1) \cap P(\psi_1') = \emptyset$, $P(\psi_1) \cap P(\psi_2') = \emptyset$, $P(\psi_2) \cap P(\psi_1') = \emptyset$, and $P(\psi_2) \cap P(\psi_2') = \emptyset$. Then $(\psi_1 \land \psi_1') \circ (\psi_2 \land \psi_2') \equiv (\psi_1 \circ \psi_2) \land (\psi_1' \circ \psi_2')$.

This assumption postulates that formulas talking about disjoint sets of variables can be updated separately. Since formulas with disjoint variables essentially speak about different aspects of the world, this seems a reasonable assumption.

Now, we start from the formula $\Phi \circ \phi$. We make replacements according to (I) and (IIa), leading to the equivalent formula $(\Phi_{IC}[C] \land \psi^C) \circ (\Phi_{IC}[C] \land \Phi_{IC}[C + E] \land \text{eff}_a)$. We can map this formula onto (IV) by taking $\psi_1$ to be $\Phi_{IC}[C] \land \psi^C$, $\psi_1'$ to be 1, $\psi_2$ to be $\Phi_{IC}[C]$, and $\psi_2'$ to be $\Phi_{IC}[C + E] \land \text{eff}_a$. Hence, we can separate our update into two parts as follows:

(A) $(\Phi \circ \phi)^C := (\Phi_{IC}[C] \land \psi^C) \circ \Phi_{IC}[C]$

(B) $(\Phi \circ \phi)^{C+E} := 1 \circ (\Phi_{IC}[C + E] \land \text{eff}_a)$

According to (IV), we then obtain our desired formula $\Phi \circ \phi$ by $\Phi \circ \phi \equiv (\Phi \circ \phi)^C \land (\Phi \circ \phi)^{C+E}$.

Illustrating this with the VTA example, we simply separate the parts of the update that talk only about $c$ from those that talk only about $d$ or the combination of both constants. The (A) part of the update is $trainTicket(c) \Rightarrow ticket(c)$ conjoined with $\psi_s$, updated with $trainTicket(c) \Rightarrow ticket(c)$. The (B) part of the update is 1 – representing the (empty) statement that the previous state $s$ makes about $d$ – updated with $(trainTicket(d) \Rightarrow ticket(d)) \land trainTicket(d) \land ticketFor(d, c)$.

It remains to examine $(\Phi \circ \phi)^C$ and $(\Phi \circ \phi)^{C+E}$. We need to prove that:

(C) $(\Phi \circ \phi)^C \equiv \Phi_{IC}[C] \land \psi^C$, and

(D) $(\Phi \circ \phi)^{C+E} \equiv \Phi_{IC}[C + E] \land \text{eff}_a$.

Essentially, this means to prove that: (C) updating a formula with something it already implies does not incur any changes; (D) updating 1 with some formula yields a belief equivalent to that formula. To see this, compare (A) with (C) and (B) with (D).

While these two statements may sound quite trivial, it is in fact far from trivial to prove them for the wide variety of, partly rather complex, belief update operations in the literature. Here we build on the works by Katzuno and Mendelzon (1991) and Herzig and Rifi (1999). We need two of the postulates made by Katzuno and Mendelzon (1991), namely:

(U1) For any $\psi_1$ and $\psi_2$: $(\psi_1 \circ \psi_2) \Rightarrow \psi_2$.

(U2) For any $\psi_1$ and $\psi_2$: if $\psi_1 \Rightarrow \psi_2$ then $(\psi_1 \circ \psi_2) \equiv \psi_1$.

Herzig and Rifi (1999) prove that (U1) is "uncontroversial", meaning it is satisfied by all belief update operators they investigated (cf. above). They also prove that (U2) is equivalent to the conjunction of two weaker statements, of which only one is uncontroversial, namely:

(U2a) For any $\psi_1$ and $\psi_2$: $(\psi_1 \wedge \psi_2) \Rightarrow (\psi_1 \circ \psi_2)$.

The other statement is not uncontroversial. However, it is proved to be satisfied by all non-causal update operators under investigation, except the so-called Winslett's standard semantics (Winslett, 1990). The latter semantics is not useful in our context anyway. The only restriction it makes on the states in $res(s, a)$ is that they differ from $s$ only on the propositions mentioned in the update formula. In our case, these include all propositions appearing in $\Phi_{IC}[C \cup E]$, which is bound to be quite a lot. So, if we were to use Winslett's standard semantics, then $res(s, a)$ would be likely to retain hardly any information from $s$.

Consider now the formula $(\Phi \circ \phi)^C$ as specified in (A), $(\Phi \circ \phi)^C = (\Phi_{IC}[C] \wedge \psi^C) \circ \Phi_{IC}[C]$. We will now prove (C). This is indeed quite simple. We have that $(\Phi_{IC}[C] \wedge \psi^C) \Rightarrow \Phi_{IC}[C]$, so we can instantiate $\psi_1$ in (U2) with $\Phi_{IC}[C] \wedge \psi^C$, and $\psi_2$ in (U2) with $\Phi_{IC}[C]$. We obtain $(\Phi_{IC}[C] \wedge \psi^C) \circ \Phi_{IC} \equiv \Phi_{IC}[C] \wedge \psi^C$, and hence $(\Phi \circ \phi)^C \equiv \Phi_{IC}[C] \wedge \psi^C$ as desired. With what was said above, this result is not uncontroversial, but holds for all non-causal update operators (except Winslett's standard semantics) investigated by Herzig and Rifi (1999). In terms of the VTA example, (U2) allowed us to conclude that the update $trainTicket(c) \Rightarrow ticket(c)$ does not make any change to the previous belief, which already contains that property.

Next, consider the formula $(\Phi \circ \phi)^{C+E}$ as specified in (B), $(\Phi \circ \phi)^{C+E} = 1 \circ (\Phi_{IC}[C+E] \wedge \text{eff}_a)$. We now prove (D). By postulate (U1), we get that $(\Phi \circ \phi)^{C+E} \Rightarrow \Phi_{IC}[C+E] \wedge \text{eff}_a$, because $\Phi_{IC}[C+E] \wedge \text{eff}_a$ is the update formula $\psi_2$. For the other direction, we exploit (U2a). We instantiate $\psi_1$ in (U2a) with 1, and get that $1 \wedge (\Phi_{IC}[C+E] \wedge \text{eff}_a) \Rightarrow 1 \circ (\Phi_{IC}[C+E] \wedge \text{eff}_a)$, which is the same as $1 \wedge (\Phi_{IC}[C+E] \wedge \text{eff}_a) \Rightarrow (\Phi \circ \phi)^{C+E}$, which is equivalent to $\Phi_{IC}[C+E] \wedge \text{eff}_a \Rightarrow (\Phi \circ \phi)^{C+E}$. This proves the claim. Note that we have used only postulates that are uncontroversial according to Herzig and Rifi (1999). Reconsidering the VTA example, we have $\Phi_{IC}[C+E] \wedge \text{eff}_a = (trainTicket(d) \Rightarrow ticket(d)) \wedge trainTicket(d) \wedge ticketFor(d, c)$. The previous state does not say anything about these propositions, and is thus represented as 1. The postulates allow us to conclude that (for all belief update operators investigated by Herzig & Rifi, 1999) the resulting belief will be equivalent to $(trainTicket(d) \Rightarrow ticket(d)) \wedge trainTicket(d) \wedge ticketFor(d, c)$.

So far, we were restricted to the case where $\Phi$, the belief to be updated, corresponds to a single world state $s$. Consider now the more general case where $\Phi$ characterizes a belief $b$, and we want to characterize the set of states $res(b, a)$. At first glance, it seems that not much changes, because Katzuno and Mendelzon (1991) also make this following postulate:

(U8) For any $\psi_1$, $\psi_2$, and $\psi$: $(\psi_1 \vee \psi_2) \circ \psi \equiv (\psi_1 \circ \psi) \vee (\psi_2 \circ \psi)$.

This means that, if $\Phi$ consists of two alternate parts, then updating $\Phi$ is the same as taking the union of the updated parts. In other words, we can compute the update on a state-by-state basis. The statement (I) from above is still true, it's just that now $\psi^C$ is the disjunction over $\psi_s$ for all states $s \in b$, rather than only the single $\psi_s$. The rest of the argumentation stays exactly the same. Herzig and Rifi (1999) prove that (U8) is uncontroversial and leave it at that.

However, matters are not that simple. The source of complications is our use of a partial matches/conditional effects semantics. *The update formula $\phi$ is different for the individual states $s \in b$*. Hence we cannot directly apply (U8). Obviously, states $s_1 \in b$ where $a$ is applicable are updated differently from states $s_2 \in b$ where $a$ is not applicable – the latter are not updated at all.[16] A somewhat more subtle distinction between states in $b$ is which constants exist in them: for different sets of constants, the integrity constraints in the update are different. Hence, to obtain a "generic" update of $\Phi$, we have to split $\Phi$ into equivalence classes $\Phi_1, \ldots, \Phi_n$ where the states within each $\Phi_i$ cannot be distinguished based on $pre_a$ and based on the existing constants. Then, (U8) and the argumentation from above can be used to show the equivalent of (III) for each $\Phi_i'$. The last step, defining the final $\Phi \circ \phi$ to be the disjunction of the individual $\Phi_i \circ \phi_i$, appears sensible. But it does not follow immediately from Katzuno and Mendelzon (1991).

For illustration, consider a variant of the VTA example where we have two preceding states, one state $s$ where we have $train(c)$ as before, and a new state $s'$ where we have $ticket(c)$ instead. In $s'$, **bookTicket**$(c, d)$ is not applicable, and hence the update is different for $s$ and $s'$. The $s$ part is as above, yielding the result $\psi_s \wedge (trainTicket(d) \Rightarrow ticket(d)) \wedge trainTicket(d) \wedge ticketFor(d, c)$. The update to $s'$ is trivial, and yields $\psi_{s'}$ as its result. The final outcome is the disjunction of these two beliefs.

We point out that the situation is much easier if we consider plug-in matches (i.e., forced preconditions) instead of partial matches. There, $a$ is applicable to all states, and it is also easy to see that every state in $b$ has the same constants. Therefore, for plug-in matches, (III) follows immediately with (U8). In the above VTA example, an update would not be computed at all since **bookTicket**$(c, d)$ would not be considered to be applicable to the preceding belief. If $s'$ satisfies $train(c)$ but disagrees in some other aspect, e.g. (quite nonsensically) that also $ticket(c)$ holds, then the updated belief is equivalent to $(\psi_s \vee \psi_{s'}) \wedge (trainTicket(d) \Rightarrow ticket(d)) \wedge trainTicket(d) \wedge ticketFor(d, c)$.

## 4.4 Computational Properties

Paralleling our analysis for general $\mathcal{WSC}$ from Section 3.3, we now perform a brief complexity analysis of the $\mathcal{WSC}|_{fwd}$ special case. As before, we consider the "propositional" case which assumes a fixed upper bound on the arity of predicates, on the number of input/output parameters of each operator, on the number of variables appearing in the goal, and on the number of variables in any clause. Also as before, we consider the decision problems of checking plans, of deciding polynomially bounded plan existence, and of deciding unbounded plan existence, in that order.

In contrast to before, we cannot reuse results from the literature as much because, of course, the particular circumstances of $\mathcal{WSC}|_{fwd}$ have not been investigated before. We include proof sketches here, and refer to Appendix A for the detailed proofs.

---

16. One might speculate that the common update would be $pre_a \Rightarrow \phi$, but that is not the case. For example, under the possible models approach that we adopt in $\mathcal{WSC}$, updating $s$ where $s \models pre_a$ with $pre_a \Rightarrow \phi$ gives rise to result states that change $s$ to violate $pre_a$ instead of changing it to satisfy $\phi$.

Thanks to the simpler semantics as per Lemma 1, plan checking is much easier in $\mathcal{WSC}|_{fwd}$ than in $\mathcal{WSC}$.

**Theorem 4 (Plan Checking in $\mathcal{WSC}|_{fwd}$)** *Assume a $\mathcal{WSC}|_{fwd}$ task with fixed arity, and a sequence $\langle a_1, \ldots, a_n \rangle$ of actions. It is coNP-complete to decide whether $\langle a_1, \ldots, a_n \rangle$ is a plan.*

**Proof Sketch:** Hardness is obvious, considering an empty sequence. Membership can be shown by a guess-and-check argument. Say $C$ is the union of $C_0$ and all output constants appearing in $\langle a_1, \ldots, a_n \rangle$. We guess an interpretation $I$ of all propositions over $\mathcal{P}$ and $C$. Further, for each $1 \leq t \leq n$, we guess a set $C_t$ of constants. $I$ needs not be time-stamped because, once an action has generated its outputs, the properties of the respective propositions remain fixed forever. Thanks to Lemma 1, we can check in polynomial time whether (a) $I$ and the $C_t$ correspond to an execution of $\langle a_1, \ldots, a_n \rangle$. Also, we can check in polynomial time whether (b) $I$ and $C_n$ satisfy $\phi_G$. $\langle a_1, \ldots, a_n \rangle$ is a plan iff there is no guess where the answer to (a) is "yes" and the answer to (b) is "no". $\quad\square$

Membership in Theorem 4 remains valid when allowing parallel actions and multiple conditional effects – *provided* one imposes restrictions ensuring that the effects/actions applied simultaneously (in one step) can never be self-contradictory. Otherwise, checking plans also involves a consistency test for each plan step, which is an NP-complete problem. Note that it is quite reasonable to demand that simultaneous actions/effects do not contradict each other. Widely used restrictions imposed to ensure this are mutually exclusive effect conditions, and/or non-conflicting sets of effect literals.

We next consider polynomially bounded plan existence. Membership follows directly from Theorem 4. To prove hardness, we reduce from validity checking of a QBF formula $\exists X. \forall Y. \phi[X, Y]$. The constructed planning task allows to choose values for $X$, and thereafter to apply actions evaluating $\phi$ for arbitrary values of $Y$. The goal is accomplished iff a setting for $X$ exists that works for all $Y$.

**Theorem 5 (Polynomially Bounded Plan Existence in $\mathcal{WSC}|_{fwd}$)** *Assume a $\mathcal{WSC}|_{fwd}$ task with fixed arity, and a natural number $b$ in unary representation. It is $\Sigma_2^p$-complete to decide whether there exists a plan of length at most $b$.*

**Proof Sketch:** For membership, guess a sequence of at most $b$ actions. By Theorem 4, we can check with an NP oracle whether the sequence is a plan.

Hardness can be proved by reduction from validity checking of a QBF formula $\exists X. \forall Y. \phi[X, Y]$ where $\phi$ is in DNF normal form, i.e., $\phi = \bigvee_{j=1}^{k} \phi_j$. The key idea is to use outputs for the creation of "time steps", and hence ensure that the operators adhere to the restrictions of $\mathcal{WSC}|_{fwd}$. Setting $x_i$ is allowed only at time step $i$. That is, for each $x_i$ we have operators $o^{x_i 1}$ and $o^{x_i 0}$. These take as input a set of time steps $\{t_0, \ldots, t_{i-1}\}$ which are required to be successive, by the precondition $start(t_0) \wedge next(t_0, t_1) \wedge \cdots \wedge next(t_{i-2}, t_{i-1})$. They output a new time step $t_i$ which they attach as a successor of $t_{i-1}$, and they set $x_i$ to 1 and 0, respectively, at time step $i$. That is, they have an effect literal of the form $x_i(t_i)$ and $\neg x_i(t_i)$, respectively. The rest of the planning task consists of: operators $o^t$ that allow extending a sequence of time steps until step $B$, for a suitable value $B$ (see below); and of operators $o^{\phi_j}$ which allow achieving the goal, given $\phi_j$ is true at the end of a time step sequence of length $B$. There are no integrity constraints ($\Phi_{IC}$ is empty). The values of the $y_i$ are not specified, i.e., those variables can take on any value in the initial belief.

If $\exists X . \forall Y . \phi[X, Y]$ is valid then obviously one can construct a plan for the task simply by setting the $x_i$ accordingly, using the $o^t$ for stepping on to time $B$, and applying all the $o^{\phi_j}$. What necessitates our complicated construction is the other direction of the proof: namely, the plan may cheat by setting a $x_i$ to both 1 and 0. The construction ensures that this is costly, because such a plan is forced to maintain *two* parallel sequences of time steps, starting from the faulty $x_i$. We can choose a sufficiently large value for $B$, together with a sufficiently small plan length bound $b$, so that cheating is not possible. □

Our final result regards unbounded plan existence. Somewhat surprisingly, it turns out that this is still undecidable in $\mathcal{WSC}|_{fwd}$. Similar to the above, the key idea again is to let actions output a new "time step", thereby ensuring membership of the constructed task in $\mathcal{WSC}|_{fwd}$.

**Theorem 6 (Unbounded Plan Existence in $\mathcal{WSC}|_{fwd}$)** *Assume a $\mathcal{WSC}|_{fwd}$ task. The decision problem asking whether a plan exists is undecidable.*

**Proof Sketch:** By reduction from the halting problem for Abacus machines, which is undecidable. An Abacus machine consists of a tuple of integer variables $v_1, \ldots, v_k$ (ranging over all positive integers including 0), and a tuple of instructions $I_1, \ldots, I_n$. A state is given by the content of $v_1, \ldots, v_k$ plus the index $pc$ of the active instruction. The machine stops iff it reaches a state where $pc = n$. All $v_i$ are initially 0, and $pc$ is initially 0. The instructions either increment a variable and jump to another instruction, or they decrement a variable and jump to different instructions depending on whether or not the variable was already 0.

It is not difficult to encode an Abacus machine as a $\mathcal{WSC}|_{fwd}$ task. The two key ideas are: (1) design an operator that "outputs" the next successor to an integer; (2) design operators simulating the instructions, by stepping to successors or predecessors of integer values. In the latter kind of operators, membership in $\mathcal{WSC}|_{fwd}$ is ensured by letting the operators output a new "time step" to which the new variable values are associated. The goal asks for the existence of a time step where the active instruction is $I_n$. □

As argued at the end of Section 3.3 already, we don't deem undecidability of unbounded plan existence a critical issue in practice. Most planning tools are by nature semi-decision procedures, anyway. In particular, web service composition is typically expected to occur in a real-time setting, where severe time-outs apply.

## 4.5 Issues in Adapting CFF

In our view, the most crucial observation about $\mathcal{WSC}|_{fwd}$ is that we can now test plans in coNP, rather than in $\Pi_2^p$ as for general $\mathcal{WSC}$. Standard notions of planning under uncertainty have the same complexity of plan testing, and research has already resulted in a sizable number of approaches and (comparatively) scalable tools (Cimatti et al., 2004; Bryce et al., 2006; Hoffmann & Brafman, 2006; Palacios & Geffner, 2007). We will show in the next section that, under certain additional restrictions on $\mathcal{WSC}|_{fwd}$, these tools can be applied off-the-shelf. Regarding general $\mathcal{WSC}|_{fwd}$, the match in the complexity of plan testing suggests that the underlying techniques can be successfully adapted. In the following, we consider in some detail the CFF tool (Hoffmann & Brafman, 2006). Other promising options would be to extend MBP (Cimatti et al., 2004) or POND (Bryce et al., 2006), or to look into the compilation techniques investigated by Palacios and Geffner (2007).

CFF can be characterized as follows:

(1) Search is performed forward in the space of action sequences.

(2) For each sequence $\bar{a}$, a CNF formula $\phi(\bar{a})$ is generated that encodes the semantics of $\bar{a}$, and SAT reasoning over $\phi(\bar{a})$ checks whether $\bar{a}$ is a plan.

(3) Some reasoning results – namely the literals that are always true after executing $\bar{a}$ – are cached to speed up future tests.

(4) Search is guided by an adaptation of FF's (Hoffmann & Nebel, 2001) relaxed plan heuristic.

(5) Relaxed planning makes use of a strengthened variant of the CNF formulas $\phi(\bar{a})$ used for reasoning about action sequences, where most of the clauses are projected onto only 2 of their literals (i.e., all but 2 of the literals are removed from each respective clause).

All of these techniques should be self-explanatory, except possibly the last one. Projecting the CNF formulas ensures that the relaxed planning remains an over-approximation of the "real" planning, because the projected formulas allow us to draw more conclusions. At the same time, the projected formulas can be handled sufficiently runtime-efficiently.[17] The method for 2-projecting "most" of the clauses is, in a nutshell, to ignore all but one of the condition literals of each conditional effect in the relaxed planning graph.

It is fairly obvious that the basic answers given by CFF, i.e., the techniques (1) – (5), also apply in $\mathcal{WSC}|_{fwd}$. Note that, indeed, the main enabling factor here is that we can check plans in coNP, rather than in $\Pi_2^p$ as for general $\mathcal{WSC}$. This enables us to design the desired CNF formulas $\phi(\bar{a})$ in a straightforward fashion. If plan checking is $\Pi_2^p$-hard, then we either need to replace the CNF formulas with QBF formulas, or we have to create worst-case exponentially large CNF formulas. Both are, at the least, technically quite challenging.

The adaptation of CFF to $\mathcal{WSC}|_{fwd}$ is of more immediate promise, but is not trivial. It involves technical challenges regarding the on-the-fly creation of constants as well as the computation of the heuristic function. The latter also brings significant new opportunities in the WSC context, pertaining to the exploitation of typical forms of ontology axioms. Let us consider these issues in a little detail.

First, like most of today's planning tools, CFF pre-instantiates PDDL into a purely propositional representation, based on which the core planning algorithms are implemented. If one allows on-the-fly creation of constants, then pre-instantiation is no longer possible, and hence the adaptation to $\mathcal{WSC}|_{fwd}$ involves re-implementing the entire tool. While this is a challenge in itself, there are more difficult obstacles to overcome. A sloppy formulation of the key question is: How many constants should we create? One can, of course, create a new tuple of constants for (the outputs of) each and every new action application. However, it seems likely that such an approach would blow up the representation size very quickly, and would hence be infeasible. So one should instead share output constants where reasonable. But how does one recognize the "reasonable" points? This issue is especially urgent inside the heuristic function. Namely, it is easy to see that, in the worst case, the relaxed planning graph grows exponentially in the number of layers. Just imagine an example where web service $w_1$ takes an input of type $A$ and generates an output of type $B$, whereas $w_2$ takes an input of type $B$ and generates an output of type $A$. Starting with one constant of type $A$ and one of type $B$, we get 2 constants of each type in the next graph layer. Then, each of $w_1$ and $w_2$ can be applied two times, and we get $4$ constants of each type in the next graph layer, and so forth. This dilemma probably cannot be handled without making further approximations in the relaxed planning graph.

---

17. Inside the heuristic function, the formulas come from relaxed planning graphs which can be quite big. So handling them without further approximations seems hopeless. This is discussed in detail by Hoffmann and Brafman (2006).

One a more positive note, it seems possible to exploit the most typical structures of ontologies in practice. In particular, most practical ontologies make extensive use of subsumption relations, structuring the domain of interest into a concept hierarchy. Additional ontology axioms often come in the form of constraints on relations (reflexivity, symmetry, transitivity) or on the typing or number of relation arguments. It may make sense to exploit some of these structures for optimizing the formulas $\phi(\bar{a})$ and the associated SAT reasoning. Certainly, it makes sense to exploit these structures inside the heuristic function. One can include specialized analysis and sub-solver techniques that recognize these structures and solve them separately in order to obtain more precise relaxed plans. One can even try to take into account *only* these structures inside the relaxed planning, and hence (potentially) obtain a very fast heuristic function.

## 5. Compilation to Initial State Uncertainty

We now show that, under certain additional restrictions, off-the-shelf scalable tools for planning under uncertainty can be exploited to solve $\mathcal{WSC}|_{fwd}$. The main limiting factors are: (1) These tools do not allow the generation of new constants. (2) These tools allow the specification of a clausal formula only for the initial state, not for all states. Our approach to deal with (1) considers a set of constants fixed a priori, namely the initially available constants plus additional "potential" constants that can be used to instantiate outputs. Our more subtle observation is that, within a special case of $\mathcal{WSC}|_{fwd}$, where the dynamics of states become predictable a priori, one can also deal with (2) in a natural way.

In what follows, we first introduce our core observation of a case where the state space becomes "predictable", in a certain sense. We then observe that predictability is naturally given in a special case of forward effects, which we term *strictly forward effects*. We discuss the strengths and limitations of this new special case. We finally provide a compilation of strictly forward effects into planning under initial state uncertainty.

### 5.1 Predictable State Spaces

Our core observation is based on a notion of *compatible actions*. Assume a $\mathcal{WSC}|_{fwd}$ task $(\mathcal{P}, \Phi_{IC}, \mathcal{O}, C_0, \phi_0, \phi_G)$. Two actions $a$, $a'$ are compatible if either $E_a \cap E_{a'} = \emptyset$, or $\mathrm{eff}_a = \mathrm{eff}_{a'}$. That is, $a$ and $a'$ either have disjunct outputs – and hence affect disjunct sets of literals since we are in $\mathcal{WSC}|_{fwd}$ – or their effects agree completely. A set $\mathcal{A}$ of actions is compatible if $E_a \cap C_0 = \emptyset$ for all $a \in \mathcal{A}$, and every pair of actions in $\mathcal{A}$ is compatible.

Lemma 2 states that, given the used actions are compatible, every state that can ever be reached satisfies all action effects, modulo the existing constants.

**Lemma 2 (Predictable State Spaces in $\mathcal{WSC}|_{fwd}$)** *Assume a $\mathcal{WSC}|_{fwd}$ task, a compatible set of actions $\mathcal{A}$, and a state $s$ that can be reached with actions from $\mathcal{A}$. Then $s \models \phi_0$ and, for all $a \in \mathcal{A}$, if $E_a \subseteq C_s$ then $s \models \mathrm{eff}_a$.*

**Proof:** The proof is by induction. In the base case, for $s \in b_0$, the claim holds by definition since $C_s \cap E_a = \emptyset$ for all $a \in \mathcal{A}$. Say $s'$ is reached from $s$ by an action $a \in \mathcal{A}$. If $a$ is not applicable to $s$, with induction assumption there is nothing to prove. Otherwise, because we are in $\mathcal{WSC}|_{fwd}$, by Lemma 1 we have that $res(s, a) = \{(C', I') \mid C' = C_s \cup E_a, I'|_{C_s} = I_s, s \models \Phi_{IC} \wedge \text{eff}_a\}$. With induction assumption applied to $s$, we have $res(s, a) = \{(C', I') \mid C' = C_s \cup E_a, s \models \phi_0 \wedge \bigwedge_{a' \in \mathcal{A}, E_{a'} \subseteq C_s} \text{eff}_{a'} \wedge \Phi_{IC} \wedge \text{eff}_a\}$. Now, if any $a' \in \mathcal{A}$ has $E_{a'} \subseteq C_s \cup E_a$ but $E_{a'} \not\subseteq C_s$, then we have $E_{a'} \cap E_a \neq \emptyset$ and hence $\text{eff}_{a'} = \text{eff}_a$ by prerequisite. This concludes the argument. □

By virtue of this lemma, the possible configurations of all constants that can be generated by actions from $\mathcal{A}$ are characterized by the formula $\Phi_{IC} \wedge \phi_0 \wedge \bigwedge_{a \in \mathcal{A}} \text{eff}_a$. Since all parts of this formula are known prior to planning, the set of possible configurations is "predictable". Before we even begin to plan, we already know how the constants will behave if they are generated. So we can list the possible behaviors of all potential constants in our initial belief, and let the actions affect only those constants which actually exist. In other words, we can compile into initial state uncertainty. We will detail this further below. First, we need to identify a setting in which Lemma 2 can actually be applied.

## 5.2 Strictly Forward Effects

Given a $\mathcal{WSC}|_{fwd}$ task, we must settle for a finite set $\mathcal{A}$ of compatible actions that the planner should try to compose the plan from. One option is to simply require every action to have its own unique output constants. This appears undesirable since planning tasks often contain many actions, and so the set of potential constants would be huge. Further, to enable chaining over several actions, the potential constants should be allowed to instantiate the input parameters of every operator, hence necessitating the creation of a new action and, with that, more new potential constants. It is unclear where to break this recursion, in a sensible way.

Herein, we focus instead on a restriction of $\mathcal{WSC}|_{fwd}$ where it suffices to assign unique output constants to individual *operators*, rather than to individual actions.

**Definition 2** *Assume a $\mathcal{WSC}$ task* $(\mathcal{P}, \Phi_{IC}, \mathcal{O}, C_0, \phi_0, \phi_G)$. *The task has* strictly forward effects *iff*:

1. *For all $o \in \mathcal{O}$, and for all $l[X] \in \text{eff}_o$, we have $|X| > 0$ and $X \subseteq Y_o$.*

2. *For all clauses $\phi \in \Phi_{IC}$, where $\phi = \forall x_1, \ldots, x_k : l_1[X_1] \vee \cdots \vee l_n[X_n]$, we have $X_1 = \cdots = X_n$.*

*The set of all $\mathcal{WSC}$ tasks with strictly forward effects is denoted with $\mathcal{WSC}|_{sfwd}$.*

The second condition is identical to the corresponding condition for $\mathcal{WSC}|_{fwd}$. The first condition is strictly stronger. While $\mathcal{WSC}|_{fwd}$ requires that *at least one* effect literal variable is taken from the outputs, $\mathcal{WSC}|_{sfwd}$ requires that *all* these variables are taken from the outputs. Therefore, obviously, $\mathcal{WSC}|_{sfwd} \subset \mathcal{WSC}|_{fwd}$. Note that the $\mathcal{WSC}$ task formulated in Example 2 is a member of $\mathcal{WSC}|_{sfwd}$.

The key property of $\mathcal{WSC}|_{sfwd}$ is that, without input variables in the effect, all actions based on the operator will have the same effect. So, for the action set to be compatible, all we need is to choose a set of unique output constants for every operator. Indeed, we can do so for every set of operators whose effects are pairwise identical. We can also choose several sets of output constants for each such group of operators.

## 5.3 Modeling Power

The limitations of $\mathcal{WSC}|_{fwd}$, discussed in Section 4.2, are naturally inherited by $\mathcal{WSC}|_{sfwd}$. Moreover, unlike $\mathcal{WSC}|_{fwd}$, we cannot state any properties in the effect that connect the inputs to the outputs. This is a serious limitation. For illustration, consider the small VTA example we have been using. The operator **bookTicket** has an effect $ticketFor(y, x)$, relating the produced ticket $y$ to the train $x$ given as input. Clearly, the notion of a "ticket" is rather weak if we cannot state what the ticket is actually valid for. Another interesting case is the one where we extend Example 2 by considering two proteins rather than just one. That is, we set $C_0 = \{c, c'\}$, $\phi_0 =$ cellProtein$(c) \wedge$ cellProtein$(c')$. We wish to encode that we need the combined presentation for both of those, i.e., $\phi_G = \exists y :$ combinedPresentation$(y, c) \wedge$ combinedPresentation$(y, c')$. In $\mathcal{WSC}|_{fwd}$, we can solve this by including, for every information providing operator, the input variable $x$ into the effect literal. For example, we set **getInfo3D**$_{1n55} := (\{x\}, 1n55(x), \{y\},$ Info3D$(y, x))$. This is not possible in $\mathcal{WSC}|_{sfwd}$.

To some extent, these difficulties can be overcome by encoding the relevant inputs into predicate names. To handle composition for the two proteins $c$ and $c'$, this would essentially mean making a copy of the entire model and renaming the part for $c'$. The goal would be $\phi_G = \exists y, y' :$ combinedPresentation$(y) \wedge$ combinedPresentation$'(y')$, and the operator preconditions would make sure that combinedPresentation$(y)$ is generated as before, while combinedPresentation$'(y')$ is generated using the new operators. Note that this a rather dirty hack, and that it depends on knowing the number of copies needed, prior to planning. The equivalent solution for the VTA would introduce a separate "ticketFor-x" predicate for every entity $x$ for which a ticket may be bought. At the very least, this would result in a rather oversized and unreadable model. A yet more troublesome case is the time-step construction outlined in Section 4.2, where we added a new output variable $t$ into each effect and related that via an effect literal $next(prevt, t)$ to a previous time step $prevt$ provided as input. In $\mathcal{WSC}|_{sfwd}$, we can no longer relate $t$ to $prevt$ so there is no way of stating which time step happens after which other one. Trying to encode this information into predicate names, we would have to include one predicate per possible time step. This necessitates assuming a bound on the number of time steps, a clear limitation with respect to the more natural encoding.

Despite the above, $\mathcal{WSC}|_{sfwd}$ is far from a pathological and irrelevant special case. An example where it applies is the domain of proteins as shown in Example 1. Similarly, the hospital domain discussed in Section 4.2 can be naturally modeled in $\mathcal{WSC}|_{sfwd}$. More generally, there is in fact a wealth of WSC formalisms which do not encode any connections between inputs and outputs. For example, that category contains all formalisms which rely exclusively on specifying the "types" of input and output parameters. The information modeled with such types is only what kind of input a service requires, and what kind of output it produces – for example, "input is a train" and "output is a ticket". Examples of such formalisms are various notions of message-based composition (Zhan et al., 2003; Constantinescu et al., 2004a; Lecue & Leger, 2006; Lecue & Delteil, 2007; Kona et al., 2007; Liu et al., 2007). In fact, the early versions of OWL-S regarded inputs and outputs as independent semantic entities, using a Description Logic formalization of their types.

Thus, the existence of a compilation from $\mathcal{WSC}|_{sfwd}$ into planning under uncertainty is quite interesting. It shows how a composition model similar to the early versions of OWL-S, in a general form with partial matches and powerful background ontologies, can be attacked by off-the-shelf planning techniques. This opens up a new connection between WSC and planning.

### 5.4 Compilation

We compile a $\mathcal{WSC}|_{sfwd}$ task into a task of conformant planning under initial state uncertainty, which takes the form $(\mathcal{P}, \mathcal{A}, \phi_0, \phi_G)$. $\mathcal{P}$ is the finite set of propositions used. $\mathcal{A}$ is a finite set of actions, where each $a \in \mathcal{A}$ takes the form $(\text{pre}(a), \text{eff}(a))$ of a pair of sets of literals over $\mathcal{P}$. $\phi_0$ is a CNF formula over $\mathcal{P}$, $\phi_G$ is a conjunction of literals over $\mathcal{P}$. These notions are given a standard belief state semantics. A state is a truth value assignment to $\mathcal{P}$. The initial belief is the set of states satisfying $\phi_0$. The result of executing an action $a$ in a state $s$ is $res(s, a) := s$ if $s \not\models \text{pre}(a)$,[18] and otherwise $res(s, a) := (s \cup add(a)) \setminus del(a)$. Here we use the standard notation that gives $s$ in terms of the set of propositions that it makes true, uses $add(a)$ to denote the positive literals in $\text{eff}(a)$, and $del(a)$ to denote the negative literals in $\text{eff}(a)$. Extension of $res$ to beliefs and the definition of a plan remain unchanged.

Assume a $\mathcal{WSC}|_{sfwd}$ task $(\mathcal{P}, \Phi_{IC}, \mathcal{O}, C_0, \phi_0, \phi_G)$. The compiled task $(\mathcal{P}', \mathcal{A}, \phi_0', \phi_G')$ makes use of a new unary predicate $Ex$ that expresses which constants have yet been brought into existence. The compilation is obtained as follows. For each operator $o \in \mathcal{O}$, with outputs $Y_o = \{y_1, \ldots, y_k\}$, we create a set of new constants $E_o = \{e_1, \ldots, e_k\}$. Then, $C := C_0 \cup \bigcup_{o \in \mathcal{O}} E_o$ will be the set of constants fixed a priori. Initialize $\mathcal{A} := \emptyset$. For each operator $o \in \mathcal{O}$, include into $\mathcal{A}$ the set of actions resulting from using $C$ to instantiate the precondition $\text{pre}_o \wedge (\bigwedge_{x \in X_o} Ex(x)) \wedge (\bigwedge_{e \in E_o} \neg Ex(e))$. Give each of these actions the same effect, $\bigwedge_{e \in E_o} Ex(e)$. In words, we instantiate $o$'s outputs with $E_o$, we enrich $o$'s precondition by saying that all inputs exist and that all outputs do not yet exist, and we replace $o$'s effect with a statement simply bringing the outputs into existence.

Replacing the effects in this way, where do the original effects go? They are *included into the initial state formula*. That is, we instantiate $\phi_0'$ as the conjunction of $\text{eff}_o[E_o/Y_o]$ for all operators $o \in \mathcal{O}$. Then, we instantiate all clauses in $\Phi_{IC}$ with $C$ and conjoin this with $\phi_0'$. We obtain our final $\phi_0'$ by further conjoining this with $\phi_0 \wedge (\bigwedge_{c \in C_0} Ex(c)) \wedge \bigwedge_{c \in C \setminus C_0} \neg Ex(c)) \wedge \neg Goal$. Here, $Goal$ is a new proposition. It serves to model the goal. Namely, we have to introduce a set of artificial goal achievement actions. The goal has the form $\phi_G = \exists x_1, \ldots, x_k.\phi[x_1, \ldots, x_k]$. The new actions are obtained by instantiating the operator $(\{x_1, \ldots, x_k\}, \phi \wedge \bigwedge_{i=1}^{k} Ex(x_i), \emptyset, Goal)$ with $C$. That is, the goal achievement actions instantiate the existentially quantified variables in the goal with all possible constants. Those actions are added to the set $\mathcal{A}$. The overall compiled task now takes the form $(\mathcal{P}', \mathcal{A}, \phi_0', Goal)$, where $\mathcal{P}'$ is simply the set of mentioned propositions.

In summary, we compile a $\mathcal{WSC}|_{sfwd}$ task $(\mathcal{P}, \Phi_{IC}, \mathcal{O}, C_0, \phi_0, \phi_G)$ into a conformant planning task $(\mathcal{P}', \mathcal{A}, \phi_0', \phi_G')$ as follows:

- For each operator $o \in \mathcal{O}$, create a unique set of new constants $E_o = \{e_1, \ldots, e_k\}$ where $Y_o = \{y_1, \ldots, y_k\}$. We denote $C := C_0 \cup \bigcup_{o \in \mathcal{O}} E_o$.

- $\mathcal{P}'$ contains all instantiations, with $C$, of $\mathcal{P}$ plus two new predicates, $Ex$ and $Goal$. $Ex$ has arity 1 and expresses which constants have yet been brought into existence. $Goal$ has arity 0 and forms the new goal, i.e., $\phi_G' = Goal$.

- The actions $\mathcal{A}$ are the instantiations of all $o \in \mathcal{O}$, where $X_o$ is instantiated with $C$, and $Y_o$ is instantiated with $E_o$. The preconditions are enriched with $(\bigwedge_{x \in X_o} Ex(x)) \wedge (\bigwedge_{e \in E_o} \neg Ex(e))$, the effects are replaced by $\bigwedge_{e \in E_o} Ex(e)$.

---

18. As before, we give the actions a conditional effects semantics, rather than the more usual distinction between forced preconditions, and non-forced effect conditions.

- Further, $\mathcal{A}$ contains goal achievement actions, achieving *Goal* under preconditions instantiating $\phi_G$ with $C$.

- The original action effects, i.e., the conjunction of $\text{eff}_o[E_o/Y_o]$ for all operators $o \in \mathcal{O}$, is moved into $\phi'_0$. Further, $\phi'_0$ contains $\phi_0$, $\Phi_{IC}$ instantiated with $C$, and $(\bigwedge_{c \in C_0} Ex(c) \wedge \bigwedge_{c \in C \setminus C_0} \neg Ex(c)) \wedge \neg Goal$.

In the terminology of Section 5.1, this means that we choose the set $\mathcal{A}$ of actions as all actions that can be obtained from an operator $o \in \mathcal{O}$ by instantiating the inputs with constants from $C$, and the outputs with $E_o$. As suggested by Lemma 2, the initial state formula $\phi'_0$ of the compiled task describes the possible configurations of the constants $C$, and the only effect of applying an action is to bring the respective output constants into existence. Note that, although the effects of the compiled actions are all positive, planning is still hard (coNP-complete, to be precise) due to the uncertainty. (If we allow $\mathcal{WSC}$ operators to also delete constants, then we have negative effects – deleting constants – in the compiled task.)

According to the above strategy, we create only one set of output constants per operator, and we do not take into account sets of operators that have identical effects. This is only to simplify the presentation. Our results carry over immediately to more complicated strategies that create more than one set of output constants per operator, as well as to strategies that share sets of output constants between operators with identical effects. It should be noted, however, that operators whose effects are not identical can not, in general, share their outputs. In particular, if the two effects are in conflict, e.g., InfoDSSP($d$) and ¬InfoDSSP($d$), then the initial state formula $\phi'_0$ as above is unsatisfiable. The compiled planning task is then trivially solved by the empty plan, and, of course, does not encode solutions in the original problem.

**Example 3** *Re-consider the planning task defined in Example 2. We specify a compiled task. We set $C = \{c, d, e, f\}$ where $c$ is the only initially available constant, and $d, e, f$ are potential constants for operator outputs. The compiled planning task $(\mathcal{P}', \mathcal{A}, \phi'_0, \phi'_G)$ is the following:*

- $\mathcal{P}' = \{$protein, cellProtein, G, H, I, 1n55, 1kw3, combinedPresentation, InfoDSSP, Info3D, Ex, Goal$\}$, *where all the predicates except Goal are unary (have one argument).*

- $\mathcal{A}$ *consists of all instantiations of:*

  - **getInfoDSSP**$_G[d/y]$: $(\{x\}, G(x) \wedge Ex(x) \wedge \neg Ex(d), Ex(d))$
  - **getInfoDSSP**$_H[d/y]$: $(\{x\}, H(x) \wedge Ex(x) \wedge \neg Ex(d), Ex(d))$
  - **getInfoDSSP**$_I[d/y]$: $(\{x\}, I(x) \wedge Ex(x) \wedge \neg Ex(d), Ex(d))$
  - **getInfo3D**$_{1n55}[e/y]$: $(\{x\}, 1n55(x) \wedge Ex(x) \wedge \neg Ex(e), Ex(e))$
  - **getInfo3D**$_{1kw3}[e/y]$: $(\{x\}, 1kw3(x) \wedge Ex(x) \wedge \neg Ex(e), Ex(e))$
  - **combineInfo**$[f/y]$: $(\{x_1, x_2\}, InfoDSSP(x_1) \wedge Info3D(x_2) \wedge Ex(x_1) \wedge Ex(x_2) \wedge \neg Ex(f), Ex(f))$
  - **GoalOp**: $(\{x\}, combinedPresentation(x) \wedge Ex(x), Goal)$

- $\phi'_0$ *is the conjunction of:*

  - *all instantiations of $\Phi_{IC}$ – [consisting of the five axioms given in Example 2]*

- – $cellProtein(c) - [\phi_0]$
- – $InfoDSSP(d) \wedge Info3D(e) \wedge combinedPresentation(f) - [original\ action\ effects]$
- – $Ex(c) \wedge \neg Ex(d) \wedge \neg Ex(e) \wedge \neg Ex(f) - [constants\ existence]$
- – $\neg Goal - [goal\ not\ yet\ achieved]$

- $\phi'_G = Goal$

*Now consider again the plan for the original task (see Example 2):* $\langle \mathbf{getInfoDSSP}_G(c,d),$
$\mathbf{getInfoDSSP}_H(c,d), \mathbf{getInfo3D}_{1n55}(c,e), \mathbf{getInfo3D}_{1kw3}(c,e), \mathbf{combineInfo}(d,e,f)\rangle.$

*To illustrate, we now verify that this plan yields a plan for the compiled task. In that task, the initial belief $b_0$ consists of all states $s$ where $c$ is the only existing constant, $d, e, f$ satisfy the respective effects, and $s \models \Phi_{IC} \wedge cellProtein(c)$. Now we apply the action sequence:*

1. Apply $\mathbf{getInfoDSSP}_G(c,d)$ to $b_0$. *We get to the belief $b_1$ which is the same as $b_0$ except that, in all $s \in b_0$ where $s \models G(c)$, $d$ now also exists.*

2. Apply $\mathbf{getInfoDSSP}_H(c,d)$ to $b_1$. *We get to the belief $b_2$ which is the same as $b_1$ except that, in all $s \in b_1$ where $s \models H(c)$, $d$ exists.*

3. Apply $\mathbf{getInfo3D}_{1n55}(c,e)$ to $b_2$, *yielding $b_3$.*

4. Apply $\mathbf{getInfo3D}_{1kw3}(c,e)$ to $b_3$. *This brings us to $b_4$ where we have $Ex(e)$ for all $s \in b_2$ with $s \models 1n55(c)$ or $s \models 1kw3(c)$.*

5. Apply $\mathbf{combineInfo}(d,e,f)$ to $b_4$. *This brings us to $b_5$ which is like $b_4$ except that all $s \in b_4$ where both $d$ and $e$ exist now also have $Ex(f)$.*

6. Apply $\mathbf{GoalOp}(f)$ to $b_5$, *yielding $b_6$.*

*The same reasoning over $\Phi_{IC}$ used in Example 2 to show that $b_5$ satisfies the original goal, can now be used to show that $\mathbf{GoalOp}(f)$ is applicable in all $s \in b_5$ and hence the resulting belief $b_6$ satisfies the goal. So we obtain a plan for the compiled task simply by attaching a goal achievement action to the original plan.*

To prove soundness and completeness of the compilation, we need to rule out *inconsistent* operators, i.e., operators whose effects are in conflict with the background theory (meaning that $\Phi_{IC} \wedge \exists X_o, Y_o : \text{eff}_o$ is unsatisfiable). For example, this is the case if $\forall x : \neg A(x) \vee \neg B(x)$ is contained in $\Phi_{IC}$, and $\text{eff}_o = A(y) \wedge B(y)$. In the presence of such an operator, the initial belief of the compiled task is empty, making the task meaningless. Note that inconsistent operators can never be part of a plan, and hence can be filtered out as a pre-process. Note also that, in $\mathcal{WSC}|_{sfwd}$, an operator is inconsistent iff all actions based on it are inconsistent.

Non-goal achievement actions in $\mathcal{A}$ correspond to actions in the original task, in the obvious way. With this connection, we can transform plans for the compiled task directly into plans for the original task, and vice versa.

**Theorem 7 (Soundness of Compilation)** *Consider the $\mathcal{WSC}|_{sfwd}$ task $(\mathcal{P}, \Phi_{IC}, \mathcal{O}, C_0, \phi_0, \phi_G)$ without inconsistent operators and a plan $\langle a_1, \ldots, a_n \rangle$ for the compiled task $(\mathcal{P}', \mathcal{A}, \phi'_0, \phi'_G)$. Then the sub-sequence of non-goal achievement actions in $\langle a_1, \ldots, a_n \rangle$ is a plan for the task $(\mathcal{P}, \Phi_{IC}, \mathcal{O}, C_0, \phi_0, \phi_G)$.*

**Proof Sketch:** For an arbitrary sequence of non-goal achievement actions, denote by $b$ the belief after execution in the original task, and by $\bar{b}$ the belief after execution in the compiled task. For a state $s$ in the original task, denote by $[s]$ the class of all compiled-task states $\bar{s}$ over the constants $C_0 \cup \bigcup_{o \in \mathcal{O}} E_o$ so that $\{c \mid \bar{s}(Ex(c)) = 1\} = C_s$, $\bar{s}|_{C_s} = I_s$, and $\bar{s} \models \Phi_{IC} \wedge \phi_0 \wedge \bigwedge_{o \in \mathcal{O}} \mathrm{eff}_o[E_o]$. One can prove that $\bar{b} = \bigcup_{s \in b}[s]$. The claim follows directly from that. □

**Theorem 8 (Completeness of Compilation)** *Consider the $\mathcal{WSC}|_{sfwd}$ task $(\mathcal{P}, \Phi_{IC}, \mathcal{O}, C_0, \phi_0, \phi_G)$ without inconsistent operators and a plan $\langle a_1, \ldots, a_n \rangle$ where every operator $o$ appears with at most one instantiation $E_o$ of the outputs. Then $\langle a_1, \ldots, a_n \rangle$ can be extended with goal achievement actions to form a plan for the compiled task $(\mathcal{P}', \mathcal{A}, \phi'_0, \phi'_G)$ obtained using the outputs $E_o$.*

**Proof Sketch:** Follows immediately from $\bar{b} = \bigcup_{s \in b}[s]$ as shown for the proof of Theorem 7. Say one executes $\langle a_1, \ldots, a_n \rangle$ in the compiled task, ending in a belief $\bar{b}$. From there, a plan for the compiled task can be obtained simply by attaching one goal achievement action for every tuple of constants satisfying $\phi_G$ in a world state from $\bar{b}$. □

The reader may have noticed that the number of instantiations of the goal achievement operator is exponential in the arity of the goal. In the worst case, all these instantiations must be included in the plan for the compiled task. In particular, this may happen in the plan constructed as per the proof of Theorem 8. However, for practical purposes it appears reasonable to assume a fixed upper bound on the number of goal variables.

As indicated, the proofs of Theorems 7 and 8 remain valid when allowing more than one $E_o$ per operator, and/or when operators with identical effects share output constants. Note that operators have identical effects if several web services provide alternative ways of achieving something. Example 3 illustrates such a situation (cf. our earlier discussion in Section 3.2). In our experiments as described in the next section, all groups of operators with identical effects are assigned the same output constants.

# 6. Empirical Results

To show that the compilation approach has merits, we now report on a number of empirical experiments using CFF as the underlying planner. We start with a discussion of the general experimental setup and then discuss the results for two different test scenarios.

## 6.1 Experiments Setup

We implemented the compilation from $\mathcal{WSC}|_{sfwd}$ into planning under uncertainty as described above, and connected it to the CFF tool. It should be noted here that, although the compiled planning tasks do not have delete effects, they are *not* solved by CFF's relaxed-plan-based heuristic function. That function makes a further relaxation ignoring all but one of the conditions of each effect (see the earlier discussion of CFF in Section 4.5). Ignoring all but one condition significantly affects the compiled tasks because their effects typically involve many conditions, particularly those conditions stating that all inputs exist and all outputs do not yet exist.

One problematic point in evaluating planning-based WSC is the choice of test cases. The field is still rather immature, and due to the widely disparate nature of existing WSC tools, there is no common set of benchmarks.[19] In fact, because web service composition is such a new topic posing so many challenges to existing techniques, the different works differ widely in terms of both their underlying purpose, and the specific aspect of WSC they address. A detailed discussion of existing WSC tools is given below in Section 7. The method we choose for evaluation is to design two test scenarios that reflect what are intuitively relevant kinds of problem structures in potential applications of planning-based WSC, and that are scalable in a number of interesting parameters. We test the reaction of our approach to these parameters.

While our test scenarios are artificial benchmarks, and cannot lead to broad conclusions of significance for practice, they do allow us to draw conclusions about planning behavior in differently structured test problems. Our solution method scales quite well in most of the tested cases, efficiently finding solutions that involve many web service calls, and that successfully employ only those services that are really necessary. Viewing these results in isolation, one can conclude that representation techniques and heuristic functions from planning under uncertainty may be useful to attack large and complex planning-like WSC instances.

A comparison to alternative WSC tools is, again, problematic, due to the broad range of problems the tools can solve, the different kinds of solutions they find, and the different kinds of input syntax/language they read. To obtain at least some notion of empirical comparison to these tools, in the following we consider only expressivity ("How general is the input language of a tool?") and scalability ("How quickly can the tool compose?"). Each of the existing WSC tools constitutes a separate point in the trade-off between these two. The question then is whether our compilation approach, restricting to $\mathcal{WSC}|_{sfwd}$ and using CFF to solve the compiled tasks, is a sensible point in that trade-off.

In terms of expressivity, our approach is located in between very general planning methods (like Eiter et al., 2003, 2004; Giunchiglia et al., 2004), inspired by the actions and change literature, and the more restricted methods that have been applied to WSC so far. The question is whether we gain scalability in comparison to the more expressive methods.

We confirm in our experiments that the answer is, as expected, "yes". We run the DLVK tool (Eiter et al., 2003, 2004), which handles a powerful planning language based on logic programming. That language in particular features "static causal rules" which are similar to the integrity constraints in fully general $\mathcal{WSC}$.[20] In that sense, from our perspective DLVK is a "native WSC tool" that handles ontology axioms directly rather than via restricting their expressivity and compiling them away. In particular, we encoded our $\mathcal{WSC}$ test problems directly in DLVK's input language, without the compilation that we use for CFF.

DLVK relies on answer set programming, instead of relaxed plan heuristics, to find plans. Further, in the style of many reasoning-based planners, DLVK requires as input a length bound on the plan, and can hence be used to find optimal plans by running it several times with different bounds. In all cases, we ran DLVK only once, with the bound corresponding to the optimal plan length. Even so, DLVK is much slower than CFF, solving only a small fraction of our test instances. We do not wish to over-interpret these results. All we conclude is that $\mathcal{WSC}|_{sfwd}$ constitutes an interesting point in the trade-off between expressivity and scalability in WSC.

---

19. While the VTA example could be considered one such benchmark, essentially every individual approach defines its own particular version of that example.

20. The similarity lies in that both static causal rules and fully general integrity constraints can, as a side effect of applying an action, yield ramifications affecting the properties inherited from the previous state.

When running some first tests with the compilation approach, we noticed that the encoding as per Section 5.4 is unnecessarily generous about the set of initial states. Observe that our compiled tasks are always easier to solve if more propositions are true in the initial state. This is, simply, because all literals in operator preconditions, effects, and the goal are positive. Hence, if a proposition $p$ does not appear positively in any initial state clause, then one can set $p$ to 0 initially, and thereby reduce the number of initial states, without introducing any new plans.[21] Setting a proposition to 0 may cause unit propagations, setting other propositions to 1 or 0. We iterate these steps until a fixpoint occurs. The resulting initial state description is stricter than before, and yields better performance both for CFF and for DLVK. We use this optimized encoding in all the experiments reported below.

We also experimented with another optimization. That optimization makes the assumption that the constants requested by the goal will be generated in a step-wise fashion, where each intermediate constant is generated with certainty before generating the next constant. Recall that in the encoding as per Section 5.4, the existence of the inputs of operators, i.e., the condition $\bigwedge_{x \in X_o} exists(x)$, is part of the operator precondition and is thus interpreted under a conditional effects semantics. However, both CFF and DLVK offer a distinction between effect conditions and *forced* preconditions that must hold in the entire belief for the action to be applicable. We can exploit that distinction to postulate that the condition $\bigwedge_{x \in X_o} exists(x)$ is forced. This reduces the state space, but may cut out solutions. The reduction is quite beneficial both for CFF and for DLVK. Since the optimization affects the set of plans, we switch it on only in part of the test cases, to point out the possible speed-up. The tests where the optimization is switched on are discussed in the text, and indicated by the keyword **forced** in the name of the test case.

We use two versions of CFF. One is CFF's default configuration which makes use of FF's "enforced Hill-climbing" search algorithm as well as its "helpful actions pruning" technique (Hoffmann & Nebel, 2001). In the other configuration, CFF helpful actions pruning is turned off, and the search proceeds in standard "greedy best-first" fashion, with an open queue ordered by increasing heuristic values. We henceforth denote the former configuration with CFF-def and the latter configuration with CFF-std.

All results were obtained on a 2.8GHz Pentium IV PC running Linux. All tests were run with a time-out of 600 seconds CPU, and limiting memory usage to 1 GB.

## 6.2 Subsumption Hierarchies

We first investigate how well our approach can deal with scaling subsumption hierarchies, and with building chains of successively created entities (outputs). For that purpose, we design a test scenario called **SH**, which demands the composition of web services realizing a chain of generation steps, where every generation step has to deal with a subsumption hierarchy.

The scenario is depicted in Figure 2. There are $n$ "top-level" concepts $TL_1, \ldots, TL_n$, depicted with "TL" in Figure 2. The goal input is $TL_1$, the goal output is $TL_n$. Beneath each $TL_i$, there is a tree-shaped hierarchy of sub-concepts. More precisely, the tree is perfectly balanced with branching factor $b$, and has depth $d$. The inner nodes of the tree are called "intermediate-level" (or simply "intermediate") concepts, depicted with "IL" in Figure 2. The leaf nodes of the tree are called "basic-level" (or simply "basic") concepts, depicted with "BL" in Figure 2. For every non-leaf concept $C$ in the tree, with children $C_1, \ldots, C_b$, we have the axioms $\forall x : C_i(x) \Rightarrow C(x)$ expressing subsumption, as well as an axiom $\forall x : C(x) \Rightarrow C_1(x) \vee \cdots \vee C_b(x)$ expressing that the parent is covered by its children.

---

21. Of course, reducing the set of initial states does not invalidate any old plans, either.

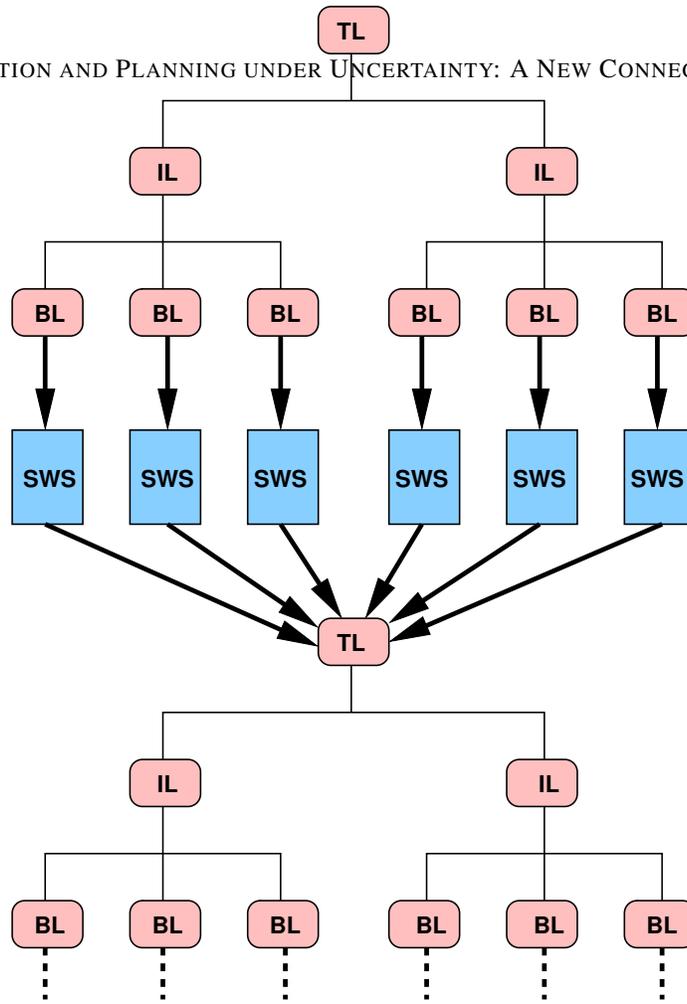

Figure 2: Schematic illustration of the **SH** scenario.

The available web services are defined as follows. For each top level concept $TL_i$, and for each leaf $BL_{i,j}$ of the corresponding tree structure, there is a web service available that takes $BL_{i,j}$ as input and that outputs $TL_{i+1}$. The corresponding $\mathcal{WSC}$ operator takes the form $o_{i,j} = (\{x\}, BL_{i,j}(x), \{y\}, TL_{i+1}(y))$. Then, by applying, for each $1 \le i < n$ in order, all services $o_{i,j}$, it is possible to make sure that a constant of concept $TL_{i+1}$ is created in all possible cases. Hence, sequencing all these steps is a plan, of length $(n-1) * b^d$. Note here that, as we already stated in Section 5.4, in our experiments groups of operators with identical effects are assigned the same output constants. For the **SH** scenario, this means that for each $1 \le i < n$, all the $o_{i,j}$ share the same output constant. Hence the total number of output constants generated, i.e., the number of "potential constants" in the initial state, is equal to the number of top-level concepts, $n$.

Although the **SH** scenario is of an abstract nature, it is representative for a variety of relevant situations. Specifically, the scenario can model situations where sets of different services must be used to address a request which none of them can handle alone. The role of each single service is then to handle some particular possible case. In our example, the "set of different services" is the set of services $o_{i,j}$ assembled for each $TL_i$. Given a constant $c$ which is a member of $TL_i$, i.e., $TL_i(c)$ holds, the "particular possible case" handled by service $o_{i,j}$ is the case where $c$ happens to be a member of leaf $BL_{i,j}$ – one of those cases must hold due to the coverage clauses in the tree.

Similar situations arise, e.g., for geographically located (regional) services when the composition request is not location-specific or addresses locations at a higher (inter-regional) level. A similar pattern can also be found in e-government scenarios where a clear-cut classification of activities leads to establishing several "parallel" services that serve different departmental areas.

Orthogonal to this "horizontal" composition, the scenario can model "vertical" composition, where one function has to be pursued by concatenating existing functions. This is the case for most complex procedures in such diverse areas as e-government or e-commerce.

The scenario can be instantiated to study different aspects of the scalability of our approach. Our empirical tests measure scalability in both the horizontal and the vertical direction. Further, we consider two extreme cases of the possible shapes of the individual concept trees in the chain, giving us instances with identical numbers of leaves. We set up the test scenario **SH-broad**, where $d = 1$ and $b$ scales over $2, 4, 8, 16, 32$. We set up the test scenario **SH-deep**, where $b = 2$ and $d$ scales over $1, 2, 3, 4, 5$. In both scenarios, $n$ scales from 2 to 20.

Further, we designed a **SH-trap** variant where a second chain of $n$ concepts can be linked, but is completely irrelevant for the goal service. This variant is suitable for testing to what extent the composition techniques are affected by irrelevant information. Finally, recall that the encoding method comes in two versions as explained above, where the default method treats input existence $- \bigwedge_{x \in X_o} exists(x) -$ by a conditional effects semantics, whereas the non-default method, **forced**, compromises completeness for efficiency by treating input existence as a forced precondition.

All in all, we have the following choices: 3 different planners (CFF-def, CFF-std, DLVK); 2 different encoding methods; **SH** with or without the trap; **SH-broad** or **SH-deep**. The cross-product of these choices yields 24 experiments, within each of which there are 19 possible values for $n$ and 5 possible values for $b$ or $d$, i.e., 95 test instances. For CFF, we measured 3 performance parameters: total runtime, number of search states inserted into the open queue, and number of actions in the plan. For DLVK, we measured total runtime and number of actions in the plan. Of course, not all of this large amount of data is interesting. In what follows, we summarize the most important observations. Figure 3 shows the data we selected for this purpose. Part (a) of the figure shows CFF-std on **SH-broad**; (b) shows CFF-std on **SH-deep**; (c) shows CFF-def on **SH-forced-broad**; (d) shows DLVK on **SH-broad** and **SH-deep**; (e) shows DLVK on **SH-forced-broad** and **SH-forced-deep**; (f) shows DLVK and CFF-std on **SH-trap**. The vertical axes all show log-scaled runtime (sec). The horizontal axes show $n$ in (a), (b) and (c). In (d), (e) and (f), $n$ is fixed to $n = 2$ and the horizontal axes show the number of leaves in the concept hierarchy.

Consider first Figure 3 (a) and (b). These plots point out how efficiently CFF can handle this kind of WSC problem, even with no **forced** optimization. Comparing the two plots points out the difference between handling broad and deep concept hierarchies. In both plots, CFF-std runtime is shown over $n$, the length of the chain to be built. In (a), we show 5 curves for the 5 different values of $b$ (the number of leaves in a hierarchy of depth 1), and in (b) we show 5 curves for the 5 different values of $d$ (the depth of a hierarchy with branching factor 2). In both cases, the scaling behavior is fairly good. With small concept hierarchies ($b = 2$ or $d = 1$), chains of almost arbitrary length can be built easily. As the hierarchies grow, runtime becomes exponentially worse. Note, however, that from one curve to the next the size of the hierarchies doubles, so that growth is itself exponential. With concept hierarchies of 16 leaves, i.e., 16 alternative cases to be handled in each step, we can still easily build chains of 6 steps, where the solution involves 96 web services. The most interesting aspect of comparing the two plots, (a) and (b), is that the underlying search spaces are actually identical: the open queues are the same. The only difference in performance stems from an overhead in CFF's reasoning techniques, which consume more runtime in the case of deep concept hierarchies. Hence the slightly worse behavior in (b).

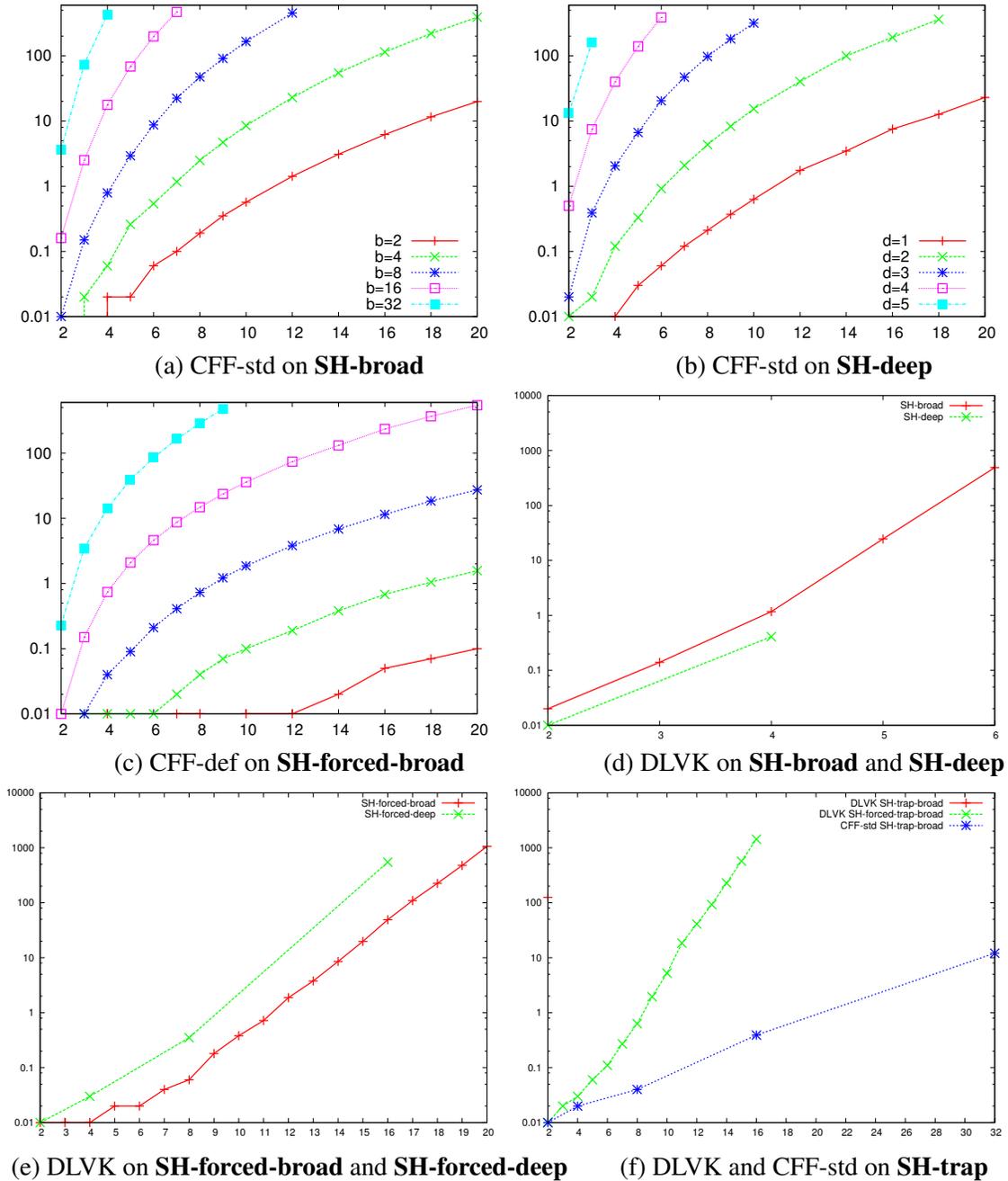

Figure 3: Selected results for **SH** scenario. See detailed explanation in text.

If we run CFF-def on the test suites of Figure 3 (a) and (b), then we obtain much worse behavior. For example, with $b = 8$ we only get up to $n = 3$. The reason seems to be that FF's helpful actions pruning and enforced hill-climbing are too greedy in this domain. A simple way to overcome this is to use a standard heuristic search algorithm instead, as done by CFF-std shown in Figure 3 (a) and (b). On the other hand, if the **forced** optimization is switched on, then helpful actions pruning and enforced hill-climbing work much better, and we obtain a significant performance boost when using CFF-def. The latter is pointed out by Figure 3 (c), showing data for CFF-def on **SH-forced-broad**. Like Figure 3 (a) for CFF-std on **SH-broad**, this plot shows 5 curves, one for each of the 5 values of $b$ (legend omitted from the plot because it would overlap the curves). We see that, in this case, we can easily build arbitrarily long chains even for $b = 16$, giving us a solution involving 320 web services for $n = 20$. Even for $b = 32$, we still get up to $n = 9$.

Figure 3 (d) and (e) show what one gets when trying to solve the same examples, encoding them directly for DLVK instead of using the compilation and solving them with CFF. As expected, the performance is much worse. Since hardly any test instance is solved for $n > 2$, we fixed $n$ to its minimum value 2 in these plots, unlike (a), (b) and (c). Each of (d) and (e) shows data for both the **broad** and **deep** variants, showing the number of leaves on the horizontal axis. In order to obtain a more fine-grained view, for the **broad** variant we increase that number by steps of 1 rather than by a multiplicative factor of 2 as before. We see that, without the **forced** optimization – Figure 3 (d) – performance is poor, and the largest case we can solve is $n = 2$, $b = 6$ where the solution involves 6 web services. As we switch **forced** on – Figure 3 (e) – performance is dramatically improved but is still on a different level than what we obtain by compilation+CFF.

Figure 3 (f), finally, exemplifies the results we get in the **trap** scenario. We show data for the **broad** version, on the default encoding with CFF-std, and on both the default and the **forced** encoding with DLVK. DLVK is quite affected by the irrelevant chain of concepts, now solving only the single instance $n = 2$, $b = 2$ for the default encoding, and getting up to $n = 2$, $b = 16$ for the **forced** encoding, instead of $n = 2$, $b = 20$ without the trap. This behavior is expected since DLVK does not make use of heuristic techniques that would be able to detect the irrelevance of the second chain of concepts. The question then is whether CFF's techniques are better at that. Figure 3 (f) shows that CFF-std is largely unaffected for $n = 2$ – one can see this by comparing that curve with the points on the vertical axis in Figure 3 (a). However, for $n > 2$ the performance of CFF-std drastically degrades: the only instances solved are $n = 3$, $b = 2$ and $n = 4$, $b = 2$. The reason seems to be that the additional actions yield a huge blow-up in the open queue used by the global heuristic search algorithm in CFF-std. Indeed, the picture is very different when using CFF-def and the **forced** encoding instead: the search spaces are then identical to those explored with no trap, and the behavior we get is identical to that shown in Figure 3 (c).

All plans found in the **SH** scenario are optimal, i.e., the plans returned contain only those web services that are needed. The single exception is DLVK in **trap**, where the solutions include some useless web services from the trap chain.[22]

---

22. Note here that DLVK's plans are parallel. Their parallel length is optimal (because we provided the correct plan length bound, cf. Section 6.1. However, each parallel step may contain unnecessary actions, on top of the necessary ones. That's what happens in **trap**.

### 6.3 Complex Concept Dependencies

The two variants of the **SH** scenario feature tightly structured relationships between the involved concepts, and allow the investigation of scalability issues by varying the size of the structure. We now consider a more advanced scenario, where the way top-level concepts are covered by lower-level concepts is subject to complex concept dependencies, similar to the axioms constraining protein classes and their characteristics in Example 1. Therefore we investigate how performance is impacted by more complex concept structures than just subsumption hierarchies.

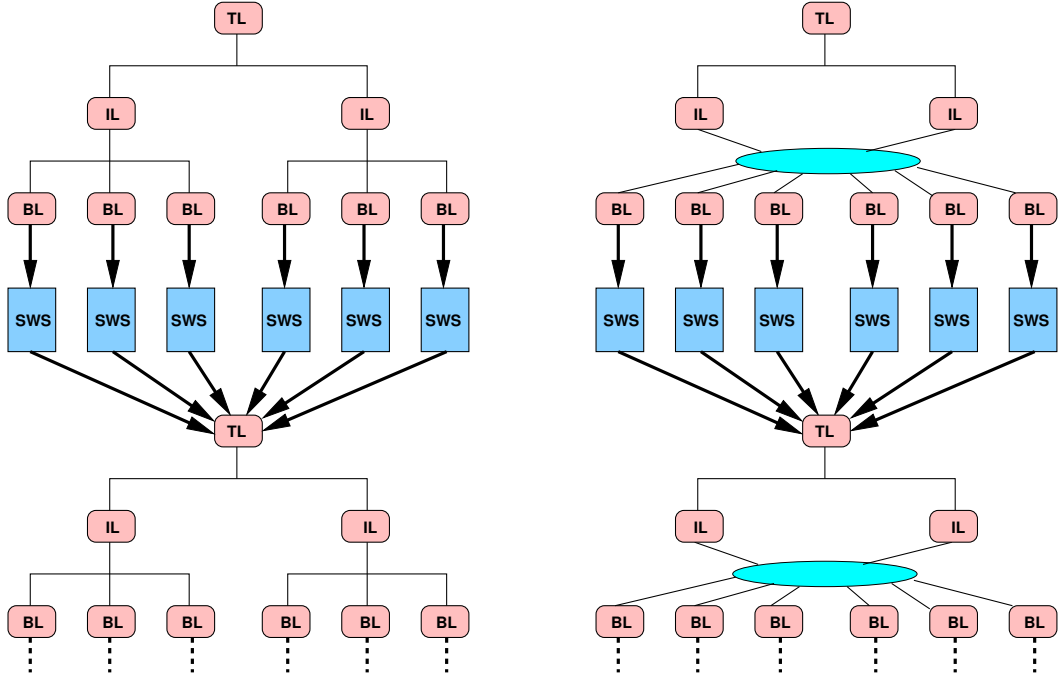

Figure 4: Schematic illustration of the **CD** scenario vs. the **SH** scenario.

Our new scenario is called **CD**, for concept dependencies. Figure 4 illustrates this scenario, and contrasts it with the **SH** scenario. Similarly to what we had in **SH**, we have top-level concepts, of which each one is associated to a set of basic sub-concepts. There are $b$ basic concepts for every top-level concept. There are $n$ top-level concepts $TL_1, \ldots, TL_n$, and the goal is to achieve $TL_n$ starting from $TL_1$. As before, this is done through combining web services that cover all possibilities. Namely, for every top-level concept $TL_i$ and for every basic concept $BL_{i,j}$ associated with it, we have the operator $o_{i,j} = ((\{x\}, BL_{i,j}(x), \{y\}, TL_{i+1}(y)).$[23]

The difference lies in the connection between the basic concepts and the top-level concepts. In **SH**, this was rigidly given in terms of a tree structure of subsumption and coverage axioms over intermediate concepts. Every basic concept – i.e., every operator $o_{i,j}$ corresponding to such a concept – had to be included in the plan in order to cover all possible cases. In **CD**, we use instead a complex set of axioms to connect the basic concepts to the top-level. Each top-level concept has $m$ intermediate concepts $IL_{i,1}, \ldots, IL_{i,m}$, for which as before we have axioms stating that each $IL_{i,j}$ is a sub-concept of $TL_i$, as well as the axiom $\forall x : TL_i(x) \Rightarrow IL_{i,1}(x) \vee \cdots \vee IL_{i,m}(x)$ stating that $TL_i$ is covered by $IL_{i,1}, \ldots, IL_{i,m}$. For the connection between the intermediate concepts and the basic concepts, complex dependencies are used. Each intermediate subconcept is constrained to be covered by some non-empty set of combinations of the basic subconcepts. Precisely, we create a random DNF, of only positive literals, using the basic concepts as the predicates. We then take that DNF to imply $IL_{i,j}$. Note here that, in the implication, the DNF is negated and hence becomes a CNF, which we can directly encode into our formalism. We do this for every $IL_{i,j}$.

---

23. Note here again that, for the same $i$, all these operators are assigned the same output constant by our compilation technique.

In such a setting, it is interesting to control *how many* combinations are required to cover the top-level concept $TL_i$. This directly corresponds to the total number of random combinations (random DNF disjuncts) that are generated, for all of the intermediate concepts $IL_{i,j}$ taken together. We control this via what we call the *coverage factor*, $c$, ranging in $(0, 1]$. From the $2^b - 1$ possible combinations of basic concepts, we pick a random subset of size $\lceil c \times (2^b - 1) \rceil$. Each such combination is associated to the DNF of a randomly chosen intermediate concept. Note that the CNF formulas generated this way may be enormous. To minimize the size of the encoding, we use the formula minimization software Espresso (Brayton, Hachtel, McMullen, & Sangiovanni-Vincentelli, 1984; McGeer, Sanghavi, Brayton, & Sangiovanni-Vincentelli, 1993).

If – hypothetically – $c$ is set to 0 then the task is unsolvable. In the experiments reported below, *whenever we write $c = 0\%$ this means that exactly one combination was selected, and associated with every intermediate concept.*

By escaping from the rigid schema of relationships presented by **SH**, the **CD** scenario is suitable to test whether the performance of our approach is tied to the specific structure of the **SH** problem. Moreover, the way **CD** is designed allows us to determine to what degree the planners react intelligently to different concept structures. In particular, the scenario allows the analysis of:

1. The ability of our approach, and in particular of the selected underlying planner CFF, to identify plans that contain only relevant actions. Especially when the "coverage factor" $c$ is low, some basic subconcepts may never appear in any partition of intermediate concepts, and thus, the plan does not need to include the respective operators. Still, due to the conditional effects/partial matches semantics, plans that include those operators are valid plans. Evaluating plan length performance over varying $c$ is therefore interesting.

2. The ability of our approach to deal with complex axiomatizations. This can be measured in terms of the impact of the coverage factor on runtime performance. The randomization of the choice of combinations of basic factors, in different settings of $c$, may induce significant differences in the CNF axiomatizations, and as a result, subject the underlying reasoning engine to very different situations.

In summary, the **CD** scenario is representative for situations where complex dependencies must be taken into account in order to select the correct services. Examples of such domains were discussed in Sections 4.2 and 5.3. In particular, the **CD** scenario corresponds closely to (a scalable version of) our protein domain example. The different values for the DSSP code correspond to different basic concepts, and the respective **getInfoDSSP** services are the operators taking them to an intermediate concept, InfoDSSP$(y)$. This is similar for amino-acids, 3-D shapes, and shapes in complexes. The top level concept combinedPresentation$(y)$ can be achieved once constants for every intermediate concept have been created. So, the only difference to **CD** lies in that, rather than having just a single top-level concept generated from its intermediates, **CD** has a sequence of top-level concepts that need to be generated in turn.

As with the **SH** scenario, the total data of our experiments is extensive, even more so since we now have 4 scenario parameters rather than 2 as before, and since individual instances now contain a random element. In Figure 5, we report selected results pointing out the main observations. Part (a)/(b) of the figure show CFF-std runtime/plan length over $n$ for $m = 4$, $b = 5$; (c)/(d) show CFF-std runtime/search nodes over $c$ for $n = 5$, $m = 3$, $b = 7$; (e) shows DLVK and CFF-std runtime over $b$ in **CD** for $n = 2$, $c = 100\%$; (f) show the latter data for CFF-def and **CD-forced**.

Figure 5 (a) and (b) consider the scalability and solution lengths of the test varying the size of the scenario, and representing different coverage factors as different lines. We report data for CFF-std. Results are very similar for **CD-forced** and CFF-def, i.e., contrary to **SH**, in **CD** this setting of options does not bring a significant performance gain. We see in Figure 5 (a) that CFF scales up pretty well, though not as well as in **SH**, being easily able to solve tasks with 7 top level concepts of which each has 4 intermediate concepts and 5 basic concepts. Tasks with minimum coverage factor, $c = 0\%$, are solved particularly effortlessly. For higher $c$ values, one can observe somewhat of an easy-hard-easy pattern, where, for example, the curve for $c = 100\%$ lies significantly below the curves for $c = 40\%$ and $c = 60\%$. We examine this easy-hard-easy pattern in more detail below.

In Figure 5 (b), an obvious and expected observation is that plan length grows linearly with $n$, i.e., with the number of top level concepts. A likewise obvious, but much more important, observation is that *plan length grows monotonically with the coverage factor $c$.* As reported above, a lower coverage factor opens up the opportunity to employ less basic services, namely only the relevant ones. Figure 5 (b) clearly shows that CFF-std is effective at determining which of the services are relevant and which are not.

Let us get back to the intriguing observation from Figure 5 (a), the easy-hard-easy pattern over growing $c$. Figure 5 (c) and (d) examine this phenomenon in more detail. Both plots scale $c$ on the horizontal axis, for a fixed setting of $n$, $m$ and $b$. Runtime is shown in (c), while (d) shows the number of search states inserted into the open queue. For each value of $c$, the plots give the average and standard deviation of the results for 30 randomized instances. We clearly see the easy-hard-easy pattern in (c) for runtime, with high variance particularly for $c = 80\%$. In (d), we see that there is no such pattern for the number of search states, and that the variance is much less pronounced. This shows that the easy-hard-easy pattern is *not* due to differences in the actual search performed by CFF, but due to the effort spent in the search nodes. We traced the behavior of CFF in detail, and found that the reason for the easy-hard-easy pattern lies in the runtime CFF spends in its SAT reasoning for "state transitions", i.e., in the reasoning it uses to determine which facts are definitely true/false in each belief. For high but non-100 values of $c$, the CNF encodings of the concept dependency structures take on a rather complex form. In the cases where CFF takes a lot of runtime, almost all of the runtime is spent *within a single call to the SAT solver.* That is, it seems that CFF's SAT solver exhibits a kind of heavy-tailed behavior on these formulas, a phenomenon well known in the SAT and CP community, see for example the work of Gomes, Selman, Crato, and Kautz (2000). It should be noted here that, in typical planning benchmarks, the CNFs have a much simpler structure, which motivates the use of a fairly naive SAT solver in CFF, using neither clause learning nor restarts, in order to save overhead on formulas that are simple anyway. It seems likely that the addition of advanced SAT techniques to the solver could ameliorate the observed problem.

Finally, Figure 5 (e) and (f) compare the performances of compilation+CFF and DLVK (with no compilation). Both plots fix $n = 2$, i.e., data is shown for only 2 top level concepts. The only instances that DLVK solves for $n > 2$ are the ones where the **forced** optimization is used and $n = 3$, $m = 2$, $b = 2$. Further, in both plots $c$ is fixed to $c = 100\%$. The reason for this is that we did not find a significant difference in the performance of DLVK for different values of $c$. DLVK was unable to exploit lower $c$ for lower runtime, and neither did it show an easy-hard-easy pattern. We speculate that DLVK's answer set programming solver tends to perform exhaustive search anyway and is accordingly not as affected by different structures as the heuristic techniques employed by CFF. However, like CFF, DLVK was able to exploit lower coverage factors $c$ for shorter plans.

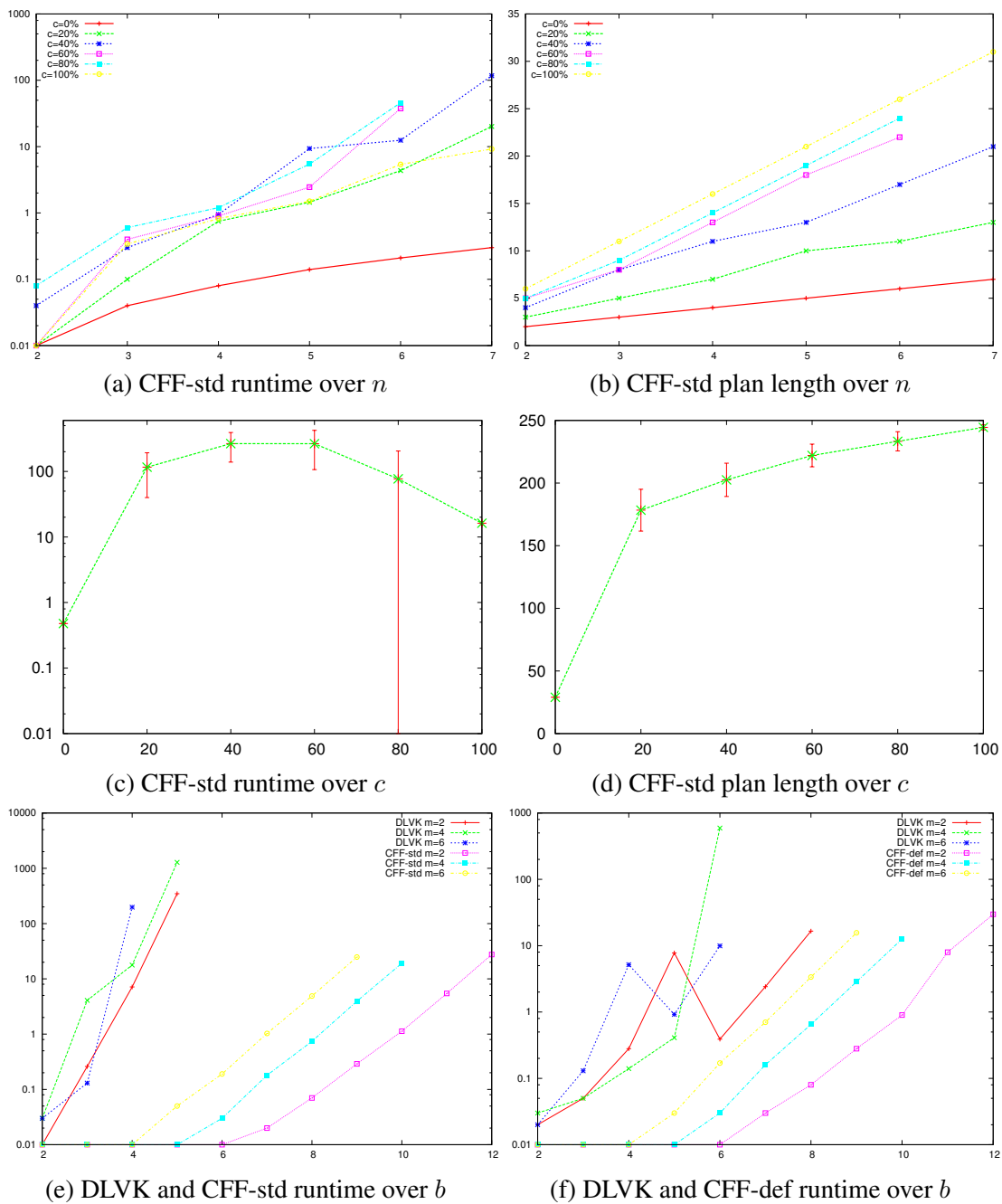

(a) CFF-std runtime over $n$

(b) CFF-std plan length over $n$

(c) CFF-std runtime over $c$

(d) CFF-std plan length over $c$

(e) DLVK and CFF-std runtime over $b$

(f) DLVK and CFF-def runtime over $b$

Figure 5: Selected results for **CD** scenario. See detailed explanation in text.

Figure 5 (e) shows the default setting without the **forced** optimization. We see that the performance of DLVK explodes quickly while CFF does not experience as much trouble. CFF fails at the upper ends of its curves, both in Figure 5 (e) and (f), only because the problem files, i.e., the CNFs describing the complex concept dependencies, become too large to parse ($> 4$ MB). That notwithstanding, CFF's runtime behavior is clearly exponential. Note, however, that the actual encodings, i.e., the problem instances to be solved, also grow exponentially over $c$.

We can further observe that DLVK exhibits quite some variance, particularly across different settings of $m$: the curves cross in Figure 5 (e). This is even more pronounced in Figure 5 (f), where we can also observe, as before for **SH**, that the **forced** optimization brings a huge advantage for DLVK. For $m = 2$ and $m = 6$ in Figure 5 (f), DLVK fails on the first unsolved problem instance due to running out of memory shortly after parsing the problem.

Concluding this section, we observe that the empirical behavior of CFF in the **SH** and **CD** scenarios is promising. These results should not be over-interpreted, though. While the test scenarios do capture problem structure typical of a variety of potential applications of WSC technology, our approach has yet to be put to the test of actual practice. The same, however, can be said of essentially all current planning-based WSC technology, since the field as a whole is still rather immature.

## 7. Related Work

The relation of our work to the belief update literature has been covered in detail already in Sections 2.2 and 4.3. As for the relation to planning, our formalism basically follows all the commonly used frameworks. Our notions of operators, actions, and conditional effects are exactly as used in the PDDL framework (McDermott et al., 1998; Bacchus, 2000; Fox & Long, 2003), except for the extension with outputs. Regarding the latter, it has been recognized for some time in the planning community, for example by Golden (2002, 2003) and Edelkamp (2003), that on-the-fly creation of constants is a relevant feature for certain kinds of planning problems. However, attempts to actually address this feature in planning tools are scarce. In fact the only attempt we are aware of is the work by Golden (2002, 2003) and Golden, Pand, Nemani, and Votava (2003). Part of the reason for this situation is probably that almost all current state of the art tools employ pre-processing procedures that compile the PDDL task into a fully grounded representation. The core algorithms are then implemented based on a propositional representation. Lifting such algorithms to a representation that involves variables and on-the-fly instantiations requires a major (implementation) effort. In the work herein, we circumvent that effort by using "potential" constants and feeding the resulting problem to CFF, which like most planners employs the said pre-processing. Extending CFF for $\mathcal{WSC}|_{fwd}$ will involve dealing with non-propositional representations as a sub-problem.

Our notion of initial state uncertainty and conformant plans closely follows the related literature from planning under uncertainty (Smith & Weld, 1998; Cimatti et al., 2004; Hoffmann & Brafman, 2006). The formalization in terms of beliefs is adapted from the work by Bonet and Geffner (2000). There are some related works in planning which allow a domain axiomatization, i.e., some form of axioms constraining the possible world states (Eiter et al., 2003; Giunchiglia et al., 2004). To the best of our knowledge, no work in planning exists, apart from the work presented herein, which considers the *combination* of domain axioms and outputs.

A few words are in order regarding our notions of "partial" and "plug-in" matches. This terminology originates from work on *service discovery* in the SWS community (see for example Paolucci et al., 2002; Li & Horrocks, 2003; Kumar et al., 2007). In service discovery, one is concerned with matching service advertisements against service requests. The discovery result is the set of services whose advertisement matches the request. The descriptions of services and requests are similar to the functional-level service descriptions, i.e., the planning operators that we use here. However, the terminology in these works is slightly different from ours, and they also describe additional kinds of matches. The notions given by Li and Horrocks (2003) have the closest relation to ours. Service descriptions are defined in terms of constructed Description Logic concepts. Say $A$ is the concept describing the advertisement, and $R$ is the concept describing the request. Then Li and Horrocks say that A and R have: an "exact match" if $A \equiv R$; a "plug-in match" if $A \sqsupseteq R$; a "subsume match" if $A \sqsubseteq R$; and an "intersection match" if $A \sqcap R \not\sqsubseteq \bot$. To compare this to our setting, consider the situation where $A$ is the effect of action $a$, and $R$ is the precondition of action $r$. Exact matches are a special case of plug-in matches which we do not distinguish herein. Intersection matches correspond to what we call partial matches. Concerning plug-in and subsume matches, matters are more subtle. The intuitive meaning of "plug-in match" is that "the advertisement fully suffices to fulfill the request". In planning terms, this means that the effect of $a$ implies the precondition of $r$. However, in service discovery this is traditionally taken to mean that every requested entity is being provided, i.e., $A \sqsupseteq R$. The latter notion – where the precondition of $r$ implies the effect of $a$ – is not meaningful in planning. Hence we use only one of the two notions, in correspondence to Li and Horrocks's "subsume matches".

In contrast to the work of Li and Horrocks (2003), and to our work, Paolucci et al. (2002) and Kumar et al. (2007) define matches for individual input/output parameters in service descriptions, rather than for service descriptions on a more global level (precondition/effect for us, constructed concept for Li & Horrocks, 2003). On the level of individual parameters, Paolucci et al. (2002) suggest the same notions as Li and Horrocks (2003) except that they do it in a less formal notation, and they do not define intersection matches. The same is true of Kumar et al. (2007). The latter authors also define notions of "contains" and "part-of" matches, relating to the building blocks of constructed concepts. Obviously, such notions do not make sense in our framework, where there aren't any constructed concepts. Finally, Kumar et al. define some ways of aggregating matches for individual parameters to matches for entire service descriptions. Again, this is not applicable in our case since we work on a more global level in the first place.

A brief survey of the existing works on WSC is as follows. There is a variety of works that compile composition into more or less standard deterministic planning formalisms (Ponnekanti & Fox, 2002; Srivastava, 2002; Sheshagiri et al., 2003). Some other works (Agarwal, Dasgupta, Karnik, Kumar, Kundu, Mittal, & Srivastava, 2005b; Agarwal et al., 2005a) additionally focus on end-to-end integration of SWS composition in the larger context. Akkiraju, Srivastava, Anca-Andreea, Goodwin, and Syeda-Mahmood (2006) investigate techniques to disambiguate concept names. McIlraith and Fadel (2002) achieve composition with particular forms of non-atomic services, by modeling the latter as atomic actions that take the meaning of a kind of macro-actions. Narayanan and McIlraith (2002) obtain a composition ability as a side-effect of verifying SWS properties using Petri Nets. Kuter, Sirin, Nau, Parsia, and Hendler (2005), Au, Kuter, and Nau (2005), and Au and Nau (2006) focus on information gathering at composition time, rather than at plan execution time. McDermott (2002) treats the actual interaction (communication) with a web service as a planning problem.

Mediratta and Srivastava (2006) design an approach to WSC based on conditional planning, i.e., a form of planning under uncertainty. While this suggests a close relation to our work, the focus of Mediratta and Srivastava's work is actually quite different from ours. Mediratta and Srivastava do not consider output variables, and neither do they consider any domain axiomatizations. The only overlap with our formalism lies in that they allow incomplete initial state descriptions, i.e., initial states that assign a value to only a subset of the propositions. They handle observation actions which allow observing the value of any unspecified proposition. To ameliorate the need for complete modeling, they consider a definition of "user acceptable" plans, where only a subset of the plan branches, as specified by the user, are guaranteed to lead to the goal. The latter may be an interesting option to look into when extending our framework to handle partial observability.

Two approaches explore how to adapt formalisms from so-called "hand-tailored planning" for SWS composition. The approaches are based on Golog (McIlraith & Son, 2002) and HTN planning (Sirin et al., 2004), respectively. These frameworks enable the human user to provide control information. However, non-deterministic action choice is allowed. If no control information is given, then planning is fully automatic. Hence, in this sense, these frameworks are strictly more powerful than planning without such control information. Further, both approaches are capable of handling advanced plan constructs such as loops and branches. In Golog, the possible plans – the possible composition solutions – are described in a kind of logic where high-level instructions are given by the programmer, and the planner will bind these instructions to concrete actions as part of the execution. In HTN, the programmer supplies the planning algorithm with a set of so-called "decomposition methods", specifying how a certain task can be accomplished in terms of a combination of sub-tasks. Recursively, there are decomposition methods for those sub-tasks. Thus the overall task can be decomposed in a step-wise fashion, until atomic actions are reached. Neither McIlraith and Son (2002) nor Sirin et al. (2004) are concerned with handling ontology axioms, as we do in this paper. Hence, combining the insights of both directions has synergetic potential, and is an interesting topic for future work.

Another approach capable of handling advanced plan constructs (loops, branches) is described by Pistore et al. (2005b), Pistore, Traverso, Bertoli, and Marconi (2005c), Pistore et al. (2005a), and Bertoli, Pistore, and Traverso (2006). In this work, "process level" composition is implemented, as opposed to the profile/capability level composition as addressed in this paper. At the process level, the semantic descriptions detail precisely how to interact with the SWS, rather than characterizing them only in terms of preconditions and effects. Pistore et al. (2005b, 2005c, 2005a) and Bertoli et al. (2006) exploit BDD (Binary Decision Diagram) based search techniques to obtain complex solutions fully automatically. However, ontology axioms are not handled and input/output types are matched based on type names.

There are only a few approaches where ontology axioms are used and the requirements on the matches are relaxed. One of those is described by Sirin, Hendler, and Parsia (2003), Sirin, Parsia, and Hendler (2004), Sirin and Parsia (2004), and Sirin et al. (2006). In the first two papers of this series (Sirin et al., 2003, 2004), a SWS composition support tool for human programmers is proposed: at any stage during the composition process, the tool provides the user with a list of matching services. The matches are found by examining the subconcept relation. An output $A$ is considered a match of input $B$ if $A \subseteq B$. This corresponds to plug-in matches. In later work (Sirin & Parsia, 2004; Sirin et al., 2006), the HTN approach (Sirin et al., 2004) mentioned above is adapted to not work on the standard planning semantics, but on the description logics semantics of OWL-S. The difficulties inherent in updating a belief are observed, but the connection to belief update as studied in the literature is not made, and it remains unclear which solution is adopted.

As far as we are aware, all other methods with more relaxed matches follow what we have here termed a message-based approach to WSC. These approaches were already discussed in some depth in Section 2.3. Next, we give a few more details on the ones most closely related to our work. The approach by Liu et al. (2007) was discussed in sufficient detail already in Section 2.3, so we do not reconsider this here.

Meyer and Weske (2006) handle ontology axioms in their WSC tool, but do not provide a semantics for action applications. Reasoning is only used to determine whether a particular output can be used to establish a particular input, so the approach can be classified as "message-based", in our terms. The kind of matches handled is said to be plug-in. To the best of our knowledge, this tool is the only existing WSC tool that employs a relaxed plan based heuristic function, like CFF. However, through various design decisions, the authors sacrifice scalability. They explicitly enumerate all world states in every belief, and hence suffer from exponentially large beliefs. They search forward with parallel actions and consequently suffer from a huge branching factor. They take their heuristic to be relaxed planning graph length (rather than relaxed plan length) and thus suffer from the fact that, most of the time, $h^{max}$ is a much less informative heuristic than $h^+$ (Bonet & Geffner, 2001; Hoffmann, 2005).

An approach rather closely related to ours, in that it can handle partial matches, is described by Constantinescu and Faltings (2003) and Constantinescu et al. (2004a, 2004b). In this work the ontology is assumed to take the form of a tree of concepts, where edges indicate the subconcept relation. Such a tree is compiled into intervals, where each interval represents a concept and the contents are arranged to correspond to the tree. The intervals are used for efficient implementation of indexing in service lookup (discovery), as well as for matching during composition. The latter searches forward in a space of "switches". Starting at the initial input, if the current input is of type $A$, then a service with input $A_i$ matches if $A \cap A_i \neq \emptyset$. Such services are collected until the set of the collected $A_i$ covers $A$ (that is, until the union of the intervals for the various $A_i$ contains the interval for $A$). The collected services form a switch, and in the next step of the search, each of their outputs becomes a new input that must be treated (i.e., the switch is an AND node). Composition is interleaved with discovery, i.e., in every search state discovery is called to find the services that match this state. The search proceeds in a depth-first fashion. Major differences to our work are the following. First, the formalization is very different, using intervals vs. using standard notions from planning based on logics. Second, the approach interleaves discovery and composition, which are separate steps in our framework (web service discovery is needed to determine the "operators" of a $\mathcal{WSC}$ task). Third, the approach considers concept trees vs. clausal integrity constraints. Last, the approach uses depth-first search, whereas one of the main points we are making is that one can exploit the heuristic techniques implemented in standard planning tools for scalable WSC.

Finally, an interesting approach related to planning is described by Ambite and Kapoor (2007). To capture the dependencies between different input variables of a web service, the input is described in terms of a relation between those variables. The same is done for the outputs. The relations are formulated in terms of logical formulas relative to an ontology. The underlying formalism is first-order logic, so the modeling language is quite expressive.[24] Reasoning is performed in order to establish links ("messages", in our terms) between inputs and outputs. The algorithmic framework in which that happens is inspired by partial-order planning (Penberthy & Weld, 1992), starting from the goal relation and maintaining a set of open links. The solution is a DAG of web services where links correspond to different kinds of data exchanges (selection, projection, join, union). Automatic insertion of mediator services, e.g., for converting a set of standard formats, is also supported.

---

24. At the cost of undecidable reasoning, which according to the authors is not a major issue in practice.

To some extent, our preconditions/effects and clausal integrity constraints can be used to model "relations" in the sense of Ambite and Kapoor (2007). Say $r$ is a $k$-ary relation with definition $\phi$, describing the input of a web service. We set the corresponding operator's precondition to $r(x_1, \ldots, x_k)$, and we transform $\phi$ into a set of universally quantified clauses. As long as the latter can be done, and as long as the ontology axioms can be likewise transformed, we obtain a model equivalent to that of Ambite and Kapoor. In that sense, the main modeling advantage of the approach of Ambite and Kapoor over $\mathcal{WSC}|_{fwd}$ is existential quantification. It is an open question whether such quantification can be accommodated in our framework. Insertion of mediator services can be supported in $\mathcal{WSC}|_{fwd}$, but only in the limited sense of recognizing, via particular preconditions, that a particular kind of mediator is required. Modeling the actual data flow is bound to be awkward. In summary, the work of Ambite and Kapoor is more advanced than ours from a data description and transformation point of view. On the other hand, Ambite and Kapoor neither consider belief update, nor do they place their work in the context of a fully-fledged planning formalism, and they are less concerned with exploiting the heuristic technologies of recent planners. Combining the virtues of both approaches – within either framework – is an interesting direction for further research.

## 8. Discussion

We have suggested a natural planning formalism for a significant notion of web service composition at the profile / capability level, incorporating on-the-fly creation of constants to model outputs, incomplete initial states to model incomplete user input, conditional effects semantics to model partial matches, and, most importantly, clausal integrity constraints to model ontology axioms. We have identified an interesting special case, forward effects, where the semantics of action applications is simpler than in the general case. We have demonstrated how this relates to the belief update literature, and we have shown how it results in reduced computational complexity. Forward effects relate closely to message-based WSC, and our results serve both to put this form of WSC into context, and to extend it towards a more general notion of partial matches. Further, we have identified a compilation into planning under (initial state) uncertainty, opening up an interesting new connection between the planning and WSC areas.

Our empirical results are encouraging, but should not be over-interpreted. While our test scenarios serve to capture some structural properties that are likely to appear in applications of WSC technology, our approach has yet to be put to the test of actual practice. The same, however, can be said of essentially all current planning-based WSC technology, since that field is still rather immature. In that sense, a more thorough evaluation of our approach, and of planning-based WSC as a whole, is a challenge for the future.

Apart from such evaluation, there are several directions for research improving and extending the technology introduced herein. A line of research that we find particularly interesting is to adapt modern planning tools for WSC, starting from our special cases, where the complications incurred by integrity constraints are more manageable. We have already outlined a few ideas for adapting CFF, and pointed out that new challenges arise. It appears particularly promising to tailor generic heuristic functions, originating in planning, to exploit the typical forms of ontology axioms as occur in practice. Considering the wealth of heuristic functions available by now, this topic alone provides material for a whole family of subsequent work.

## Acknowledgments

We thank the anonymous reviewers, as well as the managing editor Derek Long, for their comments, which were of significant help for improving the paper.

Jörg Hoffmann performed part of this work while being employed at the University of Innsbruck, Austria. His work was partly funded through the European Union's 6th Framework Programme under the SUPER project (IST FP6-026850, http://www.ip-super.org).

Piergiorgio Bertoli's and Marco Pistore's work was partly supported by the project "Software Methodology and Technology for Peer-to-Peer Systems" (STAMPS).

Malte Helmert's work was partly supported by the German Research Council (DFG) as part of the Transregional Collaborative Research Center "Automatic Verification and Analysis of Complex Systems" (SFB/TR 14 AVACS). See www.avacs.org for more information.

## Appendix A. Proofs

We first formally prove Proposition 1, stating that negative effects can be compiled away in $\mathcal{WSC}$. Before we do so, we first need to introduce the compilation formally. Assume a $\mathcal{WSC}$ task $(\mathcal{P}, \Phi_{IC}, \mathcal{O}, C_0, \phi_0, \phi_G)$. We construct a second $\mathcal{WSC}$ task $(\mathcal{P}^+, \Phi_{IC}^+, \mathcal{O}^+, C_0, \phi_0, \phi_G)$, where initially $\mathcal{P}^+, \Phi_{IC}^+$ and $\mathcal{O}^+$ are the same as $\mathcal{P}, \Phi_{IC}$ and $\mathcal{O}$, respectively. We proceed as follows. Let $G \in \mathcal{P}$ be a predicate with arity $k$, so that there exists $o \in \mathcal{O}$, $o = (X_o, \text{pre}_o, Y_o, \text{eff}_o)$ where $\text{eff}_o$ contains a negative literal $\neg G(x_1, \ldots, x_k)$. We introduce a new predicate $notG$ into $\mathcal{P}^+$, and we introduce the two new clauses $\forall x_1, \ldots, x_k : G(x_1, \ldots, x_k) \vee notG(x_1, \ldots, x_k)$ and $\forall x_1, \ldots, x_k : \neg G(x_1, \ldots, x_k) \vee \neg notG(x_1, \ldots, x_k)$. For every operator $o$ whose effect contains a negation of $G$, we replace, in $\text{eff}_o$, $\neg G(a_1, \ldots, a_k)$ with $notG(a_1, \ldots, a_k)$.[25] We continue doing so until no negative effect literals remain in $\mathcal{O}^+$.

If $a$ is an action in $(\mathcal{P}, \Phi_{IC}, \mathcal{O}, C_0, \phi_0, \phi_G)$ then we denote by $a^+$ the corresponding action in $(\mathcal{P}^+, \Phi_{IC}^+, \mathcal{O}^+, C_0, \phi_0, \phi_G)$. We also use this notation vice versa, i.e., if $a^+$ is an action in $(\mathcal{P}^+, \Phi_{IC}^+, \mathcal{O}^+, C_0, \phi_0, \phi_G)$ then $a$ denotes the corresponding action in $(\mathcal{P}, \Phi_{IC}, \mathcal{O}, C_0, \phi_0, \phi_G)$. If $s = (C_s, I_s)$ is a state using the predicates $\mathcal{P}$, then we denote by $s^+$ a state using the predicates $\mathcal{P}^+$, with the following properties: $C_{s^+} = C_s$; for all $p \in \mathcal{P}^{C_s}$ we have $I_{s^+}(p) = I_s(p)$; for all $notp$ where $p \in \mathcal{P}^{C_s}$ we have $I_{s^+}(notp) = 1$ iff $I_s(p) = 0$. Since there is, obviously, exactly one such $s^+$, we will also use this correspondence vice versa.

**Proposition 1** *Assume a $\mathcal{WSC}$ task $(\mathcal{P}, \Phi_{IC}, \mathcal{O}, C_0, \phi_0, \phi_G)$. Let $(\mathcal{P}^+, \Phi_{IC}'^+, \mathcal{O}^+, C_0, \phi_0, \phi_G)$ be the same task but with negative effects compiled away. Assume an action sequence $\langle a_1, \ldots, a_n \rangle$. Let $b$ be the result of executing $\langle a_1, \ldots, a_n \rangle$ in $(\mathcal{P}, \Phi_{IC}, \mathcal{O}, C_0, \phi_0, \phi_G)$, and $b^+$ is the result of executing $\langle a_1^+, \ldots, a_n^+ \rangle$ in $(\mathcal{P}^+, \Phi_{IC}^+, \mathcal{O}^+, C_0, \phi_0, \phi_G)$. Then, for any state $s$, we have that $s \in b$ iff $s^+ \in b^+$.*

**Proof:** By induction over the length of the action sequence in question. If the sequence is empty, then we have to consider the initial beliefs of the two tasks, for which the claim follows directly by definition. For the inductive step, say that the claim holds for $b$ and $b^+$, and $a$ is an action. We need to show that, for any state $s$, we have that $s \in res(b, a)$ iff $s^+ \in res(b^+, a^+)$.

For the direction from right to left, say $s^+ \in res(b^+, a^+)$. By definition we have $s^+ \in res(s_0^+, a^+)$ for a state $s_0^+ \in b^+$. By induction hypothesis, $s_0 \in b$. It therefore suffices to show that $s \in res(s_0, a)$. We need to show that (1) $s \models \Phi_{IC} \wedge \text{eff}_a$ and (2) $s$ differs from $s_0$ in a set-inclusion minimal set of values. (1) is obvious from the definitions. Assume to the contrary of (2) that there exists $s_1$ so that $s_1 \models \Phi_{IC} \wedge \text{eff}_a$ and $s_1$ is identical to $s$ except that there exists at least one proposition $p$ where $s_1(p) = s_0(p)$ but $s(p) \neq s_0(p)$. By definition, we get that $s_1^+ \models \Phi_{IC}^+ \wedge \text{eff}_{a^+}$. Further, we get that $s_1^+(p) = s_0^+(p)$ but $s^+(p) \neq s_0^+(p)$, and altogether that $s_1^+ <_{s_0^+} s^+$. This is a contradiction to $s^+ \in res(s^+, a^+)$, and hence proves that $s \in res(s_0, a)$ as desired.

---

25. The arguments $a_i$ here may be either variables or constants.

The direction from left to right proceeds in the same fashion. Say $s \in res(b, a)$. By definition we have $s \in res(s_0, a)$ for a state $s_0 \in b$. By induction hypothesis, $s_0^+ \in b^+$. It then suffices to show that $s^+ \in res(s_0^+, a^+)$. We need to show that (1) $s^+ \models \Phi_{IC}^+ \wedge \text{eff}_a$ and (2) $s^+$ differs from $s_0^+$ in a set-inclusion minimal set of values. (1) is obvious from the definitions. Assume to the contrary of (2) that there exists $s_1^+$ so that $s_1^+ \models \Phi_{IC}^+ \wedge \text{eff}_{a^+}$ and $s_1^+$ is identical to $s^+$ except that there exists at least one proposition $p$ where $s_1^+(p) = s_0^+(p)$ but $s^+(p) \neq s_0^+(p)$. By definition, we get that $s_1 \models \Phi_{IC} \wedge \text{eff}_a$. Further, if $p \in \mathcal{P}^{C_{s_0}}$ then we get that $s_1(p) = s_0(p)$ but $s(p) \neq s_0(p)$. If $p = notq \notin \mathcal{P}^{C_{s_0}}$ then we get the same property for $q$. Altogether, we get that $s_1 <_{s_0} s$. This is a contradiction to $s \in res(s, a)$, and hence proves that $s^+ \in res(s_0^+, a^+)$ as desired. $\qquad \square$

**Theorem 1** *Assume a $\mathcal{WSC}$ task with fixed arity, and a sequence $\langle a_1, \dots, a_n \rangle$ of actions. It is $\Pi_2^p$-complete to decide whether $\langle a_1, \dots, a_n \rangle$ is a plan.*

**Proof:** Membership is proved by a guess-and-check argument. First, observe that, for arbitrary $s$, $s'$, and $A$, we can decide within coNP whether $s' \in res(s, A)$. Guess a state $s''$ where $C_{s''} = C_s \cup E_a$. Check whether $s'' \models \Phi_{IC} \wedge \text{eff}_a$. Check whether $I_{s'} \not\leq_s I_{s''}$. Then $s' \in res(s, a)$ iff no guess succeeds. Further, for an action $a$, deciding whether $a$ is inconsistent is, obviously, equivalent to a satisfiability test, so this is contained in NP. With these instruments at hand, we can design a guess-and-check procedure to decide whether $\langle a_1, \dots, a_n \rangle$ is a plan. We guess the proposition values along $\langle a_1, \dots, a_n \rangle$. We then check whether these values comply with $res$, and lead to an inconsistent action, or to a final state that does not satisfy the goal. In detail, the checking proceeds as follows. First, check whether the initial proposition values satisfy $\Phi_{IC} \wedge \phi_0$. If not, stop without success. Otherwise, iteratively consider each action $a_i$, with pre-state $s$ and post-state $s'$. Check with an NP oracle whether $a$ is inconsistent. If yes, stop with success. If not, test with an NP oracle whether $s' \in res(s, a)$. If not, stop without success. Otherwise, if $i < n$, then go on to $a_{i+1}$. If $i = n$, then test whether $s' \models \phi_G$. Stop with success if $s' \not\models \phi_G$, stop without success if $s' \models \phi_G$. $\langle a_1, \dots, a_n \rangle$ is a plan iff no guess of proposition values is successful.

Hardness follows by the following adaptation of the proof of Lemma 6.2 from Eiter and Gottlob (1992). Validity of a QBF formula $\forall X . \exists Y . \phi[X, Y]$, where $\phi$ is in CNF, is reduced to plan testing for a single action $a$. We use the 0-ary predicates $X = \{x_1, \dots, x_m\}$, $Y = \{y_1, \dots, y_n\}$, and new 0-ary predicates $\{z_1, \dots, z_m, r, t\}$. The set of operators contains the single operator $o$ with empty in/out parameters, empty precondition, and effect $t$. The initial constants are empty; $\phi_0$ is the conjunction of all $x_i$, all $y_i$, all $z_i$, $r$, and $\neg t$; $\phi_G$ is $r$. The theory is:

$$(\bigwedge_{i=1}^{m} (\neg t \vee x_i \vee z_i)) \wedge (\bigwedge_{i=1}^{m} (\neg t \vee \neg x_i \vee \neg z_i)) \wedge (\bigwedge_{C \in \phi} (\neg t \vee \neg r \vee C)) \wedge (\bigwedge_{i=1}^{n} (\neg t \vee \neg y_i \vee r))$$

where $\phi$ is viewed as a set of clauses $C$. More readably, the theory is equivalent to:

$$t \Rightarrow [(\bigwedge_{i=1}^{m} x_i \equiv \neg z_i) \wedge (r \Rightarrow \phi) \wedge ((\bigvee_{i=1}^{n} y_i) \Rightarrow r)]$$

We refer to the initial belief as $b$. The plan to test contains the single action $a$ based on (equal to, in fact) $o$. We refer to the resulting belief as $b'$. Obviously, $b$ contains a single state $s$ where everything except $t$ is true. Also, $a$ is consistent: any interpretation that sets $r$ and all $y_i$ to 0 satisfies $\Phi_{IC} \wedge \mathrm{eff}_a$.

The theory conjuncts $x_i \equiv \neg z_i$ make sure that each $w \in b'$ makes exactly one of $x_i, z_i$ true. In particular, the different assignments to $X$ are incomparable with respect to set inclusion. Hence, we have that for every assignment $a_X$ of truth values to $X$, there exists a state $s' \in b'$ that complies with $a_X$: $a_X$ is satisfiable together with $\Phi_{IC} \wedge \mathrm{eff}_a$, and any other assignment $a'_X$ is more distant from $s$ in at least one variable (e.g., if $a'_X(x_i) = 1$ and $a_X(x_i) = 0$ then $a_X$ is closer to $s$ than $a'_X$ regarding the interpretation of $z_i$).

We now prove that, if $a$ is a plan, then $\forall X. \exists Y. \phi[X, Y]$ is valid. Let $a_X$ be a truth value assignment to $X$. With the above, we have a state $s' \in b'$ that complies with $a_X$. Since $a$ is a plan, we have $s' \models r$. Therefore, due to the theory conjunct $r \Rightarrow \phi$, we have $s' \models \phi$. Obviously, the values assigned to $Y$ by $s'$ satisfy $\phi$ for $a_X$.

For the other direction, say $\forall X. \exists Y. \phi[X, Y]$ is valid. Assume that, contrary to the claim, $a$ is not a plan. Then we have $s' \in b'$ so that $s' \not\models r$. But then, due to the theory conjunct $(\bigvee_{i=1}^{n} y_i) \Rightarrow r$, we have that $s$ sets all $y_i$ to false. Now, because $\forall X. \exists Y. \phi[X, Y]$ is valid, there exists a truth value assignment $a_Y$ to $Y$ that complies with the setting of all $x_i$ and $z_i$ in $s$. Obtain $s''$ by modifying $s'$ to comply with $a_Y$, and setting $r$ to 1. We have that $s'' \models \Phi_{IC} \wedge \mathrm{eff}_a$. But then, $s''$ is closer to $s$ than $s'$, and hence $s' \notin b'$ in contradiction. This concludes the argument. $\qquad \square$

**Theorem 2.** *Assume a $\mathcal{WSC}$ task with fixed arity, and a natural number $b$ in unary representation. It is $\Sigma_3^p$-complete to decide whether there exists a plan of length at most $b$.*

**Proof:** For membership, guess a sequences of actions containing at most $b$ actions (note that the size of such a sequence is polynomial in the size of the input representation). By Theorem 1, we can check with a $\Pi_2^p$ oracle whether the sequence is a plan.

Hardness follows by an extension of the proof of Lemma 6.2 of Eiter and Gottlob (1992). Validity of a QBF formula $\exists X. \forall Y. \exists Z. \phi[X, Y, Z]$, where $\phi$ is in CNF, is reduced to testing plan existence. We use the 0-ary predicates $X = \{x_1, \ldots, x_n\}$, $Y = \{y_1, \ldots, y_m\}$, $Z = \{z_1, \ldots, z_k\}$, and new 0-ary predicates $\{q_1, \ldots, q_m, r, t, f_1, \ldots f_n, h, g\}$. The set of operators is composed of:

- $o^t := (\emptyset, f_1 \wedge \cdots \wedge f_n \wedge h, \emptyset, t \wedge g \wedge \neg h)$

- For $1 \leq i \leq n$: $o^{x_i} := (\emptyset, h, \emptyset, x_i \wedge f_i)$

- For $1 \leq i \leq n$: $o^{\neg x_i} := (\emptyset, h, \emptyset, \neg x_i \wedge f_i)$

The initial constants are empty. The initial literal conjunction $\phi_0$ is composed of all $y_i$, all $z_i$, all $q_i$, $r$, $\neg t$, all $\neg f_i$, $h$, and $\neg g$. That is, the $y_i$, $z_i$, and $q_i$ as well as $r$ and $h$ are true, while the $f_i$ as well as $t$ and $g$ are false. No value is specified (only) for the $x_i$. The goal $\phi_G$ is $r \wedge g$. The theory is:

$$(\bigwedge_{i=1}^{m} (\neg t \vee y_i \vee q_i)) \wedge (\bigwedge_{i=1}^{m} (\neg t \vee \neg y_i \vee \neg q_i)) \wedge (\bigwedge_{C \in \phi} (\neg t \vee \neg r \vee C)) \wedge (\bigwedge_{i=1}^{n} (\neg t \vee \neg z_i \vee r))$$

where $\phi$ is viewed as a set of clauses $C$. More readably, the theory is equivalent to:

$$t \Rightarrow [(\bigwedge_{i=1}^{m} y_i \equiv \neg q_i) \wedge (r \Rightarrow \phi) \wedge ((\bigvee_{i=1}^{n} z_i) \Rightarrow r)]$$

First, note a few obvious things about this construction:

- $o^t$ must be included in any plan.

- Once $o^t$ is applied, no action can be applied anymore.

- Before $o^t$ is applied, either $o^{x_i}$ or $o^{\neg x_i}$ must be applied, for every $1 \leq i \leq n$.

- The theory is "switched off", i.e., made irrelevant because $t$ is false, up to the point where $o^t$ is applied.

That is, any plan for this task must first apply $o^{x_i}$ or $o^{\neg x_i}$, for every $1 \leq i \leq n$, thereby choosing a value for every $x_i$. Then, $o^t$ must be applied and the plan must stop. Before applying $o^t$, no changes are made to the states except that the values of $x_i$ are set and that the $f_i$ are made true one after the other. Hence, the belief $b$ in which $o^t$ is applied contains a single state $s$ which corresponds to an extension of $\phi_0$ with a value assignment for $X$, where the values of the $f_i$ have been flipped. We denote the value assignment for $X$ in $s$ with $a_X$. We further denote $b' := res(b, o^t)$. Note that $o^t$ is consistent: any interpretation that sets $r$ and all $z_i$ to 0, besides setting the immediate effects $t \wedge g \wedge \neg h$, satisfies $\Phi_{IC} \wedge \text{eff}_{o^t}$. Obviously, all of the applications of $o^{x_i}$ and $o^{\neg x_i}$ are consistent as well.

The theory conjuncts $y_i \equiv \neg q_i$ make sure that each $w \in b'$ makes exactly one of $y_i, q_i$ true. In particular, the different assignments to $Y$ are incomparable with respect to set inclusion. Hence, we have that for every assignment $a_Y$ of truth values to $Y$, there exists a state $s' \in b'$ that complies with $a_Y$: $a_Y$ is satisfiable together with $\Phi_{IC} \wedge \text{eff}_{o^t}$, and any other assignment $a'_Y$ is more distant from $s$ in at least one variable (e.g., if $a'_Y(y_i) = 1$ and $a_Y(y_i) = 0$ then $a_Y$ is closer to $s$ than $a'_Y$ regarding the interpretation of $q_i$).

We now prove that, if there exists a plan $\vec{a}$ yielding assignment $a_X$, then $\exists X. \forall Y. \exists Z. \phi[X, Y, Z]$ is valid. Let $a_Y$ be an arbitrary truth value assignment to $Y$. Then we have a state $s' \in b'$ that complies with $a_X$ and $a_Y$. $a_X$ and $a_Y$ are satisfiable together with $\Phi_{IC} \wedge \text{eff}_{o^t}$. With the above, any other assignment $a'_Y$ is more distant from $s$ in at least one variable. And, of course, if one deviates from $a_X$ then one is more distant from $s$ in the respective variable. Since $\vec{a}$ is a plan, we have $s' \models r$. Therefore, due to the theory conjunct $r \Rightarrow \phi$, we have $s' \models \phi$. Obviously, the values assigned to $Z$ by $s'$ satisfy $\phi$ for $a_X$ and $a_Y$. This proves the claim because $a_Y$ can be chosen arbitrarily.

For the other direction, say $\exists X. \forall Y. \exists Z. \phi[X, Y, Z]$ is valid. Let $a_X$ be an assignment to $X$ so that $\forall Y. \exists Z. \phi[a_X/X, Y, Z]$ is valid. Let $\vec{a}$ be the corresponding plan, i.e., $\vec{a}$ first applies, for $1 \leq i \leq n$, either $o^{x_i}$ or $o^{\neg x_i}$ according to $a_X$. Thereafter, $\vec{a}$ applies $o^t$. Assume that $\vec{a}$ is not a plan. Then we have $s' \in b'$ so that $s' \not\models r$. But then, due to the theory conjunct $(\bigvee_{i=1}^{n} z_i) \Rightarrow r$, we have that $s$ sets all $z_i$ to false. Now, because $\forall Y. \exists Z. \phi[a_X/X, Y, Z]$ is valid, there exists a truth value assignment $a_Z$ to $Z$ that complies with the setting of all $x_i$, $y_i$, and $q_i$ in $s$. Obtain $s''$ by modifying $s'$ to comply with $a_Z$, and setting $r$ to 1. We have that $s'' \models \Phi_{IC} \wedge \text{eff}_{o^t}$. But then, $s''$ is closer to $s$ than $s'$, and hence $s' \notin b'$ in contradiction. This concludes the argument. $\quad\square$

**Theorem 3.** *Assume a $\mathcal{WSC}$ task. The decision problem asking whether there exists a plan is undecidable.*

**Proof:** This result holds even with an empty background theory, a complete specification of the initial state, predicates of arity at most 2, operators of arity at most 2, a goal with no variables at all (arity 0), and only positive literals in preconditions and the goal. The result follows with a minor modification of Tom Bylander's proof (Bylander, 1994) that plan existence in propositional STRIPS planning is PSPACE-complete.[26] The original proof proceeds by a generic reduction, constructing a STRIPS task for a Turing Machine (TM) with polynomially bounded space. The latter restriction is necessary to model the machine's tape: tape cells are pre-created for all positions within the bound. What makes the difference between PSPACE-membership and undecidability is the ability to create constants. We can introduce simple operators that allow us to extend the tape, at both ends.

In detail, say the TM has (a finite number of) states $q$ and tape alphabet symbols $a$ (where $b$ is the blank); $\delta$ is the transition function, $q_0$ is the initial state, and $F$ is the set of accepting states; $\omega$ is the input word. Our planning encoding contains the following predicates. $State(q)$ indicates that the current TM state is $q$. $In(a, c)$ indicates that the current content of tape cell $c$ is $a$. $Neighbors(c, c')$ is true iff $c'$ is the (immediate) right neighbor of $c$. $At(c)$ indicates that the current position of the TM head is $c$. $Rightmost(c)$ ($Leftmost(c)$) is true iff $c$ currently has no right (left) neighbor. The set of initial constants contains all states $q$, all alphabet symbols $a$, and tape cells $c$ corresponding to $\omega$. By the initial literals, all the propositions over these constants are assigned truth values as obvious. For every transition $(q, a, q', a', R) \in \delta$ we include an operator:

$$(\{x, x'\}, State(q) \wedge In(x, a) \wedge Neighbors(x, x') \wedge At(x),$$

$$\emptyset, State(q') \wedge \neg State(q) \wedge In(x, a') \wedge \neg In(x, a) \wedge At(x') \wedge \neg At(x)).$$

Obviously, this encodes exactly that transition. We do likewise for transitions $(q, a, q', a', L) \in \delta$. To model the final states, we introduce a 0-ary predicate $G$, and include for each $q \in F$ an operator:

$$(\emptyset, State(q), \emptyset, G)$$

We finally include the operators:

$$(\{x\}, Rightmost(x), \{x'\}, Neighbors(x, x') \wedge In(b, x') \wedge Rightmost(x') \wedge \neg Rightmost(x))$$

and

$$(\{x'\}, Leftmost(x'), \{x\}, Neighbors(x, x') \wedge In(b, x) \wedge Leftmost(x) \wedge \neg Leftmost(x'))$$

With these definitions, it is easy to verify that there exists a plan iff the TM can reach an accepting state on $\omega$. □

**Lemma 1.** *Assume a $\mathcal{WSC}|_{fwd}$ task, a reachable state $s$, and an action $a$. Then $res(s, a) = res|_{fwd}(s, a)$.*

---

26. Propositional STRIPS is like our framework, but with an empty background theory, a complete specification of the initial state, a goal with no variables, only positive literals in preconditions and the goal, and with no output parameters in the operators.

**Proof:** If $a$ is not applicable to $s$, then the claim holds trivially. Consider the other case. By Equation 3, $res(s, a)$ is defined as

$$res(s, a) := \begin{cases} \{(C', I') \mid C' = C_s \cup E_a, I' \in \min(s, C', \Phi_{IC} \wedge \text{eff}_a)\} & appl(s, a) \\ \{s\} & \text{otherwise} \end{cases}$$

where $\min(s, C', \Phi_{IC} \wedge \text{eff}_a)$ is the set of all $C'$-interpretations that satisfy $\Phi_{IC} \wedge \text{eff}_a$ and that are minimal with respect to the partial order defined by $I_1 \leq_s I_2$ :iff for all propositions $p$ over $C_s$, if $I_2(p) = I_s(p)$ then $I_1(p) = I_s(p)$.

It is obvious that $res|_{fwd}(s, a) \subseteq res(s, a)$ – if $I_{s'}$ satisfies $\Phi_{IC} \wedge \text{eff}_a$ and $I_{s'}$ is identical to $I_s$ on the propositions over $C_s$, then in particular $I_{s'}$ is minimal according to $\leq_s$.

For the other direction, let $s' \in res(s, a)$. Assume that $I_{s'}(p) \neq I_s(p)$ for some proposition $p$ over $C_s$. Define $s''$ to be equal to $s'$ except that $I_{s''}(p) := I_s(p)$. Obviously, $I_{s'} \not\leq_{s''} I_2$. It now suffices to show that $s'' \models \Phi_{IC} \wedge \text{eff}_a$: then, we get $I_{s'} \notin \min(s, C', \Phi_{IC} \wedge \text{eff}_a)$ in contradiction, hence $I_{s'}$ agrees with $I_s$ on all propositions $p$ over $C_s$, hence $s' \in res|_{fwd}(s, a)$.

As before, denote with $P^{C_s + E_a}$ the set of all propositions with arguments in $C_s \cup E_a$, and with at least one argument in $E$, and denote with $\Phi_{IC}[C_s + E_a]$ the instantiation of $\Phi_{IC}$ with all constants from $C_s \cup E_a$, where in each clause at least one variable is instantiated from $E_a$. To see that $s'' \models \Phi_{IC} \wedge \text{eff}_a$, consider first that this is equivalent to $s'' \models \Phi_{IC}[C_s \cup E_a] \wedge \text{eff}_a$, which in turn is equivalent to $s'' \models \Phi_{IC}[C_s] \wedge \Phi_{IC}[C_s + E_a] \wedge \text{eff}_a$. In the last formula, because the task is in $\mathcal{WSC}|_{fwd}$, $\Phi_{IC}[C_s]$ speaks only over the propositions $P^{C_s}$, whereas $\Phi_{IC}[C_s + E_a] \wedge \text{eff}_a$ speaks only over the propositions $P^{C_s + E_a}$. So we can treat these two parts separately. We have $s'' \models \Phi_{IC}[C_s]$ because $s \models \Phi_{IC}[C_s]$ by prerequisite since $s$ is reachable. We have $s'' \models \Phi_{IC}[C_s + E_a] \wedge \text{eff}_a$ by definition. This concludes the argument. □

**Theorem 4.** *Assume a $\mathcal{WSC}|_{fwd}$ task with fixed arity, and a sequence $\langle a_1, \ldots, a_n \rangle$ of actions. It is coNP-complete to decide whether $\langle a_1, \ldots, a_n \rangle$ is a plan.*

**Proof:** Hardness is obvious, considering an empty sequence. Membership can be shown by the following guess-and-check argument. Say $C$ is the union of $C_0$ and all output constants appearing in $\langle A_1, \ldots, A_n \rangle$. We guess an interpretation $I$ of all propositions over $\mathcal{P}$ and $C$. Further, for each $1 \leq t \leq n$, we guess a set $C_t$ of constants. We can then check in polynomial time whether $I$ and the $C_t$ correspond to an execution of $\langle A_1, \ldots, A_n \rangle$. For $1 \leq t \leq n$ and $a \in A_t$, say that $a$ is applicable if $I \models \text{pre}_a$, $C_a \subseteq C_t$, and $E_a \cap C_t = \emptyset$. First, we assert $I \models \Phi_{IC}$. Second, for all $t$ and for all $a \in A_t$, assert that, if $a$ is applicable, then $I \models \text{eff}_a$. Third, assert that $C_{t+1} = C_t \cup \{E_a \mid a \in A_t, a \text{ is applicable}\}$. Using Lemma 1, it is easy to see that $I$ and the $C_t$ correspond to an execution iff all three assertions hold. Note that $I$ needs not be time-stamped because once an action has generated its outputs then the properties of the respective propositions remain fixed forever. The claim follows because, with fixed arity, we can also test in polynomial time whether $I$ and $C_n$ satisfy $\phi_G$. A guess of $I$ and $C_t$ is successful if it corresponds to an execution and does not satisfy $\phi_G$. Obviously, $\langle A_1, \ldots, A_n \rangle$ is a plan iff there is no such guess of $I$ and $C_t$. □

**Theorem 5.** *Assume a $\mathcal{WSC}|_{fwd}$ task with fixed arity, and a natural number $b$ in unary representation. It is $\Sigma_2^p$-complete to decide whether there exists a plan of length at most $b$.*

**Proof:** For membership, guess a sequence of at most $b$ actions. By Theorem 1, we can check with a $\Pi_2^p$ oracle whether the sequence is a plan.

To prove hardness, assume a QBF formula $\exists X.\forall Y.\phi[X,Y]$ where $\phi$ is in DNF normal form. (This formula class is complete for $\Sigma_2^p$.) Say $X = x_1, \ldots, x_n$, $Y = y_1, \ldots, y_m$, and $\phi = \phi_1 \vee \cdots \vee \phi_k$. We design a $\mathcal{WSC}|_{fwd}$ task which has a plan iff $\exists X.\forall Y.\phi[X,Y]$ is true. The key construction is to use outputs for the creation of "time steps", and to allow setting $x_i$ only at time step $i$. The $y_i$ can take on arbitrary values. Once all $x_i$ are set, one operator per $\phi_k$ allows to achieve the goal given $\phi_k$ is true. The main property we need to ensure in the construction is that each $x_i$ can be set at most once, i.e., either to 1 or to 0. Then there is a plan for the task iff one can set $X$ so that, for all $Y$, at least one $\phi_i$ is true – which is the case iff $\exists X.\forall Y.\phi[X,Y]$ is true.

The predicates for our task are $\mathcal{P} = \{x_1(.), \ldots, x_n(.), y_1(), \ldots, y_m(), time(.), start(.), next(..), goal(.)\}$. We indicate predicate arity here by the number of points in the parentheses. For example, the predicate $next(..)$ has arity 2. The theory $\Phi_{IC}$ is empty. The initial constants are $C_0 = \{t_0\}$. The initial literals are $\phi_0 = time(t_0)$. The goal is $\exists y.goal(y)$. The operators are as follows:

- For all $1 \leq i \leq n$, we have: $o^{x_i 1} = (\{t_0, \ldots, t_{i-1}\}, start(t_0) \wedge next(t_0, t_1) \wedge \cdots \wedge next(t_{i-2}, t_{i-1}), \{t_i\}, time(t_i) \wedge next(t_{i-1}, t_i) \wedge x_i(t_i))$. Such an operator allows generating time step $i$, and setting $x_i$ to 1 at that step.

- For all $1 \leq i \leq n$, we have: $o^{x_i 0} = (\{t_0, \ldots, t_{i-1}\}, start(t_0) \wedge next(t_0, t_1) \wedge \cdots \wedge next(t_{i-2}, t_{i-1}), \{t_i\}, time(t_i) \wedge next(t_{i-1}, t_i) \wedge \neg x_i(t_i))$. Such an operator allows generating time step $i$, and setting $x_i$ to 0 at that step.

- We will define a value $B$ below. For all $n \leq j < n + B$, we have: $o^{t_j} = (\{t_0, \ldots, t_{j-1}\}, start(t_0) \wedge next(t_0, t_1) \wedge \cdots \wedge next(t_{j-2}, t_{j-1}), \{t_j\}, time(t_j) \wedge next(t_{j-1}, t_j))$. These operators allow increasing the time step from $n$ to $n + B$.

- For $1 \leq i \leq k$, say $\phi_i = xl_{x_{j_1}} \wedge \cdots \wedge xl_{x_{j_{xn}}} \wedge yl_{y_{j_1}} \wedge \cdots \wedge yl_{y_{j_{yn}}}$ where $xl_j \in \{x_j, \neg x_j\}$ and $yl_j \in \{y_j, \neg y_j\}$. We have: $o^{\phi_i} = (\{t_0, \ldots, t_{n+B}\}, start(t_0) \wedge next(t_0, t_1) \wedge \cdots \wedge next(t_{n+B-1}, t_{n+B}) \wedge xl_{x_{j_1}}(t_{x_{j_1}}) \wedge \cdots \wedge xl_{x_{j_{xn}}}(t_{x_{j_{xn}}}) \wedge yl_{y_{j_1}}() \wedge \cdots \wedge yl_{y_{j_{yn}}}(), \{c\}, goal(c))$. Such an operator allows to achieve the goal after time step $n + B$, provided the respective $\phi_i$ is true. Note here that the $x_j$ precondition literals refer to time step $t_j$, i.e., the value set for $x_j$ at an earlier time step, while the $y_j$ precondition literals have no arguments and refer to the initial values of $y_j$, which are arbitrary.

Assume we choose any value for $B$ (polynomial in the input size). If $\exists X.\forall Y.\phi[X,Y]$ is true, then, obviously, we can find a plan of size $n + B + k$. We apply an $o^{x_i 1}$ or $o^{x_i 0}$ operator for each $x_i$, depending on whether $x_i$ must be set to 1 or 0. We apply $B$ operators $o^{t_j}$. We apply all operators $o^{\phi_i}$. The respective input parameter instantiations are all obvious.

The opposite direction – proving truth of $\exists X.\forall Y.\phi[X,Y]$ based on a plan – is more problematic. The plan might "cheat" by setting some $x_i$ to both 1 and 0. The reason why our construction is so complicated is to be able to avoid precisely this case, based on specifying a strict enough plan length bound $b$. The key property is that, in order to cheat for $x_i$, the plan has to generate *two* sequences of time steps $t_i, \ldots, t_{n+B}$. Therefore, a lower bound on the length for a cheating plan is $n + 2B$. As we have already seen, an upper bound on the length of a non-cheating plan is $n + B + k$. To determine our plan length bound $b$, we now simply choose a $B$ so that any cheating plan will have to use too many steps: $n + 2B > n + B + k$ is the case iff $B > k$. So we can set $B := k + 1$, and obtain $b := n + 2k + 1$. With this bound $b$, any plan will proceed by setting each $x_i$ to a value ($n$ actions), increasing the time step to $n+B = n+k+1$ ($k+1$ actions), and applying a sufficient subset of the $o^{\phi_i}$ (at most $k$ actions). If the plan cheats, then it needs to apply at least $n+2B = n+2k+2$ actions before being able to apply $o^{\phi_i}$ actions exploiting different value settings for a $x_i$. This concludes the argument. $\qquad\square$

**Theorem 6.** *Assume a $\mathcal{WSC}|_{fwd}$ task. The decision problem asking whether there exists a plan is undecidable.*

**Proof:** We reduce from the halting problem for Abacus machines, which is undecidable. An Abacus machine consists of a tuple of integer variables $v_1, \ldots, v_k$ (ranging over all positive integers including 0), and a tuple of instructions $I_1, \ldots, I_n$. A state is given by the content of $v_1, \ldots, v_k$ plus the index $pc$ of the active instruction. The machine stops iff it reaches a state where $pc = n$. All $v_i$ are initially 0, and $pc$ is initially 0. There are two kinds of instructions. $I_i :$ INC $j$; GOTO $I_{i'}$ increments the value of $v_j$ and jumps to $pc = i'$. $I_i :$ DEC $j$; BRANCH $I_{i^+}/I_{i^0}$ asks whether $v_j = 0$. If so, it jumps to $pc = i^0$. Otherwise, it decrements the value of $v_j$ and jumps to $pc = i^+$.

We map an arbitrary abacus program to a $\mathcal{WSC}|_{fwd}$ instance as follows:

- *Predicates:* $number(v)$, $zero(v)$, $succ(v', v)$, $value_1(v, t)$, ..., $value_k(v, t)$, $instruction_1(t)$, ..., $instruction_n(t)$

- *Background theory:* none (i.e., the trivial theory)

- *Operators:*

  – An operator $\langle \{v\}, \{number(v)\}, \{v'\}, \{number(v'), succ(v', v)\}\rangle$

  – For instructions of the form $I_i :$ INC $j$; GOTO $I_{i'}$, the operator

  $$\langle \{v_1, \ldots, v_k, t\},$$
  $$\{instruction_i(t), value_1(v_1, t), \ldots, value_k(v_k, t), succ(v', v_j)\},$$
  $$\{t'\},$$
  $$\{instruction_{i'}(t'), value_1(v_1, t'), \ldots, value_{j-1}(v_{j-1}, t'), value_j(v', t'),$$
  $$value_{j+1}(v_{j+1}, t'), \ldots, value_k(v_k, t')\}\rangle.$$

  – For instructions of the form $I_i :$ DEC $j$; BRANCH $I_{i^+}/I_{i^0}$, the operators

  $$\langle \{v_1, \ldots, v_k, t\},$$
  $$\{instruction_i(t), value_1(v_1, t), \ldots, value_k(v_k, t), succ(v_j, v')\},$$
  $$\{t'\},$$
  $$\{instruction_{i^+}(t'), value_1(v_1, t'), \ldots, value_{j-1}(v_{j-1}, t'), value_j(v', t'),$$
  $$value_{j+1}(v_{j+1}, t'), \ldots, value_k(v_k, t')\}\rangle.$$

  and

  $$\langle \{v_1, \ldots, v_k, t\},$$
  $$\{instruction_i(t), value_1(v_1, t), \ldots, value_k(v_k, t), zero(v_j)\},$$
  $$\{t'\},$$
  $$\{instruction_{i^0}(t'), value_1(v_1, t'), \ldots, value_{j-1}(v_{j-1}, t'), value_j(v_j, t'),$$
  $$value_{j+1}(v_{j+1}, t'), \ldots, value_k(v_k, t')\}\rangle.$$

- *Initial constants:* $v_0, t_0$

- *Initial literals:* $number(v_0) \wedge zero(v_0) \wedge value_1(v_0, t_0) \wedge \cdots \wedge value_k(v_0, t_0) \wedge instruction_1(t_0)$

- *Goal condition:* $\exists t.instruction_n(t)$

We now describe the intuitive meaning of the constants and predicates. There are two kinds of constants: numbers, which represent natural numbers (including 0), and time points, which represent computation steps of the Abacus machine. Variables that refer to time points are denoted as $t$ or $t'$ above. All other variables represent numbers.

Three predicates refer to numbers exclusively: $number(v)$ is true iff $v$ encodes a natural number (and not a time point); $zero(v)$ is true iff $v$ encodes the number 0; and $succ(v', v)$ is true iff $v'$ encodes the number that is one larger than the number encoded by $v$. The reduction does not enforce that every number is uniquely represented (e.g., there may be several representations of the number 3), but such a unique representation is not necessary. It is guaranteed that the number 0 is uniquely represented, though.

The remaining predicates encode configurations of the Abacus machine: $value_i(v, t)$ is true iff, at time point $t$, the $i$-th Abacus variable holds the number represented by $v$, and $instruction_j(t)$ is true iff the current instruction at time point $t$ is $I_j$.

Obviously, from an accepting run of the Abacus machine we can extract a plan for the task, and vice versa. This proves the claim. □

To prove Theorems 7 and 8, we first establish a core lemma from which both theorems follow relatively easily. We need a few notations. We denote beliefs (states) in $(\mathcal{P}, \Phi_{IC}, \mathcal{O}, C_0, \phi_0, \phi_G)$ with $b(s)$, and we denote beliefs (states) in $(\mathcal{P}', \mathcal{A}, \phi_0', \phi_G')$ with $\bar{b}(\bar{s})$. Assume a sequence $\langle a_1, \ldots, a_i \rangle$ of non-goal achievement actions. Then we denote $b := res(b_0, \langle a_1, \ldots, a_i \rangle)$ and $\bar{b} := res(\bar{b}_0, \langle a_1, \ldots, a_i \rangle)$. Note here that we overload the $res$ function to also denote state transitions in the compiled task formalism. Further, for a state $\bar{s}$, by $C(\bar{s}) := \{c \mid \bar{s}(Ex(c)) = 1\}$ we denote the constants that exist in $\bar{s}$. We denote by $\equiv_C$ the relation over states $\bar{s}$ and $\bar{s}'$ that is true iff $C(\bar{s}) = C(\bar{s}')$ and $\bar{s}|_{C(\bar{s})} = \bar{s}'|_{C(\bar{s})}$. $\equiv_C$ is an equivalence relation, where equivalent states agree on which constants exist and how they are interpreted. Note that every state $\bar{s}$ reachable in the compiled task satisfies $s \models \Phi_{IC} \wedge \phi_0 \wedge \bigwedge_{o \in \mathcal{O}} \text{eff}_o[E_o]$. Note further that $\Phi_{IC} \wedge \phi_0 \wedge \bigwedge_{o \in \mathcal{O}} \text{eff}_o[E_o]$ is actually satisfiable be prerequisite, unless $\Phi_{IC} \wedge \phi_0$ is unsatisfiable, because the outputs are instantiated with unique constants and the operators are consistent. For a state $s$, we define $[s] :=$

$$\{\bar{s} \mid \bar{s} \text{ defined over } C_0 \cup \bigcup_{o \in \mathcal{O}} E_o, C(\bar{s}) = C_s, \bar{s}|_{C_s} = I_s, s \models \Phi_{IC} \wedge \phi_0 \wedge \bigwedge_{o \in \mathcal{O}} \text{eff}_o[E_o]\}$$

That is, $[s]$ is the equivalence class of states $\bar{s}$ reachable in the compiled task that agree with $s$ on which constants exist and how they are interpreted.

**Lemma 3** *Assume a $\mathcal{WSC}|_{sfwd}$ task without inconsistent operators. Let $\langle a_1, \ldots, a_i \rangle$ consist of non-goal achievement actions, and let $b := res(b_0, \langle a_1, \ldots, a_i \rangle)$ and $\bar{b} := res(\bar{b}_0, \langle a_1, \ldots, a_i \rangle)$. Then $\bar{b} = \bigcup_{s \in b}[s]$.*

**Proof:** The proof is by induction over $i$. In the base case, we have $i = 0$, i.e., $b = b_0$ and $\bar{b} = \bar{b}_0$. We have $b_0 =$

$$\{s \mid C_s = C_0, I_s \models \Phi_{IC} \wedge \phi_0\}$$

On the other hand, we have $\overline{b}_0 =$

$$\{\overline{s} \mid C(\overline{s}) = C_0, \overline{s} \models \Phi_{IC} \wedge \phi_0 \wedge \bigwedge_{o \in \mathcal{O}} \mathsf{eff}_o[E_o]\}$$

Obviously, the latter is comprised of one equivalence class for each possibility to assign the propositions over $C_0$ in a way compliant with $\Phi_{IC} \wedge \phi_0$. This is exactly the claim.

In the inductive case, say we add another action $a$ to $\langle a_1, \ldots, a_i \rangle$. By induction assumption, we have $\overline{b} = \bigcup_{s \in b}[s]$. We need to prove that $res(\overline{b}, a) = \bigcup_{s' \in res(b,a)}[s']$. Obviously, it suffices to prove that, for every $s \in b$, we have $res([s], a) = \bigcup_{s' \in res(s,a)}[s']$. First, say $a$ is not applicable to $s$. Then $s$ is neither applicable in any $\overline{s} \in [s]$, and we have $res([s], a) = [s] = \bigcup_{s' \in res(s,a)}[s']$. Second, say $a$ is applicable to $s$. Then by Lemma 1 we have $res(s, a) =$

$$\{(C_s \cup E_a, I') \mid I'|_{C_s} = I_s, I' \models \Phi_{IC} \wedge \mathsf{eff}_a\}$$

On the other hand, we have $res([s], a) =$

$$\{\overline{s}' \mid ex.\ \overline{s} \in [s], C(\overline{s}') = C(\overline{s}) \cup E_a, \overline{s}'|_{C(\overline{s})} = \overline{s}, \overline{s}' \models \Phi_{IC} \wedge \phi_0 \wedge \bigwedge_{o \in \mathcal{O}} \mathsf{eff}_o[E_o]\}$$

We can re-write the latter into

$$\{\overline{s}' \mid C(\overline{s}') = C_s \cup E_a, \overline{s}'|_{C_s} = I_s, \overline{s}' \models \Phi_{IC} \wedge \phi_0 \wedge \bigwedge_{o \in \mathcal{O}} \mathsf{eff}_o[E_o]\}$$

Obviously, as desired, the latter set is comprised of one equivalence class for each possibility to assign the propositions over $C_s \cup E_a$ in a way compliant with $s$ and $\Phi_{IC} \wedge \mathsf{eff}_a$. This concludes the argument. $\qquad \square$

**Theorem 7.** *Assume a $\mathcal{WSC}|_{sfwd}$ task $(\mathcal{P}, \Phi_{IC}, \mathcal{O}, C_0, \phi_0, \phi_G)$ without inconsistent operators, and a plan $\langle a_1, \ldots, a_n \rangle$ for the compiled task $(\mathcal{P}', \mathcal{A}, \phi'_0, \phi'_G)$. Then the sub-sequence of non-goal achievement actions in $\langle a_1, \ldots, a_n \rangle$ is a plan for $(\mathcal{P}, \Phi_{IC}, \mathcal{O}, C_0, \phi_0, \phi_G)$.*

**Proof:** If $\Phi_{IC} \wedge \phi_0$ is unsatisfiable, there is nothing to prove, because the start belief of the original task is empty. For the non-trivial case, first note that, in any plan for the compiled task, the goal achievement actions can be moved to the back of the plan. Hence, without loss of generality, we can assume that $\langle a_1, \ldots, a_i \rangle$ consist entirely of non-goal achievement actions, and $\langle a_{i+1}, \ldots, a_i \rangle$ consist entirely of goal achievement actions. Denote $b := res(b_0, \langle a_1, \ldots, a_i \rangle)$ and $\overline{b} := res(\overline{b}_0, \langle a_1, \ldots, a_i \rangle)$. By Lemma 3, we have $\overline{b} = \bigcup_{s \in b}[s]$. Since $\langle a_1, \ldots, a_n \rangle$ is a plan for the compiled task, every $\overline{s} \in \overline{b}$ has a tuple of constants satisfying $\phi_G$. With $\overline{b} = \bigcup_{s \in b}[s]$, it follows that every $s \in b$ satisfies $\phi_G$. $\qquad \square$

**Theorem 8.** *Assume a $\mathcal{WSC}|_{sfwd}$ task $(\mathcal{P}, \Phi_{IC}, \mathcal{O}, C_0, \phi_0, \phi_G)$ without inconsistent operators, and a plan $\langle a_1, \ldots, a_n \rangle$ where every operator $o$ appears with at most one instantiation $E_o$ of the outputs. Then $\langle a_1, \ldots, a_n \rangle$ can be extended with goal achievement actions to form a plan for the compiled task $(\mathcal{P}', \mathcal{A}, \phi'_0, \phi'_G)$ obtained using the outputs $E_o$.*

**Proof:** Denote $b := res(b_0, \langle a_1, \ldots, a_n \rangle)$ and $\overline{b} := res(\overline{b}_0, \langle a_1, \ldots, a_n \rangle)$. By Lemma 3, we have $\overline{b} = \bigcup_{s \in b}[s]$. Since $\langle a_1, \ldots, a_n \rangle$ is a plan, every $s \in b$ satisfies $\phi_G$. With $\overline{b} = \bigcup_{s \in b}[s]$, it follows that every $\overline{s} \in \overline{b}$ has a tuple of constants satisfying $\phi_G$. Attaching all the respective goal achievement actions yields a plan for the compiled task. $\qquad \square$